\documentclass[sigconf, nonacm]{acmart}
\AtBeginDocument{%
  }

\usepackage{pifont}
\usepackage{makecell}
\usepackage{multirow}
\setcopyright{none}
\copyrightyear{2018}
\acmYear{2018}
\acmDOI{XXXXXXX.XXXXXXX}
\acmConference[Conference acronym 'XX]{Make sure to enter the correct
  conference title from your rights confirmation email}{June 03--05,
  2018}{Woodstock, NY}
\acmISBN{978-1-4503-XXXX-X/2018/06}

\begin{document}

\title{Seeing the End at Step Zero: Accelerating Diffusion MLLMs via MLP Sparsity-Aware Truncation}

\author{Qicheng Zhao}
\email{22447067@zju.edu.cn}
\affiliation{%
  \institution{Zhejiang University}
  \city{Hangzhou}
  \country{China}
}

\author{Qi Sun}
\email{qisunchn@zju.edu.cn}
\affiliation{%
  \institution{Zhejiang University}
  \city{Hangzhou}
  \country{China}
}

\author{Zheyu Yan}
\authornote{Corresponding author.}
\email{zyan2@zju.edu.cn}
\affiliation{%
  \institution{Zhejiang University}
  \city{Hangzhou}
  \country{China}
}
\renewcommand{\shortauthors}{Zhao et al.}

\begin{abstract}
Diffusion Multimodal Large Language Models (DMLLMs) are highly effective for multimodal reasoning, yet their inference efficiency is significantly hindered by fixed-length generation constraints. Since the actual output length is unknown, output sequences are padded to a predefined maximum length, resulting in substantial redundant computation over unnecessary \texttt{[EOS]} tokens. In this work, we discover that DMLLMs implicitly reveal their valid semantic boundary at the very first denoising step through a distinct shift in MLP activation sparsity. Leveraging this observation, we propose Seer, a training-free framework that detects this boundary using a Signal-to-Noise Ratio (SNR)-based criterion and performs one-shot truncation of the redundant suffix for all subsequent computations. To preserve these theoretical gains during batched serving, Seer incorporates a hybrid execution strategy that maximizes throughput while seamlessly accommodating dynamic sequence lengths. Experimental results demonstrate that Seer effectively eliminates padding waste, accelerating throughput by up to $\sim$31$\times$. Across 9 benchmarks, Seer robustly maintains overall performance and even improves accuracy on complex visual tasks by mitigating noise leakage (e.g., DocVQA score increases from 63.52 to 63.66), offering a highly efficient, plug-and-play solution for DMLLM acceleration.
\end{abstract}




\maketitle

\section{Introduction}

Diffusion Multimodal Large Language Models (DMLLMs), represented by LaViDa~\cite{li2025lavida}, aA~\cite{yang2025mmada}, and LLaDA-V~\cite{you2025llada}, have recently emerged as a compelling alternative to autoregressive multimodal models. Instead of decoding text strictly token by token, DMLLMs formulate generation as a global iterative denoising process over a predefined fixed-length sequence window. This paradigm naturally enables bidirectional interaction between visual and textual tokens at every diffusion step, making it attractive for holistic multimodal reasoning. However, it also introduces a severe inference bottleneck. Because denoising updates the entire sequence concurrently, the model must repeatedly process all positions layer by layer and step by step, even when only a small prefix contains semantically valid output and the remaining positions are already occupied by redundant \texttt{[EOS]} tokens.

\begin{figure}
    \centering
    
    \includegraphics[trim=0 0 0 0, clip, width=\linewidth]{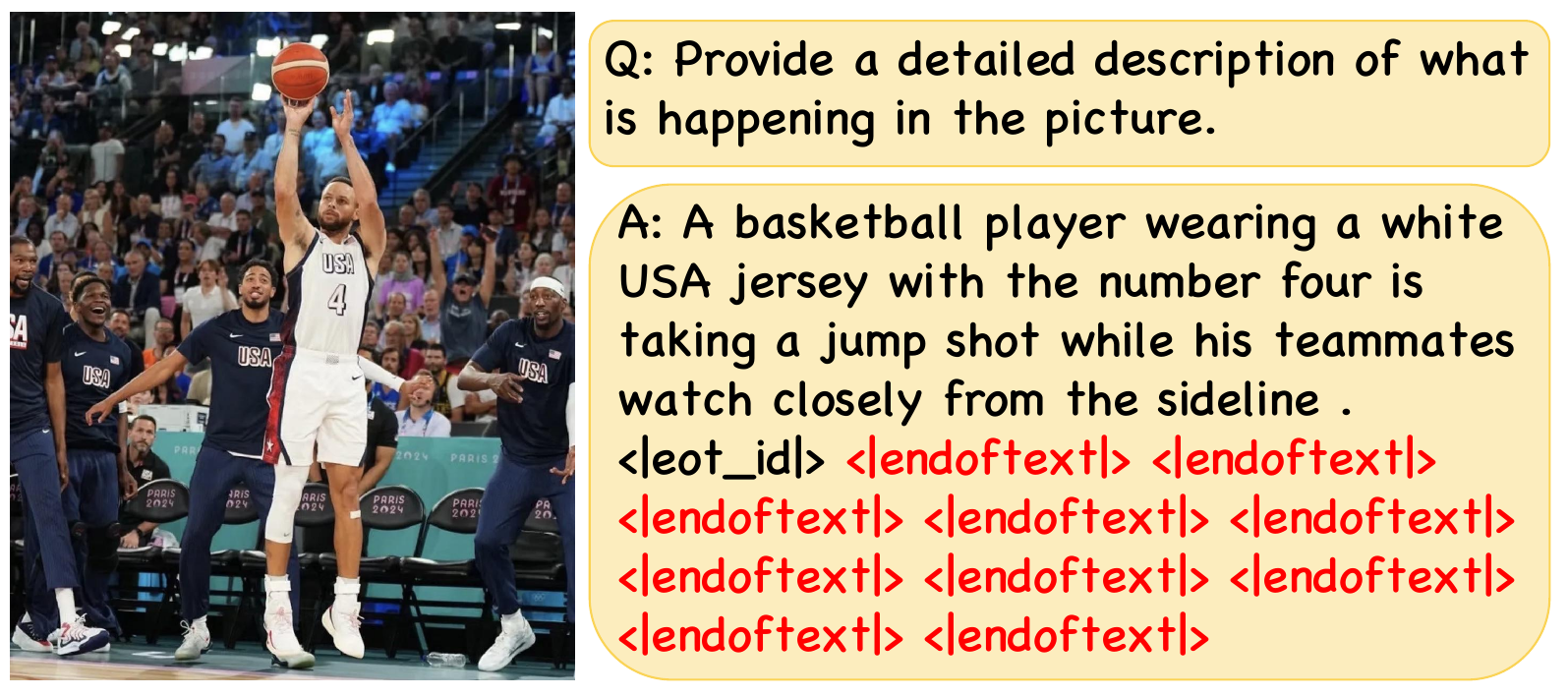}
    \Description{A diagram illustrating the "Curse of Padding" in Diffusion Multimodal Large Language Models. It shows a short generated text response to a basketball image, followed by a very long sequence of redundant end-of-text padding tokens filling up a predefined fixed-length window, representing computational waste.}
    \vspace{-6mm}
    \caption{An illustration of the ``Curse of Padding'' in DMLLMs. Due to the fixed-length generation window, a short semantically valid response is inevitably followed by a massive sequence of redundant padding tokens (e.g., \texttt{<|endoftext|>}), leading to severe computational waste.}
    \label{fig:curse_of_padding}
    \vspace{-3mm}  
\end{figure}

As illustrated in Figure~\ref{fig:curse_of_padding}, in fixed-length generation, because the ground truth response sequence length is unknown, even a short response is designated a very long generation window and the generated response is then inevitably padded with a massive sequence of redundant tokens (e.g., \texttt{<|endoftext|>}). This overhead, which we refer to as \textit{the curse of padding}, is particularly severe in multimodal settings. In pure language diffusion models, a redundant suffix mainly interacts with other text tokens. In DMLLMs, by contrast, modern architectures often ingest large visual contexts from high-resolution images or multi-frame videos, so redundant text positions repeatedly participate in dense cross-modal attention with a large visual prefix. 
This large prefix, along with the quadratic nature of the bidirectional generation, significantly exacerbates the overhead incurred by longer generation windows. 

\begin{figure}[t]
    \centering
    
    \includegraphics[trim=0 0 0 0, clip, width=\linewidth]{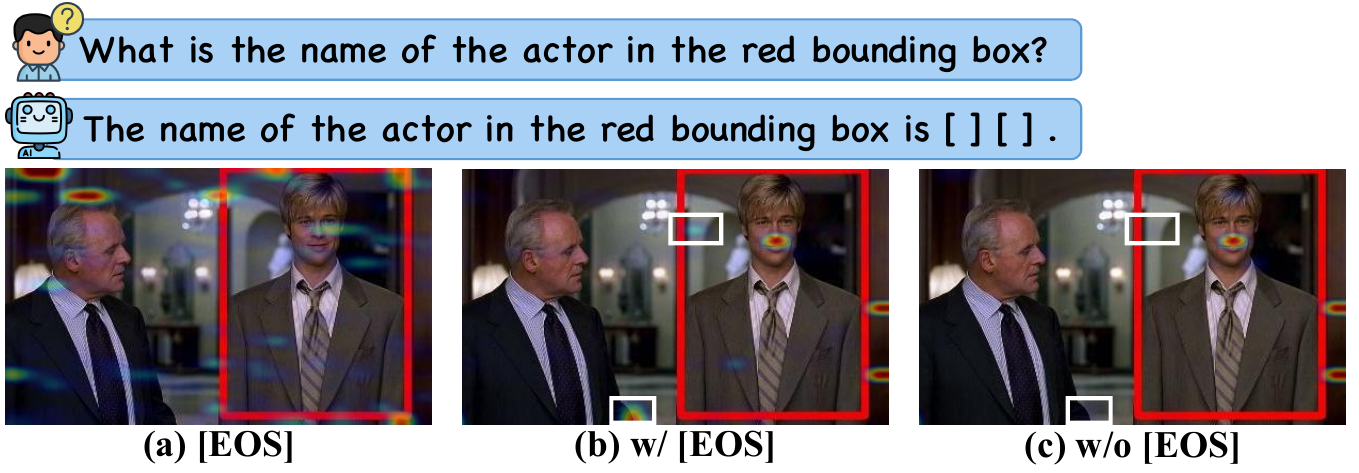}
    \Description{Three comparative visual attention heatmaps overlaying an image of an actor. Image (a) shows redundant EOS tokens attending to irrelevant background noise. Image (b) shows valid text tokens absorbing this background noise due to bidirectional attention, marked by abnormal white hotspots. Image (c) shows that after removing the EOS suffix, the model correctly concentrates its attention purely on the target, which is the actor's face.}
    \caption{Visualization of cross-modal attention showing suffix-mediated noise leakage. \textbf{(a)} Redundant \texttt{[EOS]} tokens attend to irrelevant backgrounds. \textbf{(b)} Valid tokens absorb this background noise due to bidirectional attention (white boxes show abnormal hotspots). \textbf{(c)} Removing the \texttt{[EOS]} suffix eliminates this noise, concentrating attention on the correct target (the actor's face). This shows that truncating padding improves both efficiency and visual reasoning.}
    \label{fig:noise_leakage}
    \vspace{-3mm}  
\end{figure}

Beyond efficiency, these semantically vacuous suffix tokens may also degrade prediction quality. As illustrated in Figure~\ref{fig:noise_leakage}, because padding tokens like \texttt{[EOS]} are weakly grounded, they tend to attend diffusely to task-irrelevant visual backgrounds (Figure~\ref{fig:noise_leakage}a). In a bidirectional attention paradigm, these suffix tokens act as weak relay nodes, causing background noise to leak back into the valid text prefix (Figure~\ref{fig:noise_leakage}b). By early truncating the redundant suffix, we effectively cut off this indirect contamination pathway, allowing the valid tokens to concentrate their cross-modal attention sharply on the correct semantic target (Figure~\ref{fig:noise_leakage}c). Therefore, removing the padding suffix is not merely a systems-level acceleration technique, but also a representation purification process that guards against bidirectional noise interference.

Existing acceleration methods for multimodal large models do not adequately resolve this form of text-side redundancy. In fact, recent acceleration efforts for Diffusion Multimodal Large Language Models (DMLLMs), such as D3ToM~\cite{chang2025d} and RedVTP~\cite{xu2025redvtp}, largely inherit visual token compression paradigms from autoregressive multimodal models, primarily focusing on token merging or pruning. However, these methods that heavily target visual tokens have inherent limitations. Aggressive visual compression may irreversibly remove fine-grained details that are crucial for complex vision-language reasoning. More importantly, they completely fail to address the redundant text suffix that remains on the output side under the fixed-window diffusion decoding paradigm. In short, existing multimodal acceleration techniques mainly attack visual redundancy at the input side, while leaving untouched the massive, repeated computation over semantically empty text suffixes. This raise the question: \textbf{can we improve generation efficiency by pre-defining a precise sequence length?}

\begin{figure*}[t]
    \centering
    \includegraphics[trim=0 0 0 0, clip, width=\textwidth]{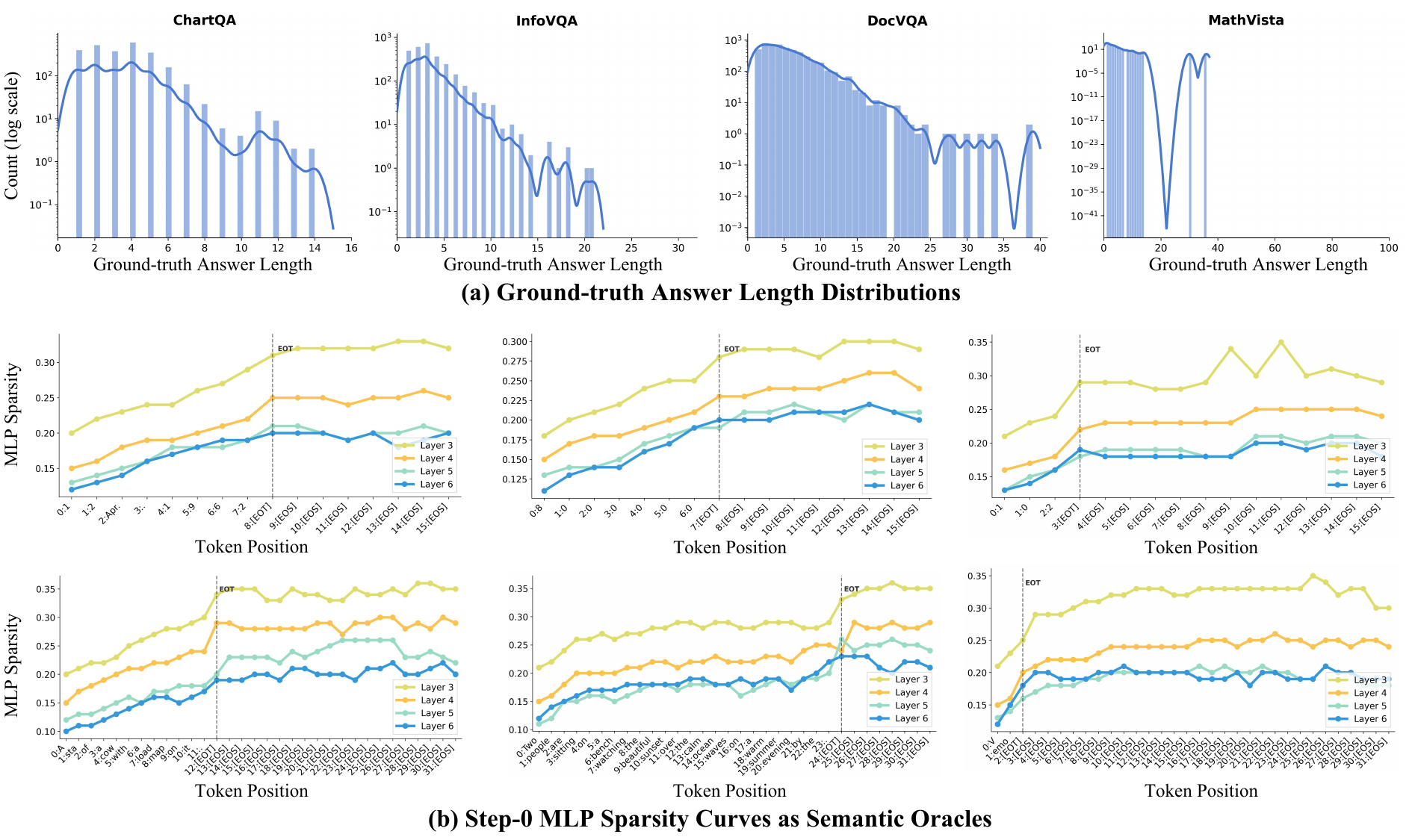}
    \Description{A two-part figure. Panel (a) displays four bar charts showing Ground-truth Answer Length Distributions across different VQA datasets, featuring right-skewed long tails which indicate a high volume of short answers. Panel (b) displays four line graphs showing Step-0 MLP Sparsity Curves across early model layers. The curves remain low for valid text tokens, spike sharply at the predicted end-of-text token forming a "Semantic Jump", and then flatten into a highly stable "Padding Plateau" for the remaining redundant tokens.}
    \vspace{-8mm}
    \caption{Motivation for dynamic semantic truncation. \textbf{(a) Ground-truth Answer Length Distributions} across four representative VQA datasets. The high variance and right-skewed long tails demonstrate that a statically fixed sequence window inevitably incurs massive padding waste for shorter answers. \textbf{(b) Step-0 MLP Sparsity Curves} across early layers (Layers 3--6). Rather than being indistinguishable, the sparsity curves reveal a sharp Semantic Jump exactly at the predicted \texttt{[EOT]} token, followed by a highly stable Padding Plateau across the redundant \texttt{[EOS]} suffix. This distinct early-layer phenomenon serves as a robust, training-free indicator for dynamic boundary detection.}
    \label{fig:motivation}
    \vspace{-3mm}  
\end{figure*}

The answer is yes, but the solution is non-trivial. A naive approach to mitigate this padding waste is to statically reduce the maximum sequence window, but as illustrated in Figure~\ref{fig:motivation}(a), ground-truth answer lengths exhibit high variance and right-skewed long-tail distributions across diverse multimodal tasks. The longest answer is often 10$\times$ to the shorter ones. Setting a static window small enough to save compute would blindly truncate valid reasoning chains, while a large safe window maintains the padding curse.

Faced with this dilemma, we ask a different question:\textit{ can a DMLLM infer, at the very beginning of denoising, where the semantically valid text ends so the redundant suffix can be dynamically removed once and for all?} Intuitively, semantically valid tokens encapsulate rich information, so their continuous MLP activation values deviate strongly from zero. Conversely, semantically vacuous padding tokens' activations naturally cluster around zero. 

Motivated by this difference, we systematically profile the text-side MLP activation sparsity at the initial denoising step (Step-0). Through empirical analysis, we discover that defining near-zero activations with an absolute threshold 
yields the most distinct boundary discrimination. As visualized in the layer-wise sparsity curves across early layers (Layers 3 to 6) in Figure~\ref{fig:motivation}(b), we uncover a striking and previously undocumented phenomenon. Long before explicit sequence generation completes, these early layers already expose a highly distinct boundary signal. The semantically valid prefix maintains relatively dense activations, whereas the predicted end region is marked by a sharp sparsity spike (the Semantic Jump), immediately followed by a prolonged and highly stable flat line (the Padding Plateau). This Step-0 indicator reveals that the model internally ``knows'' the true semantic boundary right from the start.


Based on this observation, we present Seer, an algorithm-system co-design framework for accelerating DMLLM inference through Step-0 semantic boundary detection and early suffix truncation. At the algorithmic level, Seer introduces an SNR-aware boundary detector. Rather than relying on brittle absolute thresholds, it captures the jump-to-plateau structure in early-layer MLP sparsity to estimate the semantic boundary robustly. This enables a one-shot macro-truncation strategy: once the boundary is identified at Step-0, the redundant suffix can be removed from all subsequent layers and denoising steps. In addition to reducing computation, we further observe that early suffix removal can help mitigate suffix-mediated noise leakage in bidirectional multimodal attention, which may improve reasoning quality on some visually complex examples.

Turning this semantic shortening into real latency and throughput gains, however, requires more than boundary detection alone. Under high-concurrency serving, naive per-sample truncation can easily undermine the practical benefit of shortening. When samples with different predicted boundaries are batched together, shorter sequences are often padded again to match the longest truncated sequence in the batch, which reintroduces substantial invalid computation. At the same time, if the effective sequence shape varies significantly across batches, optimized static execution paths become difficult to preserve. Consequently, the theoretical benefit of truncation does not automatically translate into reduced end-to-end latency or higher throughput.


To address this mismatch, Seer introduces a hybrid execution routing framework. Rather than padding all dynamically shortened sequences to a batch-level maximum, which risks severe padding waste when a few long sequences dominate the batch, Seer clusters sequences and quantitatively evaluates their expected Padding Waste Ratio (PWR). Based on this metric, sequence groups are dynamically dispatched across three specialized paths. Uniform groups with negligible waste are routed to a highly optimized Static-Graph path; groups with moderate length dispersion are directed to a Bucket-Varlen-Graph path that repacks sequences to eliminate padding while maintaining CUDA Graph compatibility; and highly skewed outliers or ultra-short prefixes are isolated into a lightweight Eager execution path. By mapping dynamic boundaries into these execution-friendly layouts, Seer translates theoretical FLOP reductions into consistent wall-clock acceleration under batched inference.

In summary, our main contributions are as follows:
\begin{itemize}
    \item We demonstrate that output sequence-level text redundancy creates significant overhead in multimodal diffusion, as repeated suffixes amplify unnecessary interactions with high-dimensional visual contexts and can inject suffix-driven visual noise.
    \item We are the first to identify a previously undocumented phenomenon in DMLLMs: Step-0 early-layer text-side MLP sparsity serves as a reliable semantic boundary signal for the valid text prefix.
    \item We introduce Seer, a training-free, one-shot macro-truncation framework that detects semantic boundaries using an SNR-aware sparsity criterion and eliminates redundant suffix computations throughout the rest of the denoising process.
    \item We design a systems-level execution framework that maintains the practical benefits of semantic shortening during batched serving by combining padding-waste-aware routing with device-resident latency shadowing.
\end{itemize}

\section{Related Work}
\label{sec:related_work}

\subsection{Visual Token Compression for Autoregressive MLLMs}
In autoregressive MLLMs \cite{liu2023visual, liu2024improved, chen2024internvl, wang2024qwen2, yao2024minicpm}, high-resolution images and multi-frame videos often produce excessively long visual token sequences, resulting in substantial self-attention overhead. To mitigate this input-side redundancy, early works primarily explored token merging \cite{bolya2022token}, heuristic pruning and selection \cite{chen2024image, rao2021dynamicvit, alvar2025divprune, dong2025mmtok}, or attention/sparsity-guided reduction \cite{yang2025visionzip, zhang2024sparsevlm}. Building upon these foundations, recent advances have introduced more adaptive compression criteria, including language-guided visual token pruning, hierarchical multi-stage reduction, distribution-aligned optimal transport, adaptive token skipping, search-based reduction, progressive compression, dynamic token early exit, and adaptive inference strategies \cite{ma2026apet, chen2026otprune, wu2026hidrop, baek2026agilepruner, sun2025lvpruning, zhang2025beyond, wen2025stop, dhouib2025pact, ma2025short, zhong2025aim, luo2025large, jiang2025kind, zeng2025skip, ye2025fit, he2024zipvl, zhao2025accelerating, ye2412atp, yang2025pvc, vasu2025fastvlm, wu2024accelerating, xu2025learning,wang2026zeus,jiang2025sada}. By leveraging clustering mechanisms, complex visual cues, or graph perspectives, these methods aim to preserve task-relevant semantics while discarding uninformative background features. Meanwhile, recent critical studies have re-examined these paradigms \cite{wen2025token, zou2025don}, showing that blindly prioritizing highly activated or highlighted tokens can compromise holistic visual context and motivating strategies that better preserve global structural integrity. Furthermore, output-side acceleration via speculative decoding has also been explored \cite{lin2025speculative}. However, while effective for autoregressive models, these methods are not designed to address the massive output-side padding redundancy arising in fixed-window diffusion-based multimodal generation.

\subsection{Acceleration for Diffusion Multimodal LLMs}
Accelerating DMLLMs naturally inherits optimization strategies from diffusion language models~\cite{nie2025large,ye2025dream}, which demonstrate faster-than-autoregressive potential~\cite{wang2025diffusion} and inspire comprehensive efficient inference frameworks~\cite{ma2025dinfer}. Existing methods mainly include caching intermediate states~\cite{wu2025fast, ma2025dkv, liu2025dllm, jiang2025d, song2508sparse, nguyen2025attention, huang2025mask}, speculative decoding~\cite{li2025diffuspec, cheng2025deer, liu2026dart}, parallel decoding via chunking~\cite{arriola2025block, cheng2025sdar, yu2505dimple, israel2025accelerating, bao2025learning}, and post-training quantization~\cite{zhang2025quant, xu2025dllmquant, lin2025quantization}. Extending to the multimodal setting, alongside general architectures such as Lumina-DiMOO~\cite{xin2025lumina}, recent works focus on accelerating visual inputs through dynamic token merging or pruning~\cite{chang2025d, xu2025redvtp, li2025comprehensive}. However, such lossy compression may discard fine-grained features crucial for complex reasoning, while sparse optimization mechanisms~\cite{li2025sparse} often require prohibitive fine-tuning costs. More fundamentally, these visual-centric methods fail to address the core bottleneck: the massive computational waste on text padding tokens. To resolve this issue, alternative variable-length strategies employ EOS-guided stopping~\cite{yang2025diffusion}, early-skipping and dual-boundary orchestration mechanisms~\cite{zhu2026dllm, wei2025orchestrating}, or dynamic per-step token eviction~\cite{liang2026focus, kim2025any, li2025beyond}. However, continuous token eviction shatters the static computation graphs (e.g., CUDA Graphs) required by modern inference engines, incurring substantial CPU launch overheads. In contrast, we propose a training-free, one-shot macro-truncation approach based on Step-0 text-side MLP sparsity, which resolves the padding curse with zero visual feature loss and perfect CUDA Graph compatibility.

\section{Methodology}
\label{sec:methodology}

\begin{figure*}[t]
    \centering
    \includegraphics[trim=0 0 0 0, clip, width=\textwidth]{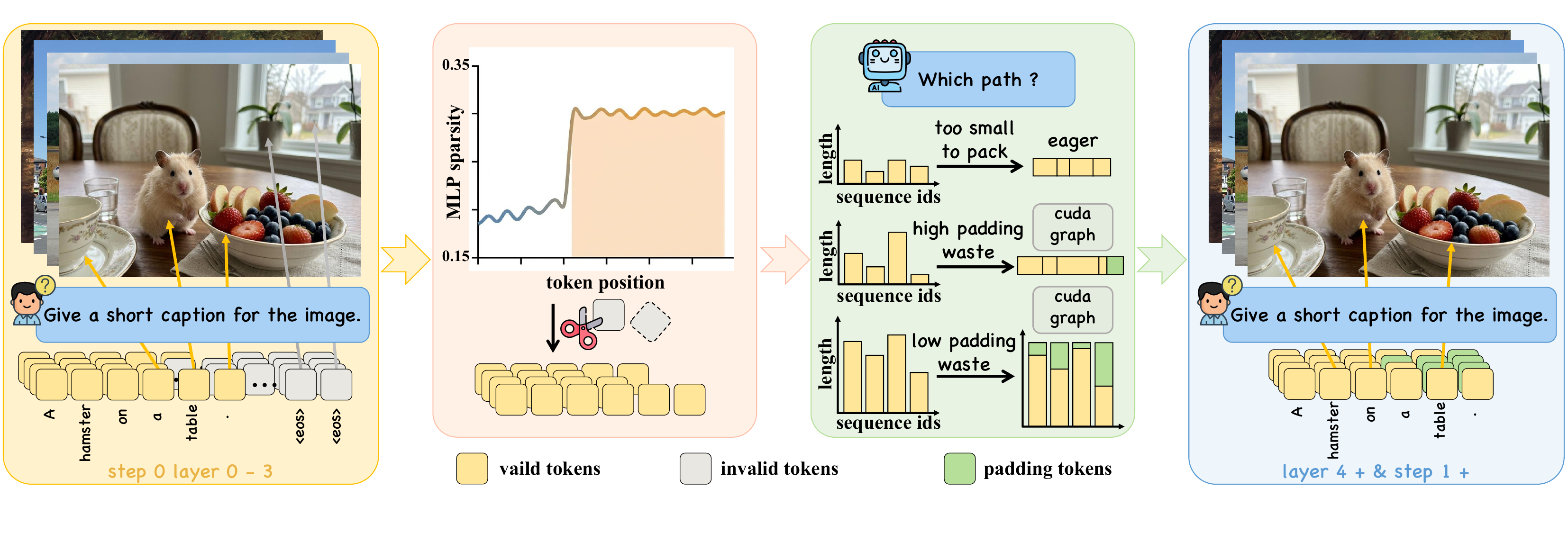}
    \Description{A flowchart illustrating the Seer framework's pipeline. On the left, an initial step processes a fixed-length sequence where redundant padding attends to image backgrounds. In the middle-left, Seer detects a semantic boundary using an MLP sparsity jump graph to truncate the suffix. In the middle-right, a hybrid router dynamically dispatches the truncated sequences into Eager, Varlen-Graph, or Static-Graph execution paths based on padding waste calculations. On the right, the compacted sequence accelerates inference and correctly focuses attention on the target visual object (a hamster).}
    \vspace{-8mm}
    \caption{Overview of the Seer framework. Left (Step 0, early layers): Redundant suffix tokens waste computation and diffusely attend to irrelevant visual backgrounds. Mid-Left: Seer identifies the exact semantic boundary via the MLP sparsity jump and strictly truncates the redundant suffix. Mid-Right: to preserve high-throughput serving, a hybrid router dynamically dispatches the truncated batch into execution-friendly paths (Eager, Varlen-Graph, or Static-Graph) based on padding waste. Right (Remaining steps \& layers): The compacted sequence accelerates inference while purifying cross-modal attention, focusing purely on the relevant visual target.}
    \label{fig:pipeline}
    \vspace{-3mm}  
\end{figure*}

In this section, we present Seer (Step-zero End Estimation and Reduction), an algorithm-system co-design framework to resolve the \textit{padding curse} in Diffusion Multimodal Large Language Models (DMLLMs). Figure~\ref{fig:pipeline} illustrates our overall pipeline. We first review the DMLLM background (Sec.~\ref{subsec:preliminary}) and reveal a pivotal text-side MLP sparsity phenomenon at the initial denoising step (Sec.~\ref{subsec:observation}). Based on this empirical indicator, we propose an SNR-guided semantic boundary detection algorithm (Sec.~\ref{subsec:snr_detection}). Finally, to elegantly translate this dynamic suffix truncation into high-throughput execution during high-concurrency serving, we introduce a padding-waste-aware static bucket lowering abstraction (Sec.~\ref{system}).

\subsection{Motivation: The Curse of Padding}
\label{subsec:preliminary}

Unlike Autoregressive (AR) models that generate tokens sequentially, Diffusion Multimodal Large Language Models (DMLLMs, e.g., LaViDa) frame text generation as a global, iterative denoising process over a discrete state space. Let $\mathbf{v}$ denote the visual context embeddings and $\mathbf{x}_t = [x_{t,1}, x_{t,2}, \dots, x_{t,L}]$ represent the text sequence of length $L$ at diffusion timestep $t \in \{1, \dots, T\}$.

During inference, the model initializes the sequence from a fully masked absorbing state $\mathbf{x}_T \sim p(\mathbf{x}_T)$ and progressively refines it into the clean data distribution $\mathbf{x}_0$. At each timestep $t$, the neural network computes the reverse transition probabilities parameterized by $\theta$:
\begin{equation}
\label{eq:denoising_step}
    p_\theta(\mathbf{x}_{t-1} | \mathbf{x}_t, \mathbf{v}) = \prod_{i=1}^{L} p_\theta(x_{t-1, i} | \mathbf{x}_t, \mathbf{v}).
\end{equation}
Equation~\ref{eq:denoising_step} reveals a key DMLLM property: the network unconditionally processes the entire sequence $\mathbf{x}_t$ of length $L$ at every denoising step to predict all token marginals simultaneously.

To enable batched execution, the $B$ sequences in a batch are padded to $L_{\max} = \max(L_1, \dots, L_B)$. The input tensor thus becomes $\mathbf{H} \in \mathbb{R}^{B \times L_{\max} \times d}$, where $d$ is the hidden dimension.

This static padding introduces a catastrophic computational burden. For a Transformer layer computing self-attention and MLP projections, the computational complexity per step is:
\begin{equation}
\label{eq:complexity}
    \text{FLOPs} \approx \mathcal{O}(L_{\max} \cdot d^2 + L_{\max}^2 \cdot d).
\end{equation}
Because the diffusion process operates globally over $T$ steps , this $\mathcal{O}(L_{\max}^2)$ attention complexity is multiplied by $T$, leading to severe performance degradation. We refer to this structural redundancy as the Curse of Padding. Even when the actual semantic output for sequence $i$ requires a fraction of tokens ($L_i \ll L_{\max}$), the system expends immense computational resources processing vacant \texttt{[EOS]} or padding tokens.

\subsection{Observation: Seeing the End at Step-0}
\label{subsec:observation}

To fundamentally break this computational bottleneck, we investigate whether the model internally ``knows'' the actual semantic boundary before explicitly generating \texttt{[EOT]} or \texttt{[EOS]} tokens.

Profiling the text-side MLP sparsity across early layers at the initial denoising step (Step-0), we consistently observe a highly distinctive phenomenon:

(1) The Semantic Jump: Information-rich prefix tokens exhibit dense activations (low sparsity), whereas the sparsity sharply spikes exactly at the predicted \texttt{[EOT]} or the first \texttt{[EOS]} padding token.

(2) The Padding Plateau: Following this jump, the sparsity of the remaining redundant suffix enters a prolonged, extremely stable plateau with minimal variance.

This Step-0 MLP sparsity acts as a global metric, enabling a prefix-preserving one-shot macro-truncation. By detecting this boundary at Step-0, the entire redundant padding suffix is permanently eliminated from all subsequent computation graphs.

\subsection{SNR-Aware Semantic Boundary Detection}
\label{subsec:snr_detection}

Hardcoded absolute sparsity thresholds are brittle because activation scales vary across tasks and prompts. We therefore formulate boundary detection as a Signal-to-Noise Ratio (SNR) test with a short local look-ahead.

Let $\mathcal{S} = [s_1, s_2, \dots, s_N]$ be the 1D sparsity array from Layer 3 at Step-0. At position $i$, we examine three consecutive positions, i.e., $\mathcal{W}_i=\{s_i, s_{i+1}, s_{i+2}\}$, against the preceding baseline $s_{i-1}$. This short look-ahead is kept fixed throughout and serves as the minimal local context to capture a sparsity jump followed by a stable plateau. The jump magnitude (Signal) and intra-window fluctuation (Noise) are defined as:
\begin{equation}
    \Delta_{jump}^{(i)} = \min(\mathcal{W}_i) - s_{i-1},
\end{equation}
\begin{equation}
    \Delta_{noise}^{(i)} = \max(\mathcal{W}_i) - \min(\mathcal{W}_i).
\end{equation}

A valid semantic boundary $i^*$ is detected only if:
\begin{itemize}
    \item \textbf{Significant Signal:} $\Delta_{jump}^{(i)} \ge \tau_{jump}$,
    \item \textbf{Stable Plateau (SNR Check):} $\Delta_{noise}^{(i)} \le \gamma \cdot \Delta_{jump}^{(i)}$.
\end{itemize}

The relative SNR constraint ($\gamma$) adapts to different absolute sparsity scales. To avoid trimming the last valid token, we keep one extra token and truncate at $L^* = i^* + 1$.

\subsection{Hybrid Execution Routing for Dynamic Boundaries}
\label{system}
Identifying a semantic boundary at Step-0 is only the first half of the problem. The more fundamental challenge is how to \emph{preserve execution regularity} after dynamic suffix truncation, especially under high-concurrency serving. A naive realization would directly slice each sequence according to its detected boundary. However, such per-sample dynamic truncation destroys batch uniformity, fragments the execution shape space, and prevents the inference engine from exploiting highly optimized static serving primitives (e.g., CUDA Graphs). As a result, theoretical FLOP reduction does not automatically translate into wall-clock acceleration.

To address this, Seer maps the dynamically shortened batch into a nearby static bucket. Formally, suppose a batch contains $B$ sequences with detected boundaries $\{L^*_1, L^*_2, \dots, L^*_B\}$. A straightforward approach computes a unified batch-level boundary:
\begin{equation}
\label{eq:batch_max}
L^*_{\max} = \max(L^*_1, L^*_2, \dots, L^*_B).
\end{equation}
This boundary is then mapped into a predefined set of static micro-buckets, $\mathcal{B} = \{8, 16, 32, \ldots, N_{\max}\}$.

While static bucket mapping successfully preserves execution regularity, naive batch-level max-pooling suffers from the \emph{straggler effect}. If a single extremely long sequence co-exists with numerous short sequences in the same batch, coercing all sequences to the maximum static bucket introduces massive padding waste, nullifying the FLOP savings gained from Step-0 truncation.

To quantitatively govern this routing and isolate stragglers, we define the \textit{Padding Waste Ratio (PWR)} for a sequence group $G$ mapped to a static bucket of size $B_G$ as:
\begin{equation}
\label{eq:pwr}
\rho = 1 - \frac{\sum_{i \in G} L_i^*}{|G| \times B_G}.
\end{equation}

Based on $\rho$ and the absolute sequence length, Seer implements a three-tier hybrid execution routing policy. Sequences are first clustered into coarse buckets, and each bucket is dynamically routed via the following execution paths:

\noindent \ding{182} \textbf{Static-Graph Path:} Triggered when lengths are concentrated and padding waste is negligible ($\rho \le \tau_{pad}$, e.g., $\tau_{pad}=0.2$). Here, variable-length repacking costs more than it saves. Seer assigns the group to a unified static bucket $B_G$ and leverages pre-captured CUDA Graphs for regular execution.

\noindent \ding{183} \textbf{Bucket-Varlen-Graph Path:} Triggered when the bucket exhibits moderate dispersion ($\rho > \tau_{\text{pad}}$), where static mapping incurs unacceptable padding waste. Seer regroups sequences and performs variable-length (varlen) packed execution. Instead of padding individual sequences, the \emph{total packed capacity} $\sum_{i \in G} L_i^*$ is mapped to a predefined 1D graph shape, preserving graph compatibility while substantially eliminating padding.

\noindent \ding{184} \textbf{Bucket-Varlen-Eager Path:} Triggered for highly skewed distributions, e.g., ultra-short prefixes ($B_G \le 4$ in our implementation) or isolated stragglers ($|G| \to 1$). Graph capture is inefficient here. Seer isolates these sequences and routes them to a lightweight eager path, with the dispatch overhead offset by padding elimination.

\subsection{Device-Resident Latency Shadowing}
A critical pitfall in dynamic inference systems is the CPU-GPU synchronization barrier. If dynamic semantic boundaries evaluated at Step-0 require CPU intervention to slice tensors and dispatch modified shapes, the resulting PCIe data transfer and sync delays will outright negate any acceleration benefits.

To decouple semantic adaptivity from host-side control flow, Seer introduces a zero-host-intervention In-GPU Compaction mechanism. We implemented a specialized fused Triton kernel that operates directly within the device memory (VRAM). Once the boundaries $L^*$ are evaluated, this kernel asynchronously performs prefix slicing, sequence repacking (for varlen paths), bucket alignment, padding, and attention mask rewriting in a single pass.

Crucially, to orchestrate the subsequent execution paths without blocking the CPU, the kernel utilizes an asynchronous pinned memory (page-locked) flag. It writes a lightweight control signal (e.g., the target bucket ID, chosen execution path, or packed length) back to the host. The CPU simply polls this signal without waiting for the massive data tensors, maintaining a fully device-resident dataflow. This abstraction ensures that the serving stack only observes a highly optimized, static-compatible execution queue, turning theoretical FLOP reductions into strict wall-clock acceleration.

\subsection{Accuracy Impact Analysis}
\label{subsec:noise_leakage}

While the preceding algorithm-system co-design focuses entirely on eliminating redundant computation for inference acceleration, we empirically observe that our macro-truncation mechanism can unexpectedly improve multimodal reasoning accuracy on visually complex datasets. In this subsection, we provide a theoretical perspective on this phenomenon. Our analysis suggests that, in bidirectional diffusion MLLMs, redundant suffix tokens may serve as leakage channels that relay task-irrelevant visual information back into the semantic text prefix. By removing these suffixes early, Seer can suppress this indirect contamination pathway in addition to reducing FLOPs.

Let the full multimodal sequence be $\mathcal{X} = \mathcal{V} \cup \mathcal{P} \cup \mathcal{S}$, where $\mathcal{V}$ denotes the visual tokens, $\mathcal{P}$ denotes the valid semantic text prefix, and $\mathcal{S}$ denotes the redundant text suffix consisting of padding and trailing \texttt{[EOS]} tokens. For a valid prefix token $p \in \mathcal{P}$, its hidden state at layer $l+1$ is updated through bidirectional self-attention over the entire sequence:
\begin{equation}
\label{eq:full_state_update}
\mathbf{h}_p^{(l+1)} = \sum_{v \in \mathcal{V}} A_{p\to v}^{(l)} \mathbf{v}_v^{(l)} + \sum_{p' \in \mathcal{P}} A_{p\to p'}^{(l)} \mathbf{v}_{p'}^{(l)} + \sum_{s \in \mathcal{S}} A_{p\to s}^{(l)} \mathbf{v}_s^{(l)}.
\end{equation}

Here, the attention weights are given by:
\begin{equation}
\label{eq:full_attention_weight}
A_{p\to j}^{(l)} = \frac{\exp(e_{p,j}^{(l)})}{\sum_{u \in \mathcal{X}} \exp(e_{p,u}^{(l)})},
\end{equation}
where the attention logit is
\begin{equation}
\label{eq:attention_logit}
e_{p,j}^{(l)} = \frac{{\mathbf{q}_p^{(l)}}^\top \mathbf{k}_j^{(l)}}{\sqrt{d}}.
\end{equation}

The key observation is that the suffix term in Eq.~\ref{eq:full_state_update} is not necessarily harmless under bidirectional attention. Since the suffix tokens in $\mathcal{S}$ are semantically weakly grounded, their visual attention can become diffuse and may absorb information from task-irrelevant visual regions. For analytical clarity, we decompose the value representation of a suffix token $s \in \mathcal{S}$ as
\begin{equation}
\label{eq:value_decomposition}
\mathbf{v}_s^{(l)} = \mathbf{c}_s^{(l)} + \mathbf{n}_s^{(l)},
\end{equation}
where $\mathbf{c}_s^{(l)}$ denotes the task-consistent component and $\mathbf{n}_s^{(l)}$ denotes the task-irrelevant visual component carried by that suffix token. Substituting Eq.~\ref{eq:value_decomposition} into the suffix term of Eq.~\ref{eq:full_state_update} yields
\begin{equation}
\label{eq:suffix_split}
\sum_{s \in \mathcal{S}} A_{p\to s}^{(l)} \mathbf{v}_s^{(l)}
=
\sum_{s \in \mathcal{S}} A_{p\to s}^{(l)} \mathbf{c}_s^{(l)}
+
\sum_{s \in \mathcal{S}} A_{p\to s}^{(l)} \mathbf{n}_s^{(l)}.
\end{equation}

The second term in Eq.~\ref{eq:suffix_split} captures an indirect contamination effect. We denote it by
\begin{equation}
\label{eq:noise_leakage_term}
\Delta \mathbf{h}_{p,\mathcal{S}}^{(l+1)} = \sum_{s \in \mathcal{S}} A_{p\to s}^{(l)} \mathbf{n}_s^{(l)}.
\end{equation}
This term reflects suffix-mediated noise leakage: even if a valid prefix token $p$ does not directly attend to irrelevant visual regions, it may still absorb such information through suffix tokens. In other words, redundant suffixes can act as semantically weak relay nodes in bidirectional attention.

After Seer detects the semantic boundary at Step-0, the redundant suffix $\mathcal{S}$ is removed from subsequent computation. The attention update for $p$ is then restricted to the retained multimodal context $\mathcal{V} \cup \mathcal{P}$:
\begin{equation}
\label{eq:truncated_state_update}
\tilde{\mathbf{h}}_p^{(l+1)} = \sum_{v \in \mathcal{V}} \tilde{A}_{p\to v}^{(l)} \mathbf{v}_v^{(l)} + \sum_{p' \in \mathcal{P}} \tilde{A}_{p\to p'}^{(l)} \mathbf{v}_{p'}^{(l)}.
\end{equation}
The corresponding normalized attention weights become
\begin{equation}
\label{eq:truncated_attention_weight}
\tilde{A}_{p\to j}^{(l)} = \frac{\exp(e_{p,j}^{(l)})}{\sum_{u \in \mathcal{V} \cup \mathcal{P}} \exp(e_{p,u}^{(l)})}, \quad j \in \mathcal{V} \cup \mathcal{P}.
\end{equation}

Comparing Eq.~\ref{eq:truncated_state_update} with Eq.~\ref{eq:full_state_update}, the suffix-mediated relay term in Eq.~\ref{eq:noise_leakage_term} is removed entirely. Under a fixed-score view of Eq.~\ref{eq:full_attention_weight} and Eq.~\ref{eq:truncated_attention_weight}, removing $\mathcal{S}$ also reallocates the normalization mass from the discarded suffix toward the retained valid context. Formally,
\begin{equation}
\label{eq:mass_conservation}
\sum_{j \in \mathcal{V} \cup \mathcal{P}} \tilde{A}_{p\to j}^{(l)} = 1, \qquad \sum_{j \in \mathcal{X}} A_{p\to j}^{(l)} = 1.
\end{equation}
Thus, truncation not only reduces the number of processed tokens, but also removes one source of indirect contamination while concentrating attention on semantically relevant subsets.

This perspective provides a plausible explanation for why Seer can occasionally improve accuracy instead of merely preserving it. If the expected magnitude of the suffix-mediated noise term is non-negligible, i.e.,
\begin{equation}
\label{eq:nonzero_noise}
\mathbb{E}\left[\left\|\Delta \mathbf{h}_{p,\mathcal{S}}^{(l+1)}\right\|_2\right] > 0,
\end{equation}
then removing $\mathcal{S}$ suppresses a meaningful contamination pathway. In this case, macro-truncation acts not only as a computational reduction mechanism, but also as a form of \emph{representation purification}, which can improve robustness and reasoning accuracy in visually noisy environments.

\begin{table*}[!ht]
\centering
\caption{Comprehensive benchmark results on LaViDa-LLaDA and MMaDA architectures. Benchmark abbreviations: MU (MMMU), MB (MMBench), IV (InfoVQA), CQ (ChartQA), SQ (SQA), DV (DocVQA), GQ (GQA), MV (MathVista).}
\vspace{-3mm}
\label{tab-comprehensive_exp}
\resizebox{\textwidth}{!}{
\begin{tabular}{l ccccccccc ccccccccc}
\toprule
\raisebox{-1.5ex}[0pt][0pt]{Method} & \multicolumn{9}{c}{\textbf{LaViDa-LLaDA}} & \multicolumn{9}{c}{\textbf{MMaDA}} \\
\cmidrule(lr){2-10} \cmidrule(lr){11-19}
 & MME & MU & MB & IV & CQ & SQ & DV & GQ & MV & MME & MU & MB & IV & CQ & SQ & DV & GQ & MV \\
\midrule
\midrule
Baseline & \textbf{1705.29} & \textbf{44.00} & \textbf{75.76} & 38.11 & \textbf{60.60} & 72.38 & 63.52 & 56.70 & \textbf{45.70} & \underline{1306.72} & \textbf{31.44} & \textbf{39.72} & 15.11 & 9.92 & 56.82 & \textbf{10.30} & \textbf{50.79} & \underline{58.00} \\
Throughput $\uparrow$ & 0.53 & 0.60 & 0.12 & 0.81 & 1.50 & 0.77 & 1.16 & 0.64 & 0.16 & 0.29 & 0.14 & 0.28 & 0.13 & 1.37 & 0.28 & 0.15 & 0.35 & 0.59 \\
Latency $\downarrow$ & 1.87 & 1.88 & 8.66 & 3.32 & 1.95 & 1.30 & 3.68 & 1.95 & 8.79 & 3.48 & 7.72 & 3.58 & 14.94 & 1.70 & 3.54 & 15.20 & 3.41 & 11.22 \\
\midrule
+ D3ToM & 1577.04 & 42.67 & 70.45 & 25.58 & 23.00 & 71.79 & 24.49 & 52.12 & 40.00 & 1269.41 & 31.11 & 25.08 & 14.76 & 9.16 & 56.82 & 6.73 & 34.87 & 56.80 \\
Throughput $\uparrow$ & 0.85 & 0.90 & 0.19 & 1.37 & 2.20 & 1.01 & 1.72 & 1.07 & 0.20 & 0.59 & 0.25 & 0.56 & 0.24 & 2.12 & 0.54 & 0.23 & 0.66 & 1.07 \\
Latency $\downarrow$ & 1.16 & 1.24 & 5.20 & 1.87 & 1.18 & 0.99 & 1.97 & 1.17 & 5.08 & 1.70 & 4.00 & 1.79 & 7.65 & 0.89 & 1.84 & 7.56 & 1.68 & 5.91 \\
\midrule
+ RedVTP & 1510.48 & 42.22 & 70.45 & 24.25 & 25.40 & 71.15 & 27.78 & 49.93 & 39.00 & 1280.91 & 30.78 & 25.39 & 14.35 & \underline{10.00} & 57.56 & 6.70 & 36.03 & 57.40 \\
Throughput $\uparrow$ & 0.88 & 0.90 & 0.21 & 1.38 & 2.22 & 1.04 & 1.83 & 1.10 & 0.28 & 0.48 & 0.21 & 0.45 & 0.21 & 1.76 & 0.46 & 0.19 & 0.53 & 0.95 \\
Latency $\downarrow$ & 1.13 & 1.24 & 4.78 & 1.75 & 1.14 & 0.96 & 1.83 & 1.12 & 4.75 & 2.07 & 4.78 & 2.20 & 8.75 & 1.09 & 2.17 & 8.89 & 2.09 & 6.62 \\
\midrule
+ VisionZip & 1438.05 & 41.89 & 72.73 & 23.87 & 33.20 & 70.45 & 32.20 & 49.78 & 38.20 & 879.05 & 29.78 & 23.84 & \textbf{15.32} & 9.64 & \textbf{58.95} & 6.37 & 34.39 & \textbf{58.30} \\
Throughput $\uparrow$ & 0.72 & 0.81 & 0.16 & 1.11 & 2.00 & 0.88 & 1.41 & 0.92 & \underline{0.44} & 1.17 & 0.51 & 1.13 & 0.47 & \underline{4.58} & 1.08 & 0.45 & 1.35 & \underline{1.74} \\
Latency $\downarrow$ & 1.39 & 1.36 & 6.18 & 2.27 & 1.41 & 1.14 & 2.49 & 1.39 & \underline{3.05} & 0.85 & 2.12 & 0.88 & 3.86 & \underline{0.41} & 0.93 & 3.79 & 0.82 & 3.26 \\
\midrule
+ MMTok & 1600.66 & 43.11 & 71.97 & 29.78 & 42.20 & \textbf{72.78} & 29.78 & 53.27 & 42.50 & 869.26 & 31.22 & 22.37 & 14.91 & 9.52 & \textbf{58.95} & 6.81 & 34.40 & 56.90 \\
Throughput $\uparrow$ & 0.86 & 0.92 & 0.19 & 1.38 & 2.42 & 0.95 & 1.38 & 1.09 & 0.28 & \underline{1.20} & \underline{0.53} & \underline{1.16} & \underline{0.50} & \textbf{4.84} & \underline{1.22} & \underline{0.46} & 1.28 & 1.72 \\
Latency $\downarrow$ & 1.17 & 1.22 & 5.20 & 1.87 & 1.16 & 1.06 & 1.87 & 1.15 & 5.00 & \underline{0.83} & 2.06 & \underline{0.86} & \underline{3.69} & \textbf{0.40} & 0.89 & 3.74 & 0.86 & \underline{3.04} \\
\midrule
+ DivPrune & 1333.60 & 41.33 & 62.88 & 22.70 & 21.80 & 71.15 & 38.78 & 46.14 & 38.40 & 860.21 & 29.89 & 21.90 & 14.79 & 9.56 & \underline{58.60} & 6.30 & 49.10 & 57.10 \\
Throughput $\uparrow$ & \underline{1.09} & \underline{1.13} & \underline{0.24} & \underline{1.67} & \underline{2.59} & \underline{1.11} & \underline{1.88} & \textbf{1.37} & 0.40 & 1.13 & 0.52 & \underline{1.16} & 0.49 & \underline{4.58} & 1.10 & 0.45 & \underline{1.38} & 1.48 \\
Latency $\downarrow$ & \underline{0.92} & \underline{0.99} & \underline{4.11} & \underline{1.40} & \textbf{0.91} & \underline{0.90} & \underline{1.90} & \textbf{0.88} & 3.28 & 0.89 & \underline{2.04} & \underline{0.86} & 3.71 & \textbf{0.40} & 0.91 & \underline{3.70} & \underline{0.80} & 3.06 \\
\midrule
+ SparseVLMs & 1217.15 & 40.67 & 59.85 & 23.65 & 18.20 & 70.20 & 14.20 & 44.20 & 39.10 & 866.13 & 29.89 & 24.23 & {15.25} & 9.36 & 57.31 & 6.48 & 34.23 & 56.80 \\
Throughput $\uparrow$ & 0.89 & 0.97 & 0.19 & 1.34 & 2.21 & 0.99 & 1.69 & 1.11 & 0.27 & 1.11 & 0.44 & 1.01 & 0.37 & 4.35 & 0.96 & 0.35 & 1.24 & 1.73 \\
Latency $\downarrow$ & 1.12 & 1.16 & 5.30 & 1.77 & 1.11 & 1.01 & 1.76 & 1.09 & 4.91 & 0.90 & 2.48 & 0.99 & 4.94 & 0.44 & 1.04 & 4.88 & 0.90 & 3.70 \\
\midrule
+ Seer (ours) & \underline{1697.94} & \underline{43.89} & \textbf{75.76} & \textbf{38.40} & \underline{57.80} & \underline{72.73} & \textbf{63.66} & \textbf{56.75} & \underline{45.10} & \textbf{1315.69} & \underline{31.33} & \underline{39.16} & \underline{15.29} & \textbf{10.12} & 57.16 & \underline{9.26} & \underline{49.45} & 55.80 \\
Throughput $\uparrow$ & \textbf{1.11} & \textbf{1.39} & \textbf{0.93} & \textbf{2.76} & \textbf{2.95} & \textbf{1.58} & \textbf{4.40} & \underline{1.36} & \textbf{1.68} & \textbf{3.39} & \textbf{2.33} & \textbf{3.17} & \textbf{4.02} & 3.95 & \textbf{2.35} & \textbf{4.48} & \textbf{4.46} & \textbf{3.39} \\
Latency $\downarrow$ & \textbf{0.90} & \textbf{0.81} & \textbf{1.08} & \textbf{0.99} & \underline{0.98} & \textbf{0.65} & \textbf{0.97} & \underline{0.92} & \textbf{0.75} & \textbf{0.30} & \textbf{0.46} & \textbf{0.32} & \textbf{0.49} & 0.59 & \textbf{0.43} & \textbf{0.50} & \textbf{0.27} & \textbf{0.39} \\
\bottomrule
\end{tabular}
}
\end{table*}

\section{Experiments}

\subsection{Evaluated Models, Baselines, and Setup}
To comprehensively evaluate the effectiveness and generalization of our proposed acceleration framework, we conduct experiments on three representative multi-modal models, specifically Mmada~\cite{yang2025mmada} and two variants from the LaViDa family~\cite{li2025lavida}, namely LaViDa-LLaDA and LaViDa-Dream. Furthermore, to demonstrate the superiority of our approach in achieving an optimal trade-off between inference efficiency and generation quality, we compare it against state-of-the-art acceleration strategies tailored for diffusion-based multi-modal large language models, namely D3TOM~\cite{chang2025d} and RedVTP~\cite{xu2025redvtp}, as well as highly influential visual token compression and pruning techniques originally designed for autoregressive multi-modal large language models but migrated to diffusion-based architectures in our implementation, including VisionZip~\cite{yang2025visionzip}, MMTok~\cite{dong2025mmtok}, DivPrune~\cite{alvar2025divprune}, and SparseVLM~\cite{zhang2024sparsevlm}. Detailed hardware specifications, implementation details, additional results on LaViDa-Dream, and further supplementary experiments and ablation studies are provided in the appendix.

\subsection{Evaluation Benchmarks and Metrics}
To comprehensively assess the multimodal understanding and reasoning capabilities of our framework, we evaluate it on nine widely used benchmarks: MME~\cite{fu2023mme}, MMMU~\cite{yue2024mmmu}, MMBench~\cite{liu2024mmbench}, InfoVQA~\cite{mathew2022infographicvqa}, ChartQA~\cite{masry2022chartqa}, ScienceQA~\cite{lu2022learn}, DocVQA~\cite{mathew2021docvqa}, GQA~\cite{hudson2019gqa}, and MathVista~\cite{lu2023mathvista}. To ensure fair and reproducible comparisons, we use the official implementation of the \texttt{lmms-eval}~\cite{zhang2025lmms} framework for all evaluations. We report task accuracy, end-to-end throughput (tokens/s), and end-to-end latency (s), with the latter two measured over the full serving pipeline.

\subsection{Main Results}
As shown in Table~\ref{tab-comprehensive_exp}, Seer achieves a favorable efficiency--quality trade-off on both LaViDa-LLaDA and MMaDA. Compared with the baseline, Seer consistently improves end-to-end throughput and reduces latency while maintaining competitive multimodal reasoning performance. Specifically, when evaluating LaViDa-LLaDA on MathVista, throughput increases from 0.16 to 1.68 tokens/s (a 10.5$\times$ gain) with only a 1.31\% relative decrease in accuracy; additionally, when evaluating MMaDA on InfoVQA, throughput increases from 0.13 to 4.02 tokens/s (a 30.9$\times$ gain), while accuracy improves from 15.11 to 15.29. In contrast, other baseline acceleration methods relying on visual token compression often suffer from significant accuracy drops when increasing speed. More importantly, under most evaluation settings in Table~\ref{tab-comprehensive_exp}, these existing baselines are inferior to Seer in both throughput and task accuracy. This highlights the advantage of Seer operating solely on redundant text suffixes, allowing it to preserve performance more steadily. Overall, these results demonstrate that text-side suffix truncation delivers substantial efficiency gains with limited impact on generation quality, and the performance improvements observed on certain visually complex tasks align with the analysis in Sec.~\ref{subsec:noise_leakage}.

\subsection{Effectiveness of System-Level Optimizations}
\begin{figure}
    \centering
    
    \includegraphics[trim=0 0 0 0, clip, width=\linewidth]{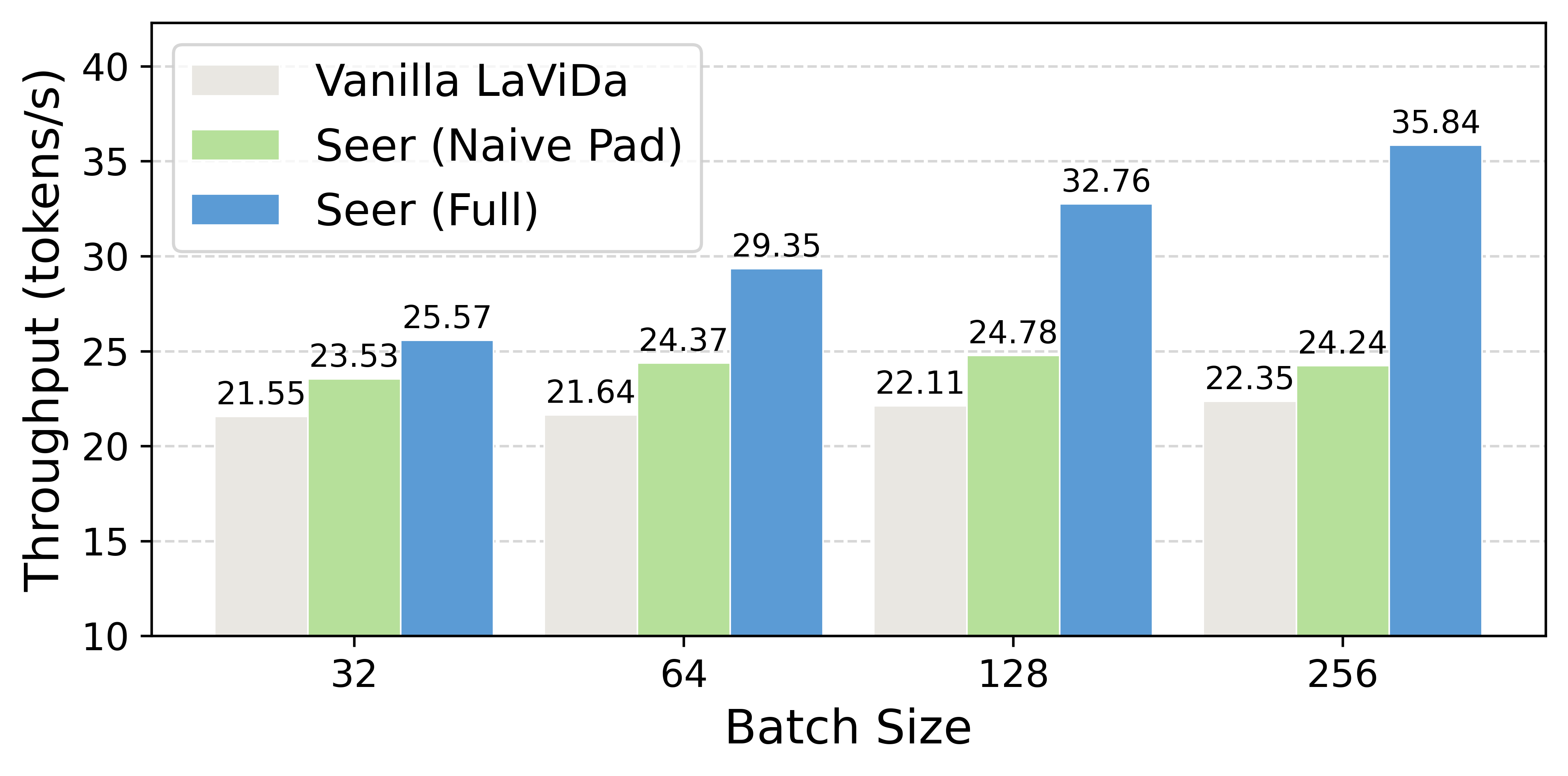}
    \Description{A grouped bar chart comparing inference throughput in tokens per second across four batch sizes (32, 64, 128, and 256). For each batch size, three execution strategies are compared: Vanilla LaViDa, Seer Naive Pad, and Seer Full. Seer Full consistently achieves the highest throughput, reaching 35.84 tokens per second at batch size 256, significantly outperforming the other two methods.}
    \vspace{-6mm}
    \caption{\textbf{Effectiveness of system-level optimizations.} Impact of different execution strategies on inference throughput. While applying algorithmic truncation alone (Seer Naive Pad) yields marginal gains over Vanilla LaViDa due to padding waste, our complete framework (Seer Full) translates theoretical reductions into substantial throughput improvements, especially at larger batch sizes.}
    \label{fig:system_opt}
    \vspace{-4mm}  
\end{figure}

We evaluate how execution strategies affect inference throughput across batch sizes. As Figure~\ref{fig:system_opt} shows, throughput for both Vanilla LaViDa and \textit{Seer Naive Pad} stagnates at larger batches due to severe padding waste from sequence-length variance. In contrast, \textit{Seer Full} resolves this bottleneck by combining padding-waste-aware hybrid routing with device-resident execution lowering. By dynamically packing sequences while preserving execution regularity, it translates theoretical reductions into substantial wall-clock acceleration, reaching 35.84 tokens/s at batch size 256 (a $1.6\times$ baseline speedup).

\begin{table}[!ht]
\centering
\caption{Joint impact of Absolute Jump Threshold ($\tau_{jump}$) and Tolerance Ratio ($\gamma$) on generation accuracy and inference throughput (Evaluated on ChartQA).}
\label{tab-hyperparams}
\resizebox{\columnwidth}{!}{%
\begin{tabular}{c cccccc}
\toprule
\textbf{Tol.} & \multicolumn{6}{c}{\textbf{Abs. Jump Threshold ($\tau_{jump}$)}} \\
\cmidrule(lr){2-7}
\textbf{($\gamma$)} & 0.01 & 0.02 & 0.03 & 0.04 & 0.05 & 0.06 \\
\midrule
0.2 & 60.6/1.62 & 60.6/1.55 & 60.4/1.48 & 60.2/1.50 & 60.6/1.53 & 60.6/1.48 \\
0.3 & 59.8/2.15 & 60.6/1.59 & 60.6/1.55 & 60.6/1.45 & 60.6/1.51 & 60.6/1.47 \\
0.4 & 54.2/3.11 & 59.5/2.15 & 60.4/1.67 & 59.3/1.69 & 60.3/1.57 & 59.6/1.54 \\
0.5 & 48.4/4.18 & 55.6/3.41 & 58.4/2.43 & 60.1/1.81 & 60.6/1.55 & 60.6/1.45 \\
0.6 & 35.8/5.15 & 51.4/3.80 & \textbf{57.8/2.93} & 59.8/2.35 & 60.5/1.60 & 60.6/1.58 \\
0.7 & 18.5/6.05 & 42.6/4.88 & 54.5/3.51 & 58.3/2.64 & 60.1/2.11 & 60.6/1.54 \\
0.8 & 8.4/6.67  & 30.2/5.62 & 46.8/4.35 & 53.4/3.36 & 55.1/3.13 & 60.4/1.95 \\
0.9 & 4.2/7.15  & 16.5/6.37 & 34.5/5.18 & 48.6/4.25 & 52.2/3.50 & 56.1/2.82 \\
1.0 & 2.1/7.49  & 8.6/6.88  & 20.5/5.85 & 38.2/4.80 & 51.4/3.93 & 55.4/3.12 \\
\bottomrule
\end{tabular}%
} 
\end{table}

\subsection{Ablation Studies}

\textbf{Hyperparameters with Accuracy-Throughput Trade-Off.} Table~\ref{tab-hyperparams} demonstrates how the absolute jump threshold ($\tau_{jump}$) and tolerance ratio ($\gamma$) govern our SNR-based detector's sensitivity. Rather than requiring heuristic tuning, they provide an explicit interface to navigate the quality-throughput Pareto frontier. Smaller $\tau_{jump}$ or larger $\gamma$ prioritizes aggressive truncation for maximal speed (e.g., 7.45 tokens/s) in latency-sensitive applications. Conversely, larger $\tau_{jump}$ or smaller $\gamma$ ensures strict boundary preservation, maintaining the 60.6 baseline accuracy for quality-critical tasks.

\begin{figure}[h]
    \centering
    \includegraphics[width=0.8\linewidth]{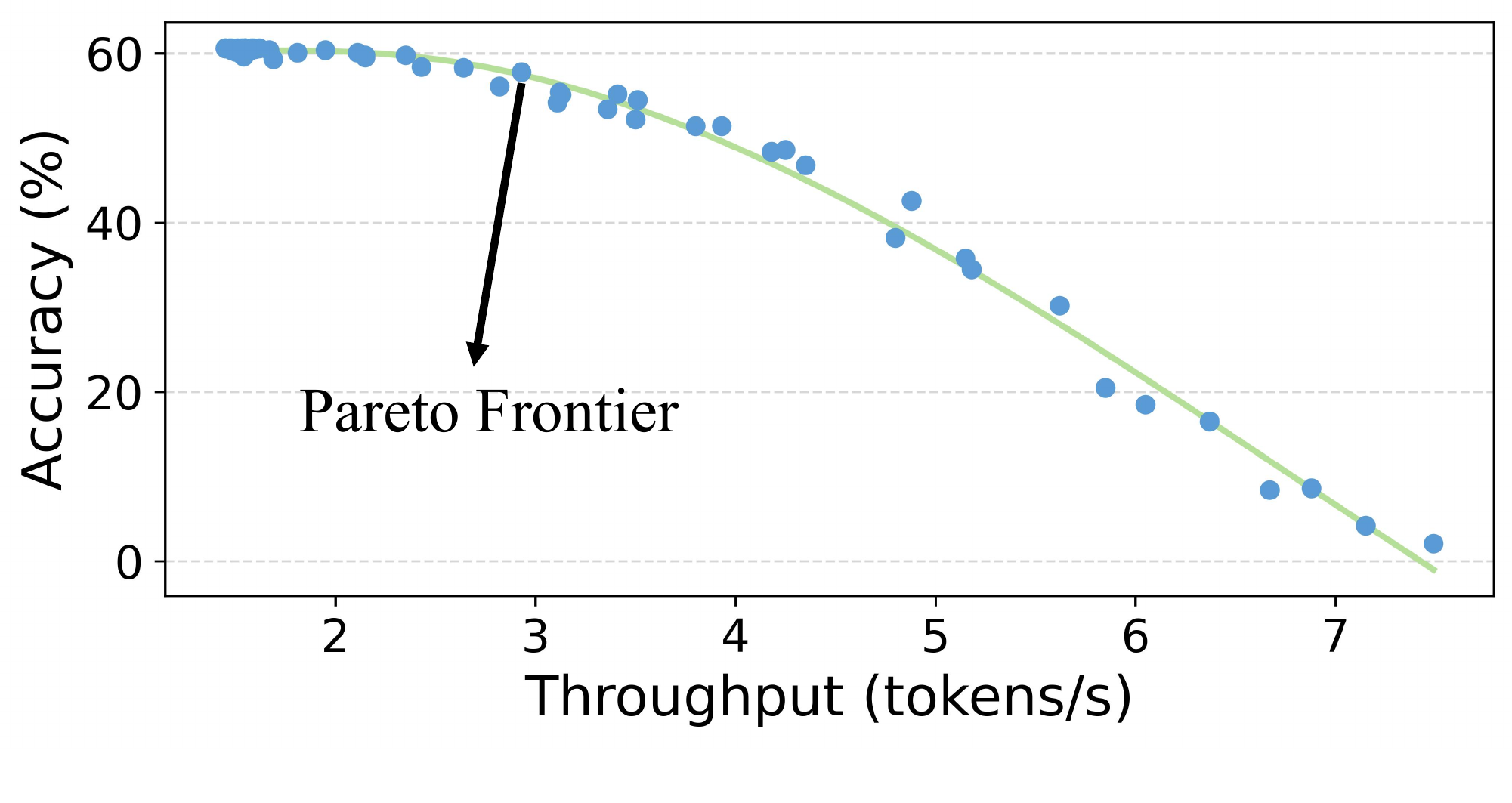} 
    \Description{A scatter plot depicting a downward-sloping Pareto frontier curve that maps generation accuracy percentage on the y-axis against inference throughput on the x-axis. An arrow points to a specific data point at the 'knee' of the curve, representing the optimal tradeoff balance between high accuracy and fast throughput.}
    \vspace{-6mm}
    \caption{Pareto frontier of generation accuracy versus inference throughput based on Table~\ref{tab-hyperparams}. The default configuration ($\tau_{jump}=0.03, \gamma=0.6$) is located at the optimal ``knee'' point.}
    \label{fig:pareto_curve}
    \vspace{-3mm}
\end{figure}

\textbf{Formulation of the Pareto Optimal Configuration.} To justify our default parameters ($\tau_{jump} = 0.03, \gamma = 0.6$) in Table~\ref{tab-comprehensive_exp} and avoid arbitrary tuning, we formalize this trade-off as a multi-objective optimization problem. Let $\hat{\mathcal{A}}(\tau, \gamma)$ and $\hat{\mathcal{T}}(\tau, \gamma)$ denote baseline-normalized accuracy and throughput, with joint utility $\mathcal{U}$ defined as:
\begin{equation}
    \mathcal{U}(\tau, \gamma) = \hat{\mathcal{A}}(\tau, \gamma) + \lambda \log \hat{\mathcal{T}}(\tau, \gamma) ,
\end{equation}
where $\lambda$ dictates efficiency preference, and the logarithm models diminishing returns. Mapping Table~\ref{tab-hyperparams} onto the Pareto frontier identifies the ``knee'' (Figure~\ref{fig:pareto_curve}), revealing $(\tau_{jump} = 0.03, \gamma = 0.6)$ as the theoretical optimum $\arg\max_{\tau, \gamma} \mathcal{U}$. This achieves a $\sim$$2\times$ speedup ($\hat{\mathcal{T}} \approx 1.98$) with marginal $4.6\%$ accuracy degradation ($\hat{\mathcal{A}} \approx 0.954$), confirming it as a principled solution rather than mere empirical coincidence.

\paragraph{Sensitivity of Padding Waste Threshold.} As shown in Figure~7, the padding waste threshold ($\tau_{pad}$) crucially dictates our routing policy. Setting $\tau_{pad} \leq 0.7$ effectively filters highly skewed batches to prevent excessive padding. However, exceeding this tipping point causes a sharp throughput drop across all batch sizes, because more long-tailed batches are forced into static execution paths where a few long sequences dominate the bucket size and reintroduce substantial invalid computation. An excessively high threshold forces long-tailed sequences into static execution paths, incurring severe padding waste that negates algorithmic truncation benefits. This confirms the indispensability of our padding-waste-aware hybrid routing.

\begin{figure}
    \centering
    
    \includegraphics[trim=0 0 0 0, clip, width=\linewidth]{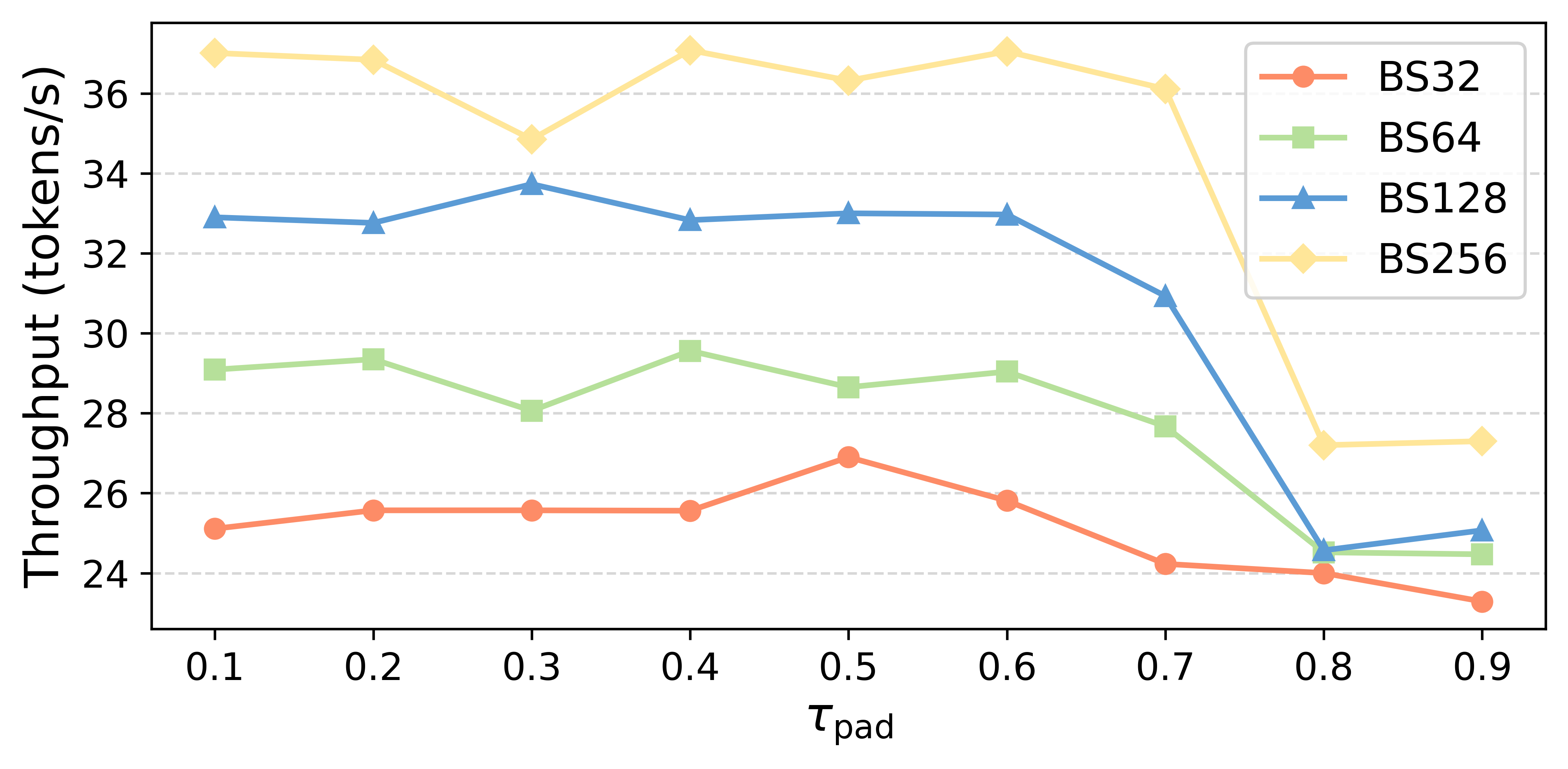}
    \Description{A line graph showing the effect of the tau_pad threshold on inference throughput for four different batch sizes: 32, 64, 128, and 256. The x-axis represents the tau_pad value from 0.1 to 0.9. Throughput remains relatively stable for threshold values up to 0.6 or 0.7, but plummets sharply across all batch sizes as the threshold approaches 0.8 and 0.9.}
    \vspace{-6mm}
    \caption{\textbf{Effect of $\tau_{pad}$ on throughput.} Experiments with LaViDa-LLaDA on ChartQA show that as $\tau_{pad}$ increases beyond a critical point, throughput drops sharply across all batch sizes. This indicates that an excessively high threshold forces too many skewed sequences into static execution paths, incurring severe padding waste that negates the benefits of semantic truncation.}
    \label{fig:system_ablation}
    \vspace{-3mm}  
\end{figure}

\textbf{Boundary Detection Quality.} Table~\ref{tab:boundary_quality} reports MAE and over-truncation rate, which measure the mean absolute boundary prediction error and the risk of premature truncation, respectively. As shown in Table~\ref{tab:boundary_quality}, Step-0 sparsity provides reliable boundary estimates across diverse tasks. DocVQA achieves the best performance, while MathVista is more challenging and exhibits larger boundary error. The consistently low over-truncation rates suggest that Seer rarely removes valid tokens in practice.

\begin{table}[h]
\centering
\caption{Boundary detection quality on three benchmarks.}
\label{tab:boundary_quality}
\begin{tabular}{lccc}
\toprule
\textbf{Metric} & \textbf{ChartQA} & \textbf{DocVQA} & \textbf{MathVista} \\
\midrule
\textbf{MAE $\downarrow$} & 0.86 & 0.40 & 1.18 \\
\textbf{Over-truncation (\%) $\downarrow$} & 3.4 & 0.8 & 3.8 \\
\bottomrule
\end{tabular}
\end{table}


\section{Conclusion}
We identify output-side padding as an important yet underexplored efficiency bottleneck in Diffusion Multimodal Large Language Models, and show that the semantic boundary of a valid response is implicitly revealed at the first denoising step through early-layer text-side MLP sparsity. Based on this observation, we propose Seer, a training-free framework that detects semantic boundaries via an SNR-based criterion and removes redundant suffix computation through one-shot truncation. Combined with a hybrid execution design, Seer translates semantic shortening into practical acceleration under realistic batched serving. Experiments show that Seer consistently improves throughput and latency while maintaining, and in some cases even improving, task performance, highlighting output-side semantic redundancy as a promising direction for efficient diffusion-based multimodal generation. Nevertheless, extending Seer to broader settings and model architectures remains an important direction for future work.

\clearpage



\begin{acks}
This study is partially supported by the National Natural Science Foundation of China (Grant No.62504204).
\end{acks}

\bibliographystyle{ACM-Reference-Format}
\bibliography{sample-base}


\begin{thebibliography}{84}


\ifx \showCODEN    \undefined \def \showCODEN     #1{\unskip}     \fi
\ifx \showISBNx    \undefined \def \showISBNx     #1{\unskip}     \fi
\ifx \showISBNxiii \undefined \def \showISBNxiii  #1{\unskip}     \fi
\ifx \showISSN     \undefined \def \showISSN      #1{\unskip}     \fi
\ifx \showLCCN     \undefined \def \showLCCN      #1{\unskip}     \fi
\ifx \shownote     \undefined \def \shownote      #1{#1}          \fi
\ifx \showarticletitle \undefined \def \showarticletitle #1{#1}   \fi
\ifx \showURL      \undefined \def \showURL       {\relax}        \fi
\providecommand\bibfield[2]{#2}
\providecommand\bibinfo[2]{#2}
\providecommand\natexlab[1]{#1}
\providecommand\showeprint[2][]{arXiv:#2}

\bibitem[Alvar et~al\mbox{.}(2025)]%
        {alvar2025divprune}
\bibfield{author}{\bibinfo{person}{Saeed~Ranjbar Alvar}, \bibinfo{person}{Gursimran Singh}, \bibinfo{person}{Mohammad Akbari}, {and} \bibinfo{person}{Yong Zhang}.} \bibinfo{year}{2025}\natexlab{}.
\newblock \showarticletitle{Divprune: Diversity-based visual token pruning for large multimodal models}. In \bibinfo{booktitle}{\emph{Proceedings of the Computer Vision and Pattern Recognition Conference}}. \bibinfo{pages}{9392--9401}.
\newblock


\bibitem[Arriola et~al\mbox{.}(2025)]%
        {arriola2025block}
\bibfield{author}{\bibinfo{person}{Marianne Arriola}, \bibinfo{person}{Aaron Gokaslan}, \bibinfo{person}{Justin~T Chiu}, \bibinfo{person}{Zhihan Yang}, \bibinfo{person}{Zhixuan Qi}, \bibinfo{person}{Jiaqi Han}, \bibinfo{person}{Subham~Sekhar Sahoo}, {and} \bibinfo{person}{Volodymyr Kuleshov}.} \bibinfo{year}{2025}\natexlab{}.
\newblock \showarticletitle{Block diffusion: Interpolating between autoregressive and diffusion language models}.
\newblock \bibinfo{journal}{\emph{arXiv preprint arXiv:2503.09573}} (\bibinfo{year}{2025}).
\newblock


\bibitem[Baek et~al\mbox{.}(2026)]%
        {baek2026agilepruner}
\bibfield{author}{\bibinfo{person}{Changwoo Baek}, \bibinfo{person}{Jouwon Song}, \bibinfo{person}{Sohyeon Kim}, {and} \bibinfo{person}{Kyeongbo Kong}.} \bibinfo{year}{2026}\natexlab{}.
\newblock \showarticletitle{AgilePruner: An Empirical Study of Attention and Diversity for Adaptive Visual Token Pruning in Large Vision-Language Models}.
\newblock \bibinfo{journal}{\emph{arXiv preprint arXiv:2603.01236}} (\bibinfo{year}{2026}).
\newblock


\bibitem[Bao et~al\mbox{.}(2025)]%
        {bao2025learning}
\bibfield{author}{\bibinfo{person}{Wenrui Bao}, \bibinfo{person}{Zhiben Chen}, \bibinfo{person}{Dan Xu}, {and} \bibinfo{person}{Yuzhang Shang}.} \bibinfo{year}{2025}\natexlab{}.
\newblock \showarticletitle{Learning to parallel: Accelerating diffusion large language models via learnable parallel decoding}.
\newblock \bibinfo{journal}{\emph{arXiv preprint arXiv:2509.25188}} (\bibinfo{year}{2025}).
\newblock


\bibitem[Bolya et~al\mbox{.}(2022)]%
        {bolya2022token}
\bibfield{author}{\bibinfo{person}{Daniel Bolya}, \bibinfo{person}{Cheng-Yang Fu}, \bibinfo{person}{Xiaoliang Dai}, \bibinfo{person}{Peizhao Zhang}, \bibinfo{person}{Christoph Feichtenhofer}, {and} \bibinfo{person}{Judy Hoffman}.} \bibinfo{year}{2022}\natexlab{}.
\newblock \showarticletitle{Token merging: Your vit but faster}.
\newblock \bibinfo{journal}{\emph{arXiv preprint arXiv:2210.09461}} (\bibinfo{year}{2022}).
\newblock


\bibitem[Chang et~al\mbox{.}(2025)]%
        {chang2025d}
\bibfield{author}{\bibinfo{person}{Shuochen Chang}, \bibinfo{person}{Xiaofeng Zhang}, \bibinfo{person}{Qingyang Liu}, {and} \bibinfo{person}{Li Niu}.} \bibinfo{year}{2025}\natexlab{}.
\newblock \showarticletitle{{D$^3$ ToM}: Decider-Guided Dynamic Token Merging for Accelerating Diffusion MLLMs}.
\newblock \bibinfo{journal}{\emph{arXiv preprint arXiv:2511.12280}} (\bibinfo{year}{2025}).
\newblock


\bibitem[Chen et~al\mbox{.}(2024b)]%
        {chen2024image}
\bibfield{author}{\bibinfo{person}{Liang Chen}, \bibinfo{person}{Haozhe Zhao}, \bibinfo{person}{Tianyu Liu}, \bibinfo{person}{Shuai Bai}, \bibinfo{person}{Junyang Lin}, \bibinfo{person}{Chang Zhou}, {and} \bibinfo{person}{Baobao Chang}.} \bibinfo{year}{2024}\natexlab{b}.
\newblock \bibinfo{title}{An Image is Worth 1/2 Tokens After Layer 2: Plug-and-Play Inference Acceleration for Large Vision-Language Models}.
\newblock
\showeprint[arxiv]{2403.06764}~[cs.CV]


\bibitem[Chen et~al\mbox{.}(2026)]%
        {chen2026otprune}
\bibfield{author}{\bibinfo{person}{Xiwen Chen}, \bibinfo{person}{Wenhui Zhu}, \bibinfo{person}{Gen Li}, \bibinfo{person}{Xuanzhao Dong}, \bibinfo{person}{Yujian Xiong}, \bibinfo{person}{Hao Wang}, \bibinfo{person}{Peijie Qiu}, \bibinfo{person}{Qingquan Song}, \bibinfo{person}{Zhipeng Wang}, \bibinfo{person}{Shao Tang}, {et~al\mbox{.}}} \bibinfo{year}{2026}\natexlab{}.
\newblock \showarticletitle{OTPrune: Distribution-Aligned Visual Token Pruning via Optimal Transport}.
\newblock \bibinfo{journal}{\emph{arXiv preprint arXiv:2602.20205}} (\bibinfo{year}{2026}).
\newblock


\bibitem[Chen et~al\mbox{.}(2024a)]%
        {chen2024internvl}
\bibfield{author}{\bibinfo{person}{Zhe Chen}, \bibinfo{person}{Jiannan Wu}, \bibinfo{person}{Wenhai Wang}, \bibinfo{person}{Weijie Su}, \bibinfo{person}{Guo Chen}, \bibinfo{person}{Sen Xing}, \bibinfo{person}{Muyan Zhong}, \bibinfo{person}{Qinglong Zhang}, \bibinfo{person}{Xizhou Zhu}, \bibinfo{person}{Lewei Lu}, {et~al\mbox{.}}} \bibinfo{year}{2024}\natexlab{a}.
\newblock \showarticletitle{Internvl: Scaling up vision foundation models and aligning for generic visual-linguistic tasks}. In \bibinfo{booktitle}{\emph{Proceedings of the IEEE/CVF conference on computer vision and pattern recognition}}. \bibinfo{pages}{24185--24198}.
\newblock


\bibitem[Cheng et~al\mbox{.}(2025a)]%
        {cheng2025sdar}
\bibfield{author}{\bibinfo{person}{Shuang Cheng}, \bibinfo{person}{Yuhua Jiang}, \bibinfo{person}{Zineng Zhou}, \bibinfo{person}{Dawei Liu}, \bibinfo{person}{Wang Tao}, \bibinfo{person}{Linfeng Zhang}, \bibinfo{person}{Biqing Qi}, {and} \bibinfo{person}{Bowen Zhou}.} \bibinfo{year}{2025}\natexlab{a}.
\newblock \showarticletitle{Sdar-vl: Stable and efficient block-wise diffusion for vision-language understanding}.
\newblock \bibinfo{journal}{\emph{arXiv preprint arXiv:2512.14068}} (\bibinfo{year}{2025}).
\newblock


\bibitem[Cheng et~al\mbox{.}(2025b)]%
        {cheng2025deer}
\bibfield{author}{\bibinfo{person}{Zicong Cheng}, \bibinfo{person}{Guo-Wei Yang}, \bibinfo{person}{Jia Li}, \bibinfo{person}{Zhijie Deng}, \bibinfo{person}{Meng-Hao Guo}, {and} \bibinfo{person}{Shi-Min Hu}.} \bibinfo{year}{2025}\natexlab{b}.
\newblock \showarticletitle{Deer: Draft with diffusion, verify with autoregressive models}.
\newblock \bibinfo{journal}{\emph{arXiv preprint arXiv:2512.15176}} (\bibinfo{year}{2025}).
\newblock


\bibitem[Dhouib et~al\mbox{.}(2025)]%
        {dhouib2025pact}
\bibfield{author}{\bibinfo{person}{Mohamed Dhouib}, \bibinfo{person}{Davide Buscaldi}, \bibinfo{person}{Sonia Vanier}, {and} \bibinfo{person}{Aymen Shabou}.} \bibinfo{year}{2025}\natexlab{}.
\newblock \showarticletitle{Pact: Pruning and clustering-based token reduction for faster visual language models}. In \bibinfo{booktitle}{\emph{Proceedings of the Computer Vision and Pattern Recognition Conference}}. \bibinfo{pages}{14582--14592}.
\newblock


\bibitem[Dong et~al\mbox{.}(2025)]%
        {dong2025mmtok}
\bibfield{author}{\bibinfo{person}{Sixun Dong}, \bibinfo{person}{Juhua Hu}, \bibinfo{person}{Mian Zhang}, \bibinfo{person}{Ming Yin}, \bibinfo{person}{Yanjie Fu}, {and} \bibinfo{person}{Qi Qian}.} \bibinfo{year}{2025}\natexlab{}.
\newblock \showarticletitle{Mmtok: Multimodal coverage maximization for efficient inference of vlms}.
\newblock \bibinfo{journal}{\emph{arXiv preprint arXiv:2508.18264}} (\bibinfo{year}{2025}).
\newblock


\bibitem[Fu et~al\mbox{.}(2023)]%
        {fu2023mme}
\bibfield{author}{\bibinfo{person}{Chaoyou Fu}, \bibinfo{person}{Peixian Chen}, \bibinfo{person}{Yunhang Shen}, \bibinfo{person}{Yulei Qin}, \bibinfo{person}{Mengdan Zhang}, \bibinfo{person}{Xu Lin}, \bibinfo{person}{Jinrui Yang}, \bibinfo{person}{Xiawu Zheng}, \bibinfo{person}{Ke Li}, \bibinfo{person}{Xing Sun}, {et~al\mbox{.}}} \bibinfo{year}{2023}\natexlab{}.
\newblock \showarticletitle{Mme: A comprehensive evaluation benchmark for multimodal large language models}.
\newblock \bibinfo{journal}{\emph{arXiv preprint arXiv:2306.13394}} (\bibinfo{year}{2023}).
\newblock


\bibitem[He et~al\mbox{.}(2024)]%
        {he2024zipvl}
\bibfield{author}{\bibinfo{person}{Yefei He}, \bibinfo{person}{Feng Chen}, \bibinfo{person}{Jing Liu}, \bibinfo{person}{Wenqi Shao}, \bibinfo{person}{Hong Zhou}, \bibinfo{person}{Kaipeng Zhang}, {and} \bibinfo{person}{Bohan Zhuang}.} \bibinfo{year}{2024}\natexlab{}.
\newblock \showarticletitle{Zipvl: Efficient large vision-language models with dynamic token sparsification}.
\newblock \bibinfo{journal}{\emph{arXiv preprint arXiv:2410.08584}} (\bibinfo{year}{2024}).
\newblock


\bibitem[Huang et~al\mbox{.}(2025)]%
        {huang2025mask}
\bibfield{author}{\bibinfo{person}{Jianuo Huang}, \bibinfo{person}{Yaojie Zhang}, \bibinfo{person}{Yicun Yang}, \bibinfo{person}{Benhao Huang}, \bibinfo{person}{Biqing Qi}, \bibinfo{person}{Dongrui Liu}, {and} \bibinfo{person}{Linfeng Zhang}.} \bibinfo{year}{2025}\natexlab{}.
\newblock \showarticletitle{Mask tokens as prophet: Fine-grained cache eviction for efficient dllm inference}.
\newblock \bibinfo{journal}{\emph{arXiv preprint arXiv:2510.09309}} (\bibinfo{year}{2025}).
\newblock


\bibitem[Hudson and Manning(2019)]%
        {hudson2019gqa}
\bibfield{author}{\bibinfo{person}{Drew~A Hudson} {and} \bibinfo{person}{Christopher~D Manning}.} \bibinfo{year}{2019}\natexlab{}.
\newblock \showarticletitle{Gqa: A new dataset for real-world visual reasoning and compositional question answering}. In \bibinfo{booktitle}{\emph{Proceedings of the IEEE/CVF conference on computer vision and pattern recognition}}. \bibinfo{pages}{6700--6709}.
\newblock


\bibitem[Israel et~al\mbox{.}(2025)]%
        {israel2025accelerating}
\bibfield{author}{\bibinfo{person}{Daniel Israel}, \bibinfo{person}{Guy Van~den Broeck}, {and} \bibinfo{person}{Aditya Grover}.} \bibinfo{year}{2025}\natexlab{}.
\newblock \showarticletitle{Accelerating diffusion llms via adaptive parallel decoding}.
\newblock \bibinfo{journal}{\emph{arXiv preprint arXiv:2506.00413}} (\bibinfo{year}{2025}).
\newblock


\bibitem[Jiang et~al\mbox{.}(2025b)]%
        {jiang2025sada}
\bibfield{author}{\bibinfo{person}{Ting Jiang}, \bibinfo{person}{Yixiao Wang}, \bibinfo{person}{Hancheng Ye}, \bibinfo{person}{Zishan Shao}, \bibinfo{person}{Jingwei Sun}, \bibinfo{person}{Jingyang Zhang}, \bibinfo{person}{Zekai Chen}, \bibinfo{person}{Jianyi Zhang}, \bibinfo{person}{Yiran Chen}, {and} \bibinfo{person}{Hai Li}.} \bibinfo{year}{2025}\natexlab{b}.
\newblock \showarticletitle{Sada: Stability-guided adaptive diffusion acceleration}.
\newblock \bibinfo{journal}{\emph{arXiv preprint arXiv:2507.17135}} (\bibinfo{year}{2025}).
\newblock


\bibitem[Jiang et~al\mbox{.}(2025a)]%
        {jiang2025d}
\bibfield{author}{\bibinfo{person}{Yuchu Jiang}, \bibinfo{person}{Yue Cai}, \bibinfo{person}{Xiangzhong Luo}, \bibinfo{person}{Jiale Fu}, \bibinfo{person}{Jiarui Wang}, \bibinfo{person}{Chonghan Liu}, {and} \bibinfo{person}{Xu Yang}.} \bibinfo{year}{2025}\natexlab{a}.
\newblock \showarticletitle{d{$^2$}Cache: Accelerating Diffusion-Based LLMs via Dual Adaptive Caching}.
\newblock \bibinfo{journal}{\emph{arXiv preprint arXiv:2509.23094}} (\bibinfo{year}{2025}).
\newblock


\bibitem[Jiang et~al\mbox{.}(2025c)]%
        {jiang2025kind}
\bibfield{author}{\bibinfo{person}{Yutao Jiang}, \bibinfo{person}{Qiong Wu}, \bibinfo{person}{Wenhao Lin}, \bibinfo{person}{Wei Yu}, {and} \bibinfo{person}{Yiyi Zhou}.} \bibinfo{year}{2025}\natexlab{c}.
\newblock \showarticletitle{What kind of visual tokens do we need? training-free visual token pruning for multi-modal large language models from the perspective of graph}. In \bibinfo{booktitle}{\emph{Proceedings of the AAAI Conference on Artificial Intelligence}}, Vol.~\bibinfo{volume}{39}. \bibinfo{pages}{4075--4083}.
\newblock


\bibitem[Kim et~al\mbox{.}(2025)]%
        {kim2025any}
\bibfield{author}{\bibinfo{person}{Jaeyeon Kim}, \bibinfo{person}{Lee Cheuk-Kit}, \bibinfo{person}{Carles Domingo-Enrich}, \bibinfo{person}{Yilun Du}, \bibinfo{person}{Sham Kakade}, \bibinfo{person}{Timothy Ngotiaoco}, \bibinfo{person}{Sitan Chen}, {and} \bibinfo{person}{Michael Albergo}.} \bibinfo{year}{2025}\natexlab{}.
\newblock \showarticletitle{Any-order flexible length masked diffusion}.
\newblock \bibinfo{journal}{\emph{arXiv preprint arXiv:2509.01025}} (\bibinfo{year}{2025}).
\newblock


\bibitem[Li et~al\mbox{.}(2025e)]%
        {li2025comprehensive}
\bibfield{author}{\bibinfo{person}{Duo Li}, \bibinfo{person}{Zuhao Yang}, \bibinfo{person}{Xiaoqin Zhang}, \bibinfo{person}{Ling Shao}, {and} \bibinfo{person}{Shijian Lu}.} \bibinfo{year}{2025}\natexlab{e}.
\newblock \showarticletitle{A Comprehensive Study on Visual Token Redundancy for Discrete Diffusion-based Multimodal Large Language Models}.
\newblock \bibinfo{journal}{\emph{arXiv preprint arXiv:2511.15098}} (\bibinfo{year}{2025}).
\newblock


\bibitem[Li et~al\mbox{.}(2025b)]%
        {li2025diffuspec}
\bibfield{author}{\bibinfo{person}{Guanghao Li}, \bibinfo{person}{Zhihui Fu}, \bibinfo{person}{Min Fang}, \bibinfo{person}{Qibin Zhao}, \bibinfo{person}{Ming Tang}, \bibinfo{person}{Chun Yuan}, {and} \bibinfo{person}{Jun Wang}.} \bibinfo{year}{2025}\natexlab{b}.
\newblock \showarticletitle{Diffuspec: Unlocking diffusion language models for speculative decoding}.
\newblock \bibinfo{journal}{\emph{arXiv preprint arXiv:2510.02358}} (\bibinfo{year}{2025}).
\newblock


\bibitem[Li et~al\mbox{.}(2025a)]%
        {li2025beyond}
\bibfield{author}{\bibinfo{person}{Jinsong Li}, \bibinfo{person}{Xiaoyi Dong}, \bibinfo{person}{Yuhang Zang}, \bibinfo{person}{Yuhang Cao}, \bibinfo{person}{Jiaqi Wang}, {and} \bibinfo{person}{Dahua Lin}.} \bibinfo{year}{2025}\natexlab{a}.
\newblock \showarticletitle{Beyond fixed: Training-free variable-length denoising for diffusion large language models}.
\newblock \bibinfo{journal}{\emph{arXiv preprint arXiv:2508.00819}} (\bibinfo{year}{2025}).
\newblock


\bibitem[Li et~al\mbox{.}(2025c)]%
        {li2025sparse}
\bibfield{author}{\bibinfo{person}{Shufan Li}, \bibinfo{person}{Jiuxiang Gu}, \bibinfo{person}{Kangning Liu}, \bibinfo{person}{Zhe Lin}, \bibinfo{person}{Zijun Wei}, \bibinfo{person}{Aditya Grover}, {and} \bibinfo{person}{Jason Kuen}.} \bibinfo{year}{2025}\natexlab{c}.
\newblock \showarticletitle{Sparse-LaViDa: Sparse Multimodal Discrete Diffusion Language Models}.
\newblock \bibinfo{journal}{\emph{arXiv preprint arXiv:2512.14008}} (\bibinfo{year}{2025}).
\newblock


\bibitem[Li et~al\mbox{.}(2025d)]%
        {li2025lavida}
\bibfield{author}{\bibinfo{person}{Shufan Li}, \bibinfo{person}{Konstantinos Kallidromitis}, \bibinfo{person}{Hritik Bansal}, \bibinfo{person}{Akash Gokul}, \bibinfo{person}{Yusuke Kato}, \bibinfo{person}{Kazuki Kozuka}, \bibinfo{person}{Jason Kuen}, \bibinfo{person}{Zhe Lin}, \bibinfo{person}{Kai-Wei Chang}, {and} \bibinfo{person}{Aditya Grover}.} \bibinfo{year}{2025}\natexlab{d}.
\newblock \showarticletitle{Lavida: A large diffusion language model for multimodal understanding}.
\newblock \bibinfo{journal}{\emph{arXiv preprint arXiv:2505.16839}} (\bibinfo{year}{2025}).
\newblock


\bibitem[Liang et~al\mbox{.}(2026)]%
        {liang2026focus}
\bibfield{author}{\bibinfo{person}{Kaihua Liang}, \bibinfo{person}{Xin Tan}, \bibinfo{person}{An Zhong}, \bibinfo{person}{Hong Xu}, {and} \bibinfo{person}{Marco Canini}.} \bibinfo{year}{2026}\natexlab{}.
\newblock \showarticletitle{FOCUS: DLLMs Know How to Tame Their Compute Bound}.
\newblock \bibinfo{journal}{\emph{arXiv preprint arXiv:2601.23278}} (\bibinfo{year}{2026}).
\newblock


\bibitem[Lin et~al\mbox{.}(2025b)]%
        {lin2025quantization}
\bibfield{author}{\bibinfo{person}{Haokun Lin}, \bibinfo{person}{Haobo Xu}, \bibinfo{person}{Yichen Wu}, \bibinfo{person}{Ziyu Guo}, \bibinfo{person}{Renrui Zhang}, \bibinfo{person}{Zhichao Lu}, \bibinfo{person}{Ying Wei}, \bibinfo{person}{Qingfu Zhang}, {and} \bibinfo{person}{Zhenan Sun}.} \bibinfo{year}{2025}\natexlab{b}.
\newblock \showarticletitle{Quantization meets dllms: A systematic study of post-training quantization for diffusion llms}.
\newblock \bibinfo{journal}{\emph{arXiv preprint arXiv:2508.14896}} (\bibinfo{year}{2025}).
\newblock


\bibitem[Lin et~al\mbox{.}(2025a)]%
        {lin2025speculative}
\bibfield{author}{\bibinfo{person}{Luxi Lin}, \bibinfo{person}{Zhihang Lin}, \bibinfo{person}{Zhanpeng Zeng}, {and} \bibinfo{person}{Rongrong Ji}.} \bibinfo{year}{2025}\natexlab{a}.
\newblock \showarticletitle{Speculative decoding reimagined for multimodal large language models}.
\newblock \bibinfo{journal}{\emph{arXiv preprint arXiv:2505.14260}} (\bibinfo{year}{2025}).
\newblock


\bibitem[Liu et~al\mbox{.}(2026)]%
        {liu2026dart}
\bibfield{author}{\bibinfo{person}{Fuliang Liu}, \bibinfo{person}{Xue Li}, \bibinfo{person}{Ketai Zhao}, \bibinfo{person}{Yinxi Gao}, \bibinfo{person}{Ziyan Zhou}, \bibinfo{person}{Zhonghui Zhang}, \bibinfo{person}{Zhibin Wang}, \bibinfo{person}{Wanchun Dou}, \bibinfo{person}{Sheng Zhong}, {and} \bibinfo{person}{Chen Tian}.} \bibinfo{year}{2026}\natexlab{}.
\newblock \showarticletitle{DART: Diffusion-Inspired Speculative Decoding for Fast LLM Inference}.
\newblock \bibinfo{journal}{\emph{arXiv preprint arXiv:2601.19278}} (\bibinfo{year}{2026}).
\newblock


\bibitem[Liu et~al\mbox{.}(2024b)]%
        {liu2024improved}
\bibfield{author}{\bibinfo{person}{Haotian Liu}, \bibinfo{person}{Chunyuan Li}, \bibinfo{person}{Yuheng Li}, {and} \bibinfo{person}{Yong~Jae Lee}.} \bibinfo{year}{2024}\natexlab{b}.
\newblock \showarticletitle{Improved baselines with visual instruction tuning}. In \bibinfo{booktitle}{\emph{Proceedings of the IEEE/CVF conference on computer vision and pattern recognition}}. \bibinfo{pages}{26296--26306}.
\newblock


\bibitem[Liu et~al\mbox{.}(2023)]%
        {liu2023visual}
\bibfield{author}{\bibinfo{person}{Haotian Liu}, \bibinfo{person}{Chunyuan Li}, \bibinfo{person}{Qingyang Wu}, {and} \bibinfo{person}{Yong~Jae Lee}.} \bibinfo{year}{2023}\natexlab{}.
\newblock \showarticletitle{Visual instruction tuning}.
\newblock \bibinfo{journal}{\emph{Advances in neural information processing systems}}  \bibinfo{volume}{36} (\bibinfo{year}{2023}), \bibinfo{pages}{34892--34916}.
\newblock


\bibitem[Liu et~al\mbox{.}(2024a)]%
        {liu2024mmbench}
\bibfield{author}{\bibinfo{person}{Yuan Liu}, \bibinfo{person}{Haodong Duan}, \bibinfo{person}{Yuanhan Zhang}, \bibinfo{person}{Bo Li}, \bibinfo{person}{Songyang Zhang}, \bibinfo{person}{Wangbo Zhao}, \bibinfo{person}{Yike Yuan}, \bibinfo{person}{Jiaqi Wang}, \bibinfo{person}{Conghui He}, \bibinfo{person}{Ziwei Liu}, {et~al\mbox{.}}} \bibinfo{year}{2024}\natexlab{a}.
\newblock \showarticletitle{Mmbench: Is your multi-modal model an all-around player?}. In \bibinfo{booktitle}{\emph{European conference on computer vision}}. Springer, \bibinfo{pages}{216--233}.
\newblock


\bibitem[Liu et~al\mbox{.}(2025)]%
        {liu2025dllm}
\bibfield{author}{\bibinfo{person}{Zhiyuan Liu}, \bibinfo{person}{Yicun Yang}, \bibinfo{person}{Yaojie Zhang}, \bibinfo{person}{Junjie Chen}, \bibinfo{person}{Chang Zou}, \bibinfo{person}{Qingyuan Wei}, \bibinfo{person}{Shaobo Wang}, {and} \bibinfo{person}{Linfeng Zhang}.} \bibinfo{year}{2025}\natexlab{}.
\newblock \showarticletitle{dllm-cache: Accelerating diffusion large language models with adaptive caching}.
\newblock \bibinfo{journal}{\emph{arXiv preprint arXiv:2506.06295}} (\bibinfo{year}{2025}).
\newblock


\bibitem[Lu et~al\mbox{.}(2023)]%
        {lu2023mathvista}
\bibfield{author}{\bibinfo{person}{Pan Lu}, \bibinfo{person}{Hritik Bansal}, \bibinfo{person}{Tony Xia}, \bibinfo{person}{Jiacheng Liu}, \bibinfo{person}{Chunyuan Li}, \bibinfo{person}{Hannaneh Hajishirzi}, \bibinfo{person}{Hao Cheng}, \bibinfo{person}{Kai-Wei Chang}, \bibinfo{person}{Michel Galley}, {and} \bibinfo{person}{Jianfeng Gao}.} \bibinfo{year}{2023}\natexlab{}.
\newblock \showarticletitle{Mathvista: Evaluating mathematical reasoning of foundation models in visual contexts}.
\newblock \bibinfo{journal}{\emph{arXiv preprint arXiv:2310.02255}} (\bibinfo{year}{2023}).
\newblock


\bibitem[Lu et~al\mbox{.}(2022)]%
        {lu2022learn}
\bibfield{author}{\bibinfo{person}{Pan Lu}, \bibinfo{person}{Swaroop Mishra}, \bibinfo{person}{Tanglin Xia}, \bibinfo{person}{Liang Qiu}, \bibinfo{person}{Kai-Wei Chang}, \bibinfo{person}{Song-Chun Zhu}, \bibinfo{person}{Oyvind Tafjord}, \bibinfo{person}{Peter Clark}, {and} \bibinfo{person}{Ashwin Kalyan}.} \bibinfo{year}{2022}\natexlab{}.
\newblock \showarticletitle{Learn to explain: Multimodal reasoning via thought chains for science question answering}.
\newblock \bibinfo{journal}{\emph{Advances in neural information processing systems}}  \bibinfo{volume}{35} (\bibinfo{year}{2022}), \bibinfo{pages}{2507--2521}.
\newblock


\bibitem[Luo et~al\mbox{.}(2025)]%
        {luo2025large}
\bibfield{author}{\bibinfo{person}{Junwei Luo}, \bibinfo{person}{Yingying Zhang}, \bibinfo{person}{Xue Yang}, \bibinfo{person}{Kang Wu}, \bibinfo{person}{Qi Zhu}, \bibinfo{person}{Lei Liang}, \bibinfo{person}{Jingdong Chen}, {and} \bibinfo{person}{Yansheng Li}.} \bibinfo{year}{2025}\natexlab{}.
\newblock \showarticletitle{When large vision-language model meets large remote sensing imagery: Coarse-to-fine text-guided token pruning}. In \bibinfo{booktitle}{\emph{Proceedings of the IEEE/CVF International Conference on Computer Vision}}. \bibinfo{pages}{9206--9217}.
\newblock


\bibitem[Ma et~al\mbox{.}(2025b)]%
        {ma2025short}
\bibfield{author}{\bibinfo{person}{Ji Ma}, \bibinfo{person}{Wei Suo}, \bibinfo{person}{Peng Wang}, {and} \bibinfo{person}{Yanning Zhang}.} \bibinfo{year}{2025}\natexlab{b}.
\newblock \showarticletitle{Short-lvlm: Compressing and accelerating large vision-language models by pruning redundant layers}. In \bibinfo{booktitle}{\emph{Proceedings of the 33rd ACM International Conference on Multimedia}}. \bibinfo{pages}{3575--3584}.
\newblock


\bibitem[Ma et~al\mbox{.}(2026)]%
        {ma2026apet}
\bibfield{author}{\bibinfo{person}{Qiankun Ma}, \bibinfo{person}{Ziyao Zhang}, \bibinfo{person}{Haofei Wang}, \bibinfo{person}{Jie Chen}, \bibinfo{person}{Zhen Song}, {and} \bibinfo{person}{Hairong Zheng}.} \bibinfo{year}{2026}\natexlab{}.
\newblock \showarticletitle{ApET: Approximation-Error Guided Token Compression for Efficient VLMs}.
\newblock \bibinfo{journal}{\emph{arXiv preprint arXiv:2602.19870}} (\bibinfo{year}{2026}).
\newblock


\bibitem[Ma et~al\mbox{.}(2025c)]%
        {ma2025dkv}
\bibfield{author}{\bibinfo{person}{Xinyin Ma}, \bibinfo{person}{Runpeng Yu}, \bibinfo{person}{Gongfan Fang}, {and} \bibinfo{person}{Xinchao Wang}.} \bibinfo{year}{2025}\natexlab{c}.
\newblock \showarticletitle{dkv-cache: The cache for diffusion language models}.
\newblock \bibinfo{journal}{\emph{arXiv preprint arXiv:2505.15781}} (\bibinfo{year}{2025}).
\newblock


\bibitem[Ma et~al\mbox{.}(2025a)]%
        {ma2025dinfer}
\bibfield{author}{\bibinfo{person}{Yuxin Ma}, \bibinfo{person}{Lun Du}, \bibinfo{person}{Lanning Wei}, \bibinfo{person}{Kun Chen}, \bibinfo{person}{Qian Xu}, \bibinfo{person}{Kangyu Wang}, \bibinfo{person}{Guofeng Feng}, \bibinfo{person}{Guoshan Lu}, \bibinfo{person}{Lin Liu}, \bibinfo{person}{Xiaojing Qi}, {et~al\mbox{.}}} \bibinfo{year}{2025}\natexlab{a}.
\newblock \showarticletitle{dinfer: An efficient inference framework for diffusion language models}.
\newblock \bibinfo{journal}{\emph{arXiv preprint arXiv:2510.08666}} (\bibinfo{year}{2025}).
\newblock


\bibitem[Masry et~al\mbox{.}(2022)]%
        {masry2022chartqa}
\bibfield{author}{\bibinfo{person}{Ahmed Masry}, \bibinfo{person}{Xuan~Long Do}, \bibinfo{person}{Jia~Qing Tan}, \bibinfo{person}{Shafiq Joty}, {and} \bibinfo{person}{Enamul Hoque}.} \bibinfo{year}{2022}\natexlab{}.
\newblock \showarticletitle{Chartqa: A benchmark for question answering about charts with visual and logical reasoning}. In \bibinfo{booktitle}{\emph{Findings of the association for computational linguistics: ACL 2022}}. \bibinfo{pages}{2263--2279}.
\newblock


\bibitem[Mathew et~al\mbox{.}(2022)]%
        {mathew2022infographicvqa}
\bibfield{author}{\bibinfo{person}{Minesh Mathew}, \bibinfo{person}{Viraj Bagal}, \bibinfo{person}{Rub{\`e}n Tito}, \bibinfo{person}{Dimosthenis Karatzas}, \bibinfo{person}{Ernest Valveny}, {and} \bibinfo{person}{CV Jawahar}.} \bibinfo{year}{2022}\natexlab{}.
\newblock \showarticletitle{Infographicvqa}. In \bibinfo{booktitle}{\emph{Proceedings of the IEEE/CVF Winter Conference on Applications of Computer Vision}}. \bibinfo{pages}{1697--1706}.
\newblock


\bibitem[Mathew et~al\mbox{.}(2021)]%
        {mathew2021docvqa}
\bibfield{author}{\bibinfo{person}{Minesh Mathew}, \bibinfo{person}{Dimosthenis Karatzas}, {and} \bibinfo{person}{CV Jawahar}.} \bibinfo{year}{2021}\natexlab{}.
\newblock \showarticletitle{Docvqa: A dataset for vqa on document images}. In \bibinfo{booktitle}{\emph{Proceedings of the IEEE/CVF winter conference on applications of computer vision}}. \bibinfo{pages}{2200--2209}.
\newblock


\bibitem[Nguyen-Tri et~al\mbox{.}(2025)]%
        {nguyen2025attention}
\bibfield{author}{\bibinfo{person}{Quan Nguyen-Tri}, \bibinfo{person}{Mukul Ranjan}, {and} \bibinfo{person}{Zhiqiang Shen}.} \bibinfo{year}{2025}\natexlab{}.
\newblock \showarticletitle{Attention is all you need for kv cache in diffusion llms}.
\newblock \bibinfo{journal}{\emph{arXiv preprint arXiv:2510.14973}} (\bibinfo{year}{2025}).
\newblock


\bibitem[Nie et~al\mbox{.}(2025)]%
        {nie2025large}
\bibfield{author}{\bibinfo{person}{Shen Nie}, \bibinfo{person}{Fengqi Zhu}, \bibinfo{person}{Zebin You}, \bibinfo{person}{Xiaolu Zhang}, \bibinfo{person}{Jingyang Ou}, \bibinfo{person}{Jun Hu}, \bibinfo{person}{Jun Zhou}, \bibinfo{person}{Yankai Lin}, \bibinfo{person}{Ji-Rong Wen}, {and} \bibinfo{person}{Chongxuan Li}.} \bibinfo{year}{2025}\natexlab{}.
\newblock \showarticletitle{Large language diffusion models}.
\newblock \bibinfo{journal}{\emph{arXiv preprint arXiv:2502.09992}} (\bibinfo{year}{2025}).
\newblock


\bibitem[Rao et~al\mbox{.}(2021)]%
        {rao2021dynamicvit}
\bibfield{author}{\bibinfo{person}{Yongming Rao}, \bibinfo{person}{Wenliang Zhao}, \bibinfo{person}{Benlin Liu}, \bibinfo{person}{Jiwen Lu}, \bibinfo{person}{Jie Zhou}, {and} \bibinfo{person}{Cho-Jui Hsieh}.} \bibinfo{year}{2021}\natexlab{}.
\newblock \showarticletitle{Dynamicvit: Efficient vision transformers with dynamic token sparsification}.
\newblock \bibinfo{journal}{\emph{Advances in neural information processing systems}}  \bibinfo{volume}{34} (\bibinfo{year}{2021}), \bibinfo{pages}{13937--13949}.
\newblock


\bibitem[Song et~al\mbox{.}({[n.\,d.]})]%
        {song2508sparse}
\bibfield{author}{\bibinfo{person}{Yuerong Song}, \bibinfo{person}{Xiaoran Liu}, \bibinfo{person}{Ruixiao Li}, \bibinfo{person}{Zhigeng Liu}, \bibinfo{person}{Zengfeng Huang}, \bibinfo{person}{Qipeng Guo}, \bibinfo{person}{Ziwei He}, {and} \bibinfo{person}{Xipeng Qiu}.} \bibinfo{year}{[n.\,d.]}\natexlab{}.
\newblock \showarticletitle{Sparse-dllm: Accelerating diffusion llms with dynamic cache eviction, 2025a}.
\newblock \bibinfo{journal}{\emph{URL https://arxiv. org/abs/2508.02558}} (\bibinfo{year}{[n.\,d.]}).
\newblock


\bibitem[Sun et~al\mbox{.}(2025)]%
        {sun2025lvpruning}
\bibfield{author}{\bibinfo{person}{Yizheng Sun}, \bibinfo{person}{Yanze Xin}, \bibinfo{person}{Hao Li}, \bibinfo{person}{Jingyuan Sun}, \bibinfo{person}{Chenghua Lin}, {and} \bibinfo{person}{Riza~Theresa Batista-Navarro}.} \bibinfo{year}{2025}\natexlab{}.
\newblock \showarticletitle{Lvpruning: An effective yet simple language-guided vision token pruning approach for multi-modal large language models}. In \bibinfo{booktitle}{\emph{Findings of the Association for Computational Linguistics: NAACL 2025}}. \bibinfo{pages}{4299--4308}.
\newblock


\bibitem[Vasu et~al\mbox{.}(2025)]%
        {vasu2025fastvlm}
\bibfield{author}{\bibinfo{person}{Pavan Kumar~Anasosalu Vasu}, \bibinfo{person}{Fartash Faghri}, \bibinfo{person}{Chun-Liang Li}, \bibinfo{person}{Cem Koc}, \bibinfo{person}{Nate True}, \bibinfo{person}{Albert Antony}, \bibinfo{person}{Gokula Santhanam}, \bibinfo{person}{James Gabriel}, \bibinfo{person}{Peter Grasch}, \bibinfo{person}{Oncel Tuzel}, {et~al\mbox{.}}} \bibinfo{year}{2025}\natexlab{}.
\newblock \showarticletitle{Fastvlm: Efficient vision encoding for vision language models}. In \bibinfo{booktitle}{\emph{Proceedings of the Computer Vision and Pattern Recognition Conference}}. \bibinfo{pages}{19769--19780}.
\newblock


\bibitem[Wang et~al\mbox{.}(2024)]%
        {wang2024qwen2}
\bibfield{author}{\bibinfo{person}{Peng Wang}, \bibinfo{person}{Shuai Bai}, \bibinfo{person}{Sinan Tan}, \bibinfo{person}{Shijie Wang}, \bibinfo{person}{Zhihao Fan}, \bibinfo{person}{Jinze Bai}, \bibinfo{person}{Keqin Chen}, \bibinfo{person}{Xuejing Liu}, \bibinfo{person}{Jialin Wang}, \bibinfo{person}{Wenbin Ge}, {et~al\mbox{.}}} \bibinfo{year}{2024}\natexlab{}.
\newblock \showarticletitle{Qwen2-vl: Enhancing vision-language model's perception of the world at any resolution}.
\newblock \bibinfo{journal}{\emph{arXiv preprint arXiv:2409.12191}} (\bibinfo{year}{2024}).
\newblock


\bibitem[Wang et~al\mbox{.}(2025)]%
        {wang2025diffusion}
\bibfield{author}{\bibinfo{person}{Xu Wang}, \bibinfo{person}{Chenkai Xu}, \bibinfo{person}{Yijie Jin}, \bibinfo{person}{Jiachun Jin}, \bibinfo{person}{Hao Zhang}, {and} \bibinfo{person}{Zhijie Deng}.} \bibinfo{year}{2025}\natexlab{}.
\newblock \showarticletitle{Diffusion llms can do faster-than-ar inference via discrete diffusion forcing}.
\newblock \bibinfo{journal}{\emph{arXiv preprint arXiv:2508.09192}} (\bibinfo{year}{2025}).
\newblock


\bibitem[Wang et~al\mbox{.}(2026)]%
        {wang2026zeus}
\bibfield{author}{\bibinfo{person}{Yixiao Wang}, \bibinfo{person}{Ting Jiang}, \bibinfo{person}{Zishan Shao}, \bibinfo{person}{Hancheng Ye}, \bibinfo{person}{Jingwei Sun}, \bibinfo{person}{Mingyuan Ma}, \bibinfo{person}{Jianyi Zhang}, \bibinfo{person}{Yiran Chen}, {and} \bibinfo{person}{Hai Li}.} \bibinfo{year}{2026}\natexlab{}.
\newblock \showarticletitle{Zeus: Accelerating diffusion models with only second-order predictor}.
\newblock \bibinfo{journal}{\emph{arXiv preprint arXiv:2604.01552}} (\bibinfo{year}{2026}).
\newblock


\bibitem[Wei et~al\mbox{.}(2025)]%
        {wei2025orchestrating}
\bibfield{author}{\bibinfo{person}{Linye Wei}, \bibinfo{person}{Wenjue Chen}, \bibinfo{person}{Pingzhi Tang}, \bibinfo{person}{Xiaotian Guo}, \bibinfo{person}{Le Ye}, \bibinfo{person}{Runsheng Wang}, {and} \bibinfo{person}{Meng Li}.} \bibinfo{year}{2025}\natexlab{}.
\newblock \showarticletitle{Orchestrating Dual-Boundaries: An Arithmetic Intensity Inspired Acceleration Framework for Diffusion Language Models}.
\newblock \bibinfo{journal}{\emph{arXiv preprint arXiv:2511.21759}} (\bibinfo{year}{2025}).
\newblock


\bibitem[Wen et~al\mbox{.}(2025a)]%
        {wen2025token}
\bibfield{author}{\bibinfo{person}{Zichen Wen}, \bibinfo{person}{Yifeng Gao}, \bibinfo{person}{Weijia Li}, \bibinfo{person}{Conghui He}, {and} \bibinfo{person}{Linfeng Zhang}.} \bibinfo{year}{2025}\natexlab{a}.
\newblock \showarticletitle{Token pruning in multimodal large language models: Are we solving the right problem?}. In \bibinfo{booktitle}{\emph{Findings of the Association for Computational Linguistics: ACL 2025}}. \bibinfo{pages}{15537--15549}.
\newblock


\bibitem[Wen et~al\mbox{.}(2025b)]%
        {wen2025stop}
\bibfield{author}{\bibinfo{person}{Zichen Wen}, \bibinfo{person}{Yifeng Gao}, \bibinfo{person}{Shaobo Wang}, \bibinfo{person}{Junyuan Zhang}, \bibinfo{person}{Qintong Zhang}, \bibinfo{person}{Weijia Li}, \bibinfo{person}{Conghui He}, {and} \bibinfo{person}{Linfeng Zhang}.} \bibinfo{year}{2025}\natexlab{b}.
\newblock \showarticletitle{Stop Looking for “Important Tokens” in Multimodal Language Models: Duplication Matters More}. In \bibinfo{booktitle}{\emph{Proceedings of the 2025 Conference on Empirical Methods in Natural Language Processing}}. \bibinfo{pages}{9972--9991}.
\newblock


\bibitem[Wu et~al\mbox{.}(2025)]%
        {wu2025fast}
\bibfield{author}{\bibinfo{person}{Chengyue Wu}, \bibinfo{person}{Hao Zhang}, \bibinfo{person}{Shuchen Xue}, \bibinfo{person}{Zhijian Liu}, \bibinfo{person}{Shizhe Diao}, \bibinfo{person}{Ligeng Zhu}, \bibinfo{person}{Ping Luo}, \bibinfo{person}{Song Han}, {and} \bibinfo{person}{Enze Xie}.} \bibinfo{year}{2025}\natexlab{}.
\newblock \showarticletitle{Fast-dllm: Training-free acceleration of diffusion llm by enabling kv cache and parallel decoding}.
\newblock \bibinfo{journal}{\emph{arXiv preprint arXiv:2505.22618}} (\bibinfo{year}{2025}).
\newblock


\bibitem[Wu et~al\mbox{.}(2026)]%
        {wu2026hidrop}
\bibfield{author}{\bibinfo{person}{Hao Wu}, \bibinfo{person}{Yingqi Fan}, \bibinfo{person}{Jinyang Dai}, \bibinfo{person}{Junlong Tong}, \bibinfo{person}{Yunpu Ma}, {and} \bibinfo{person}{Xiaoyu Shen}.} \bibinfo{year}{2026}\natexlab{}.
\newblock \showarticletitle{HiDrop: Hierarchical Vision Token Reduction in MLLMs via Late Injection, Concave Pyramid Pruning, and Early Exit}.
\newblock \bibinfo{journal}{\emph{arXiv preprint arXiv:2602.23699}} (\bibinfo{year}{2026}).
\newblock


\bibitem[Wu et~al\mbox{.}(2024)]%
        {wu2024accelerating}
\bibfield{author}{\bibinfo{person}{Qiong Wu}, \bibinfo{person}{Wenhao Lin}, \bibinfo{person}{Yiyi Zhou}, \bibinfo{person}{Weihao Ye}, \bibinfo{person}{Zhanpeng Zen}, \bibinfo{person}{Xiaoshuai Sun}, {and} \bibinfo{person}{Rongrong Ji}.} \bibinfo{year}{2024}\natexlab{}.
\newblock \showarticletitle{Accelerating multimodal large language models via dynamic visual-token exit and the empirical findings}.
\newblock \bibinfo{journal}{\emph{arXiv preprint arXiv:2411.19628}} (\bibinfo{year}{2024}).
\newblock


\bibitem[Xin et~al\mbox{.}(2025)]%
        {xin2025lumina}
\bibfield{author}{\bibinfo{person}{Yi Xin}, \bibinfo{person}{Qi Qin}, \bibinfo{person}{Siqi Luo}, \bibinfo{person}{Kaiwen Zhu}, \bibinfo{person}{Juncheng Yan}, \bibinfo{person}{Yan Tai}, \bibinfo{person}{Jiayi Lei}, \bibinfo{person}{Yuewen Cao}, \bibinfo{person}{Keqi Wang}, \bibinfo{person}{Yibin Wang}, {et~al\mbox{.}}} \bibinfo{year}{2025}\natexlab{}.
\newblock \showarticletitle{Lumina-dimoo: An omni diffusion large language model for multi-modal generation and understanding}.
\newblock \bibinfo{journal}{\emph{arXiv preprint arXiv:2510.06308}} (\bibinfo{year}{2025}).
\newblock


\bibitem[Xu and Yang(2025)]%
        {xu2025dllmquant}
\bibfield{author}{\bibinfo{person}{Chen Xu} {and} \bibinfo{person}{Dawei Yang}.} \bibinfo{year}{2025}\natexlab{}.
\newblock \showarticletitle{Dllmquant: Quantizing diffusion-based large language models}.
\newblock \bibinfo{journal}{\emph{arXiv preprint arXiv:2508.14090}} (\bibinfo{year}{2025}).
\newblock


\bibitem[Xu et~al\mbox{.}(2025a)]%
        {xu2025redvtp}
\bibfield{author}{\bibinfo{person}{Jingqi Xu}, \bibinfo{person}{Jingxi Lu}, \bibinfo{person}{Chenghao Li}, \bibinfo{person}{Sreetama Sarkar}, \bibinfo{person}{Souvik Kundu}, {and} \bibinfo{person}{Peter~A Beerel}.} \bibinfo{year}{2025}\natexlab{a}.
\newblock \showarticletitle{RedVTP: Training-Free Acceleration of Diffusion Vision-Language Models Inference via Masked Token-Guided Visual Token Pruning}.
\newblock \bibinfo{journal}{\emph{arXiv preprint arXiv:2511.12428}} (\bibinfo{year}{2025}).
\newblock


\bibitem[Xu et~al\mbox{.}(2025b)]%
        {xu2025learning}
\bibfield{author}{\bibinfo{person}{Zhuoyan Xu}, \bibinfo{person}{Khoi~Duc Nguyen}, \bibinfo{person}{Preeti Mukherjee}, \bibinfo{person}{Saurabh Bagchi}, \bibinfo{person}{Somali Chaterji}, \bibinfo{person}{Yingyu Liang}, {and} \bibinfo{person}{Yin Li}.} \bibinfo{year}{2025}\natexlab{b}.
\newblock \showarticletitle{Learning to inference adaptively for multimodal large language models}. In \bibinfo{booktitle}{\emph{Proceedings of the IEEE/CVF International Conference on Computer Vision}}. \bibinfo{pages}{3552--3563}.
\newblock


\bibitem[Yang et~al\mbox{.}(2025b)]%
        {yang2025pvc}
\bibfield{author}{\bibinfo{person}{Chenyu Yang}, \bibinfo{person}{Xuan Dong}, \bibinfo{person}{Xizhou Zhu}, \bibinfo{person}{Weijie Su}, \bibinfo{person}{Jiahao Wang}, \bibinfo{person}{Hao Tian}, \bibinfo{person}{Zhe Chen}, \bibinfo{person}{Wenhai Wang}, \bibinfo{person}{Lewei Lu}, {and} \bibinfo{person}{Jifeng Dai}.} \bibinfo{year}{2025}\natexlab{b}.
\newblock \showarticletitle{PVC: Progressive Visual Token Compression for Unified Image and Video Processing in Large Vision-Language Models}. In \bibinfo{booktitle}{\emph{Proceedings of the Computer Vision and Pattern Recognition Conference}}. \bibinfo{pages}{24939--24949}.
\newblock


\bibitem[Yang et~al\mbox{.}(2025c)]%
        {yang2025mmada}
\bibfield{author}{\bibinfo{person}{Ling Yang}, \bibinfo{person}{Ye Tian}, \bibinfo{person}{Bowen Li}, \bibinfo{person}{Xinchen Zhang}, \bibinfo{person}{Ke Shen}, \bibinfo{person}{Yunhai Tong}, {and} \bibinfo{person}{Mengdi Wang}.} \bibinfo{year}{2025}\natexlab{c}.
\newblock \showarticletitle{Mmada: Multimodal large diffusion language models}.
\newblock \bibinfo{journal}{\emph{arXiv preprint arXiv:2505.15809}} (\bibinfo{year}{2025}).
\newblock


\bibitem[Yang et~al\mbox{.}(2025a)]%
        {yang2025visionzip}
\bibfield{author}{\bibinfo{person}{Senqiao Yang}, \bibinfo{person}{Yukang Chen}, \bibinfo{person}{Zhuotao Tian}, \bibinfo{person}{Chengyao Wang}, \bibinfo{person}{Jingyao Li}, \bibinfo{person}{Bei Yu}, {and} \bibinfo{person}{Jiaya Jia}.} \bibinfo{year}{2025}\natexlab{a}.
\newblock \showarticletitle{Visionzip: Longer is better but not necessary in vision language models}. In \bibinfo{booktitle}{\emph{Proceedings of the IEEE/CVF Conference on Computer Vision and Pattern Recognition}}. \bibinfo{pages}{19792--19802}.
\newblock


\bibitem[Yang et~al\mbox{.}(2025d)]%
        {yang2025diffusion}
\bibfield{author}{\bibinfo{person}{Yicun Yang}, \bibinfo{person}{Cong Wang}, \bibinfo{person}{Shaobo Wang}, \bibinfo{person}{Zichen Wen}, \bibinfo{person}{Biqing Qi}, \bibinfo{person}{Hanlin Xu}, {and} \bibinfo{person}{Linfeng Zhang}.} \bibinfo{year}{2025}\natexlab{d}.
\newblock \showarticletitle{Diffusion llm with native variable generation lengths: Let [eos] lead the way}.
\newblock \bibinfo{journal}{\emph{arXiv preprint arXiv:2510.24605}} (\bibinfo{year}{2025}).
\newblock


\bibitem[Yao et~al\mbox{.}(2024)]%
        {yao2024minicpm}
\bibfield{author}{\bibinfo{person}{Yuan Yao}, \bibinfo{person}{Tianyu Yu}, \bibinfo{person}{Ao Zhang}, \bibinfo{person}{Chongyi Wang}, \bibinfo{person}{Junbo Cui}, \bibinfo{person}{Hongji Zhu}, \bibinfo{person}{Tianchi Cai}, \bibinfo{person}{Haoyu Li}, \bibinfo{person}{Weilin Zhao}, \bibinfo{person}{Zhihui He}, {et~al\mbox{.}}} \bibinfo{year}{2024}\natexlab{}.
\newblock \showarticletitle{Minicpm-v: A gpt-4v level mllm on your phone}.
\newblock \bibinfo{journal}{\emph{arXiv preprint arXiv:2408.01800}} (\bibinfo{year}{2024}).
\newblock


\bibitem[Ye et~al\mbox{.}(2025b)]%
        {ye2025dream}
\bibfield{author}{\bibinfo{person}{Jiacheng Ye}, \bibinfo{person}{Zhihui Xie}, \bibinfo{person}{Lin Zheng}, \bibinfo{person}{Jiahui Gao}, \bibinfo{person}{Zirui Wu}, \bibinfo{person}{Xin Jiang}, \bibinfo{person}{Zhenguo Li}, {and} \bibinfo{person}{Lingpeng Kong}.} \bibinfo{year}{2025}\natexlab{b}.
\newblock \showarticletitle{Dream 7b: Diffusion large language models}.
\newblock \bibinfo{journal}{\emph{arXiv preprint arXiv:2508.15487}} (\bibinfo{year}{2025}).
\newblock


\bibitem[Ye et~al\mbox{.}(2025a)]%
        {ye2025fit}
\bibfield{author}{\bibinfo{person}{Weihao Ye}, \bibinfo{person}{Qiong Wu}, \bibinfo{person}{Wenhao Lin}, {and} \bibinfo{person}{Yiyi Zhou}.} \bibinfo{year}{2025}\natexlab{a}.
\newblock \showarticletitle{Fit and prune: Fast and training-free visual token pruning for multi-modal large language models}. In \bibinfo{booktitle}{\emph{Proceedings of the AAAI Conference on Artificial Intelligence}}, Vol.~\bibinfo{volume}{39}. \bibinfo{pages}{22128--22136}.
\newblock


\bibitem[Ye et~al\mbox{.}({[n.\,d.]})]%
        {ye2412atp}
\bibfield{author}{\bibinfo{person}{X Ye}, \bibinfo{person}{Y Gan}, \bibinfo{person}{Y Ge}, \bibinfo{person}{XP Zhang}, {and} \bibinfo{person}{Y Tang}.} \bibinfo{year}{[n.\,d.]}\natexlab{}.
\newblock \showarticletitle{ATP-LLAVA: adaptive token pruning for large vision language models (2024)}.
\newblock \bibinfo{journal}{\emph{URL https://arxiv. org/abs/2412}}  \bibinfo{volume}{447} (\bibinfo{year}{[n.\,d.]}).
\newblock


\bibitem[You et~al\mbox{.}(2025)]%
        {you2025llada}
\bibfield{author}{\bibinfo{person}{Zebin You}, \bibinfo{person}{Shen Nie}, \bibinfo{person}{Xiaolu Zhang}, \bibinfo{person}{Jun Hu}, \bibinfo{person}{Jun Zhou}, \bibinfo{person}{Zhiwu Lu}, \bibinfo{person}{Ji-Rong Wen}, {and} \bibinfo{person}{Chongxuan Li}.} \bibinfo{year}{2025}\natexlab{}.
\newblock \showarticletitle{Llada-v: Large language diffusion models with visual instruction tuning}.
\newblock \bibinfo{journal}{\emph{arXiv preprint arXiv:2505.16933}} (\bibinfo{year}{2025}).
\newblock


\bibitem[Yu et~al\mbox{.}({[n.\,d.]})]%
        {yu2505dimple}
\bibfield{author}{\bibinfo{person}{Runpeng Yu}, \bibinfo{person}{Xinyin Ma}, {and} \bibinfo{person}{Xinchao Wang}.} \bibinfo{year}{[n.\,d.]}\natexlab{}.
\newblock \showarticletitle{Dimple: Discrete diffusion multimodal large language model with parallel decoding, 2025}.
\newblock \bibinfo{journal}{\emph{URL https://arxiv. org/abs/2505.16990}} (\bibinfo{year}{[n.\,d.]}).
\newblock


\bibitem[Yue et~al\mbox{.}(2024)]%
        {yue2024mmmu}
\bibfield{author}{\bibinfo{person}{Xiang Yue}, \bibinfo{person}{Yuansheng Ni}, \bibinfo{person}{Kai Zhang}, \bibinfo{person}{Tianyu Zheng}, \bibinfo{person}{Ruoqi Liu}, \bibinfo{person}{Ge Zhang}, \bibinfo{person}{Samuel Stevens}, \bibinfo{person}{Dongfu Jiang}, \bibinfo{person}{Weiming Ren}, \bibinfo{person}{Yuxuan Sun}, {et~al\mbox{.}}} \bibinfo{year}{2024}\natexlab{}.
\newblock \showarticletitle{Mmmu: A massive multi-discipline multimodal understanding and reasoning benchmark for expert agi}. In \bibinfo{booktitle}{\emph{Proceedings of the IEEE/CVF conference on computer vision and pattern recognition}}. \bibinfo{pages}{9556--9567}.
\newblock


\bibitem[Zeng et~al\mbox{.}(2025)]%
        {zeng2025skip}
\bibfield{author}{\bibinfo{person}{Weili Zeng}, \bibinfo{person}{Ziyuan Huang}, \bibinfo{person}{Kaixiang Ji}, {and} \bibinfo{person}{Yichao Yan}.} \bibinfo{year}{2025}\natexlab{}.
\newblock \showarticletitle{Skip-vision: Efficient and scalable acceleration of vision-language models via adaptive token skipping}. In \bibinfo{booktitle}{\emph{Proceedings of the IEEE/CVF International Conference on Computer Vision}}. \bibinfo{pages}{21384--21397}.
\newblock


\bibitem[Zhang et~al\mbox{.}(2025c)]%
        {zhang2025lmms}
\bibfield{author}{\bibinfo{person}{Kaichen Zhang}, \bibinfo{person}{Bo Li}, \bibinfo{person}{Peiyuan Zhang}, \bibinfo{person}{Fanyi Pu}, \bibinfo{person}{Joshua~Adrian Cahyono}, \bibinfo{person}{Kairui Hu}, \bibinfo{person}{Shuai Liu}, \bibinfo{person}{Yuanhan Zhang}, \bibinfo{person}{Jingkang Yang}, \bibinfo{person}{Chunyuan Li}, {et~al\mbox{.}}} \bibinfo{year}{2025}\natexlab{c}.
\newblock \showarticletitle{Lmms-eval: Reality check on the evaluation of large multimodal models}. In \bibinfo{booktitle}{\emph{Findings of the Association for Computational Linguistics: NAACL 2025}}. \bibinfo{pages}{881--916}.
\newblock


\bibitem[Zhang et~al\mbox{.}(2025a)]%
        {zhang2025beyond}
\bibfield{author}{\bibinfo{person}{Qizhe Zhang}, \bibinfo{person}{Aosong Cheng}, \bibinfo{person}{Ming Lu}, \bibinfo{person}{Renrui Zhang}, \bibinfo{person}{Zhiyong Zhuo}, \bibinfo{person}{Jiajun Cao}, \bibinfo{person}{Shaobo Guo}, \bibinfo{person}{Qi She}, {and} \bibinfo{person}{Shanghang Zhang}.} \bibinfo{year}{2025}\natexlab{a}.
\newblock \showarticletitle{Beyond text-visual attention: Exploiting visual cues for effective token pruning in vlms}. In \bibinfo{booktitle}{\emph{Proceedings of the IEEE/CVF International Conference on Computer Vision}}. \bibinfo{pages}{20857--20867}.
\newblock


\bibitem[Zhang et~al\mbox{.}(2025b)]%
        {zhang2025quant}
\bibfield{author}{\bibinfo{person}{Tianao Zhang}, \bibinfo{person}{Zhiteng Li}, \bibinfo{person}{Xianglong Yan}, \bibinfo{person}{Haotong Qin}, \bibinfo{person}{Yong Guo}, {and} \bibinfo{person}{Yulun Zhang}.} \bibinfo{year}{2025}\natexlab{b}.
\newblock \showarticletitle{Quant-dllm: Post-training extreme low-bit quantization for diffusion large language models}.
\newblock \bibinfo{journal}{\emph{arXiv preprint arXiv:2510.03274}} (\bibinfo{year}{2025}).
\newblock


\bibitem[Zhang et~al\mbox{.}(2024)]%
        {zhang2024sparsevlm}
\bibfield{author}{\bibinfo{person}{Yuan Zhang}, \bibinfo{person}{Chun-Kai Fan}, \bibinfo{person}{Junpeng Ma}, \bibinfo{person}{Wenzhao Zheng}, \bibinfo{person}{Tao Huang}, \bibinfo{person}{Kuan Cheng}, \bibinfo{person}{Denis Gudovskiy}, \bibinfo{person}{Tomoyuki Okuno}, \bibinfo{person}{Yohei Nakata}, \bibinfo{person}{Kurt Keutzer}, {et~al\mbox{.}}} \bibinfo{year}{2024}\natexlab{}.
\newblock \showarticletitle{Sparsevlm: Visual token sparsification for efficient vision-language model inference}.
\newblock \bibinfo{journal}{\emph{arXiv preprint arXiv:2410.04417}} (\bibinfo{year}{2024}).
\newblock


\bibitem[Zhao et~al\mbox{.}(2025)]%
        {zhao2025accelerating}
\bibfield{author}{\bibinfo{person}{Shiyu Zhao}, \bibinfo{person}{Zhenting Wang}, \bibinfo{person}{Felix Juefei-Xu}, \bibinfo{person}{Xide Xia}, \bibinfo{person}{Miao Liu}, \bibinfo{person}{Xiaofang Wang}, \bibinfo{person}{Mingfu Liang}, \bibinfo{person}{Ning Zhang}, \bibinfo{person}{Dimitris~N Metaxas}, {and} \bibinfo{person}{Licheng Yu}.} \bibinfo{year}{2025}\natexlab{}.
\newblock \showarticletitle{Accelerating multimodal large language models by searching optimal vision token reduction}. In \bibinfo{booktitle}{\emph{Proceedings of the Computer Vision and Pattern Recognition Conference}}. \bibinfo{pages}{29869--29879}.
\newblock


\bibitem[Zhong et~al\mbox{.}(2025)]%
        {zhong2025aim}
\bibfield{author}{\bibinfo{person}{Yiwu Zhong}, \bibinfo{person}{Zhuoming Liu}, \bibinfo{person}{Yin Li}, {and} \bibinfo{person}{Liwei Wang}.} \bibinfo{year}{2025}\natexlab{}.
\newblock \showarticletitle{Aim: Adaptive inference of multi-modal llms via token merging and pruning}. In \bibinfo{booktitle}{\emph{Proceedings of the IEEE/CVF International Conference on Computer Vision}}. \bibinfo{pages}{20180--20192}.
\newblock


\bibitem[Zhu et~al\mbox{.}(2026)]%
        {zhu2026dllm}
\bibfield{author}{\bibinfo{person}{Zijian Zhu}, \bibinfo{person}{Fei Ren}, \bibinfo{person}{Zhanhong Tan}, {and} \bibinfo{person}{Kaisheng Ma}.} \bibinfo{year}{2026}\natexlab{}.
\newblock \showarticletitle{ES-dLLM: Efficient Inference for Diffusion Large Language Models by Early-Skipping}.
\newblock \bibinfo{journal}{\emph{arXiv preprint arXiv:2603.10088}} (\bibinfo{year}{2026}).
\newblock


\bibitem[Zou et~al\mbox{.}(2025)]%
        {zou2025don}
\bibfield{author}{\bibinfo{person}{Xin Zou}, \bibinfo{person}{Di Lu}, \bibinfo{person}{Yizhou Wang}, \bibinfo{person}{Yibo Yan}, \bibinfo{person}{Yuanhuiyi Lyu}, \bibinfo{person}{Xu Zheng}, \bibinfo{person}{Linfeng Zhang}, {and} \bibinfo{person}{Xuming Hu}.} \bibinfo{year}{2025}\natexlab{}.
\newblock \showarticletitle{Don't Just Chase" Highlighted Tokens" in MLLMs: Revisiting Visual Holistic Context Retention}.
\newblock \bibinfo{journal}{\emph{arXiv preprint arXiv:2510.02912}} (\bibinfo{year}{2025}).
\newblock


\end{thebibliography}

\clearpage
\appendix

\section{Experimental Details}

\subsection{Hardware Setup and Evaluation Stability}
All experiments were conducted on NVIDIA A100 GPUs. Specifically, the experiments for evaluating the ``Effectiveness of system-level optimizations'' were performed on 4$\times$ A100 GPUs to simulate high-concurrency batched serving, while all other baseline and ablation experiments were executed on a single A100 GPU. 

To ensure the reliability and reproducibility of our results, all experiments were run three times, and the average values are reported. The maximum variance observed across all evaluation metrics was strictly bounded:
\begin{itemize}
    \item \textbf{Accuracy:} The maximum variance was 0.0025, observed when evaluating the DivPrune method on the MMaDA model using the DocVQA benchmark.
    \item \textbf{Throughput:} The maximum variance was 0.0759 tokens/s, observed when evaluating the Seer method on the MMaDA model using the MME benchmark.
    \item \textbf{Latency:} The maximum variance was 0.1164 s, observed when evaluating the D3TOM method on the LaViDa-LLaDA model using the MathVista benchmark.
\end{itemize}

\subsection{Hyperparameter Configurations}
After careful hyperparameter tuning, the specific configurations for the baseline methods across different model architectures are detailed below:

\vspace{0.5em}
\noindent\textbf{D3TOM:} Adaptive merging was enabled. Token merging was strictly performed at the 3rd layer with a base merge rate set to 0.8.

\vspace{0.5em}
\noindent\textbf{RedVTP:} The visual token retention rate was set to 0.25.

\vspace{0.5em}
{\sloppy
\noindent\textbf{VisionZip:} For the LaViDa-LLaDA and LaViDa-Dream architectures, VISIONZIP\_DOMINANT was set to 80, and VISIONZIP\_CONTEXTUAL was set to 10. For the MMaDA model, these parameters were adjusted to 32 and 16, respectively. Note that because MMaDA utilizes MAGVIT-v2 VQ tokens and lacks continuous attention mechanisms, we adapted the VisionZip implementation to use text-visual cosine similarity for token filtering.\par}

\vspace{0.5em}
{\sloppy
\noindent\textbf{MMTok:} For LaViDa-LLaDA and LaViDa-Dream, the parameters were configured as MMTOK\_TOKENS = 64, MMTOK\_ALPHA = 0.5, MMTOK\_TV\_TEMP = 0.02, and MMTOK\_VV\_TEMP = 0.2. For MMaDA, MMTOK\_TOKENS was reduced to 32, while the remaining temperature and alpha parameters were kept identical.\par}

\vspace{0.5em}
{\sloppy
\noindent\textbf{DivPrune:} The token retention rate was set to 0.1 for LaViDa-LLaDA and LaViDa-Dream. Because MMaDA's visual encoder (MAGVIT-v2 VQ tokenizer) operates on a discrete codebook, we adapted the pruning strategy to use an absolute number to retain tokens, setting the retention count to exactly 32.\par}

\vspace{0.5em}
\noindent\textbf{SparseVLM:} The absolute token retention count was set to 128 for LaViDa-LLaDA and LaViDa-Dream, and 64 for MMaDA. Pruning was executed at layers 2, 6, and 15 across all models.

\vspace{0.5em}
{\sloppy
\noindent\textbf{Focus:} Focus is an acceleration method originally designed for pure diffusion language models, falling into the category of dynamic per-step token eviction. To evaluate its cross-modal effectiveness, we migrated this method to the diffusion multimodal large language model (DMLLM) architectures. In our experiments, we kept its default parameters.\par}

\vspace{0.5em}
{\sloppy
\noindent\textbf{Daedal:} Similarly, Daedal is a variable-length generation strategy (dynamic token eviction) originally proposed for diffusion language models, which we adapted for DMLLMs in our comparative study. Its parameters were configured as DAEDAL\_INITIAL\_LENGTH = 16, DAEDAL\_MAX\_LENGTH = 100, and DAEDAL\_EXPANSION\_FACTOR = 8.\par}

\subsection{Evaluation Settings}
The generation configurations for each benchmark strictly followed the default lmms-eval settings. Table \ref{tab:output_length} details the predefined denoising timestep $T$, which directly corresponds to the padded output sequence length $|O|$, used for each dataset during inference.

\begin{table}[htbp]
\centering
\caption{Output length settings to produce the main results. By default, we use the denoising timestep $T$ corresponding to the valid output window $|O|$.}
\label{tab:output_length}
\begin{tabular}{lc|lc}
\toprule
\textbf{Dataset} & $\mathbf{|O|}$ & \textbf{Dataset} & $\mathbf{|O|}$ \\
\midrule
MME & 16 & DocVQA & 32 \\
MMMU & 16 & InfoVQA & 32 \\
ChartQA & 16 & MathVista & 100 \\
SQA & 16 & MMBench & 100 \\
GQA & 16 & & \\
\bottomrule
\end{tabular}
\end{table}

\begin{table*}[!ht]
\centering
\caption{Additional comparative benchmark results on LaViDa-LLaDA and MMaDA architectures. Benchmark abbreviations: MU (MMMU), MB (MMBench), IV (InfoVQA), CQ (ChartQA), SQ (SQA), DV (DocVQA), GQ (GQA), MV (MathVista). The best results are highlighted in \textbf{bold}, the second best are \underline{underlined}, and the third best are marked in \textcolor{blue}{blue}.}
\label{5_comparsion}
\vspace{-3mm}

{
\begin{tabular}{l ccccccccc}
\toprule
Method & MME & MU & MB & IV & CQ & SQ & DV & GQ & MV \\
\midrule
\multicolumn{10}{c}{\textbf{LaViDa-LLaDA}} \\
\midrule
Baseline & \underline{1705.29} & \textbf{44.00} & \textbf{75.76} & \underline{38.11} & \underline{60.60} & \textcolor{blue}{72.38} & \underline{63.52} & \textcolor{blue}{56.70} & \underline{45.70} \\
Throughput $\uparrow$ & \textcolor{blue}{0.53} & \textbf{0.60} & \textcolor{blue}{0.12} & \textcolor{blue}{0.81} & \textcolor{blue}{1.50} & \textcolor{blue}{0.77} & \textcolor{blue}{1.16} & \textcolor{blue}{0.64} & \textcolor{blue}{0.16} \\
Latency $\downarrow$ & \textcolor{blue}{1.87} & \textcolor{blue}{1.88} & \textcolor{blue}{8.66} & \textcolor{blue}{3.32} & \textcolor{blue}{1.95} & \textcolor{blue}{1.30} & \textcolor{blue}{3.68} & \textcolor{blue}{1.95} & \textcolor{blue}{8.79} \\
\midrule
+ Focus & \textbf{1707.11} & \textbf{44.00} & \underline{75.00} & \textcolor{blue}{37.02} & \textbf{60.60} & \underline{72.48} & \textcolor{blue}{62.16} & \underline{56.74} & \textbf{46.00} \\
Throughput $\uparrow$ & 0.45 & 0.44 & 0.09 & 0.66 & 1.25 & 0.69 & \textbf{6.91} & 0.54 & 0.12 \\
Latency $\downarrow$ & 2.23 & 2.56 & 10.61 & 3.99 & 2.33 & 1.45 & 4.59 & 2.29 & 11.0634 \\
\midrule
+ Daedal & 1683.28 & \textcolor{blue}{43.21} & \textcolor{blue}{73.48} & \textcolor{blue}{37.43} & \textcolor{blue}{54.40} & 71.34 & 58.62 & 56.35 & {45.00} \\
Throughput $\uparrow$ & \underline{1.01} & \underline{1.32} & \underline{0.70} & \underline{2.57} & \textbf{3.35} & \underline{1.45} & \textcolor{blue}{3.48} & \underline{1.23} & \textbf{1.73} \\
Latency $\downarrow$ & \underline{0.99} & \underline{0.84} & \underline{1.43} & \underline{1.01} & \textbf{0.86} & \underline{0.69} & \underline{1.13} & \underline{1.00} & \underline{0.80} \\
\midrule
+ Seer (ours) & \textcolor{blue}{1697.94} & \underline{43.89} & \textbf{75.76} & \textbf{38.40} & \underline{57.80} & \textbf{72.73} & \textbf{63.66} & \textbf{56.75} & \textcolor{blue}{45.10} \\
Throughput $\uparrow$ & \textbf{1.11} & \textbf{1.39} & \textbf{0.93} & \textbf{2.76} & \underline{2.95} & \textbf{1.58} & \underline{4.40} & \textbf{1.36} & \underline{1.68} \\
Latency $\downarrow$ & \textbf{0.90} & \textbf{0.81} & \textbf{1.08} & \textbf{0.99} & \underline{0.98} & \textbf{0.65} & \textbf{0.97} & \textbf{0.92} & \textbf{0.75} \\
\midrule
\multicolumn{10}{c}{\textbf{MMaDA}} \\
\midrule
Baseline & \textcolor{blue}{1306.72} & \textbf{31.44} & \textbf{39.72} & \textbf{15.11} & \textcolor{blue}{9.92} & 56.82 & \textbf{10.30} & \textbf{50.79} & \textbf{58.00} \\
Throughput $\uparrow$ & \textcolor{blue}{0.29} & \textcolor{blue}{0.14} & \textcolor{blue}{0.28} & \textcolor{blue}{0.13} & \textcolor{blue}{1.37} & \textcolor{blue}{0.28} & \textcolor{blue}{0.15} & \textcolor{blue}{0.35} & \textcolor{blue}{0.59} \\
Latency $\downarrow$ & \textcolor{blue}{3.48} & \textcolor{blue}{7.72} & \textcolor{blue}{3.58} & \textcolor{blue}{14.94} & \textcolor{blue}{1.70} & \textcolor{blue}{3.54} & \textcolor{blue}{15.20} & \textcolor{blue}{3.41} & \textcolor{blue}{11.22} \\
\midrule
+ Focus & \underline{1312.10} & \textcolor{blue}{31.00} & \underline{39.70} & \textcolor{blue}{14.83} & \underline{10.04} & \textcolor{blue}{56.87} & \underline{10.01} & \textcolor{blue}{50.14} & \underline{57.70} \\
Throughput $\uparrow$ & 0.19 & 0.09 & 0.19 & 0.09 & 0.82 & 0.17 & 0.10 & 0.25 & 0.40 \\
Latency $\downarrow$ & 5.38 & 12.04 & 5.29 & 20.81 & 2.76 & 5.74 & 21.45 & 4.76 & 16.28 \\
\midrule
+ Daedal & 1289.24 & 30.78 & {38.78} & {14.75} & 9.80 & \underline{57.21} & {9.05} & {50.03} & \textcolor{blue}{57.50} \\
Throughput $\uparrow$ & \underline{2.34} & \underline{1.20} & \underline{2.55} & \underline{3.53} & \underline{2.49} & \textbf{2.53} & \underline{3.17} & \underline{2.52} & \textbf{6.74} \\
Latency $\downarrow$ & \underline{0.43} & \underline{0.91} & \underline{0.39} & \underline{0.56} & \underline{0.95} & \textbf{0.39} & \underline{0.73} & \underline{0.48} & \underline{0.67} \\
\midrule
+ Seer (ours) & \textbf{1315.69} & \underline{31.33} & \textcolor{blue}{39.16} & \underline{14.87} & \textbf{10.12} & \textbf{58.16} & \textcolor{blue}{9.27} & \underline{50.65} & 55.80 \\
Throughput $\uparrow$ & \textbf{3.39} & \textbf{2.33} & \textbf{3.17} & \textbf{4.02} & \textbf{3.95} & \underline{2.35} & \textbf{4.48} & \textbf{4.46} & \underline{3.39} \\
Latency $\downarrow$ & \textbf{0.30} & \textbf{0.46} & \textbf{0.32} & \textbf{0.49} & \textbf{0.59} & \underline{0.43} & \textbf{0.50} & \textbf{0.27} & \textbf{0.39} \\
\bottomrule
\end{tabular}
}
\end{table*}

\begin{table*}[!ht]
\centering
\caption{Comprehensive benchmark results on the LaViDa-Dream architecture. Benchmark abbreviations: MU (MMMU), MB (MMBench), IV (InfoVQA), CQ (ChartQA), SQ (SQA), DV (DocVQA), GQ (GQA), MV (MathVista). The best results are highlighted in \textbf{bold}, the second best are \underline{underlined}, and the third best are marked in \textcolor{blue}{blue}.}
\label{6_comparsion}
\vspace{-3mm}
\label{tab-comprehensive_exp_dream}
{
\begin{tabular}{l ccccccccc}
\toprule
Method & MME & MU & MB & IV & CQ & SQ & DV & GQ & MV \\
\midrule
\multicolumn{10}{c}{\textbf{LaViDa-Dream}} \\
\midrule
Baseline & 1822.04 & 42.44 & \textcolor{blue}{76.52} & \textcolor{blue}{40.61} & \textbf{55.40} & 75.21 & \textcolor{blue}{59.28} & {58.05} & 41.60 \\
Throughput $\uparrow$ & 2.54 & 2.53 & 0.20 & 2.04 & 4.09 & 3.97 & 2.25 & 2.67 & 0.46 \\
Latency $\downarrow$ & 1.18 & 1.24 & 7.16 & 2.30 & 1.19 & 0.76 & 2.70 & 1.23 & 7.24 \\
\midrule
+ D3ToM & 1564.48 & \underline{42.89} & 75.00 & 26.77 & 19.00 & 73.87 & 15.15 & 48.97 & 38.10 \\
Throughput $\uparrow$ & 4.09 & 3.73 & 0.63 & 3.22 & 5.79 & 4.98 & 3.35 & 4.31 & 0.72 \\
Latency $\downarrow$ & 0.73 & 0.83 & 4.77 & 1.36 & 0.75 & 0.60 & 1.47 & 0.75 & 4.61 \\
\midrule
+ RedVTP & 1547.32 & 42.33 & 68.94 & 30.12 & 25.20 & 72.24 & 34.38 & 50.94 & 37.60 \\
Throughput $\uparrow$ & 4.64 & 4.18 & 0.75 & 3.74 & 6.68 & 6.01 & 4.28 & 4.86 & 0.84 \\
Latency $\downarrow$ & 0.65 & 0.75 & 4.01 & 1.21 & 0.68 & 0.50 & 1.29 & 0.67 & 3.93 \\
\midrule
+ VisionZip & 1491.66 & 42.33 & 75.00 & 25.43 & 25.60 & 72.34 & 25.75 & 51.65 & 39.20 \\
Throughput $\uparrow$ & 4.25 & 4.23 & 0.64 & 3.12 & 6.22 & 5.32 & 3.49 & 4.61 & 0.70 \\
Latency $\downarrow$ & 0.71 & 0.74 & 4.71 & 1.40 & 0.73 & 0.56 & 1.43 & 0.71 & 4.75 \\
\midrule
+ MMTok & 1741.38 & \textbf{43.44} & \underline{77.27} & 33.64 & 34.40 & 75.16 & 38.80 & 54.95 & 41.50 \\
Throughput $\uparrow$ & 4.20 & 4.29 & 0.68 & 3.40 & 6.46 & 5.39 & 3.90 & 4.51 & 0.75 \\
Latency $\downarrow$ & 0.71 & 0.73 & 4.42 & 1.35 & 0.74 & 0.56 & 1.43 & 0.73 & 4.45 \\
\midrule
+ DivPrune & 1113.88 & 41.11 & 49.24 & 24.41 & 15.60 & 71.15 & 9.67 & 39.20 & 37.90 \\
Throughput $\uparrow$ & \textcolor{blue}{8.87} & \textbf{6.53} & \textcolor{blue}{1.01} & \underline{5.97} & \textbf{11.66} & \textcolor{blue}{7.90} & \textcolor{blue}{6.24} & \underline{9.42} & \textcolor{blue}{1.19} \\
Latency $\downarrow$ & \textcolor{blue}{0.34} & \textbf{0.47} & \textcolor{blue}{2.98} & \underline{0.71} & \textbf{0.35} & \underline{0.38} & \textbf{0.68} & \textbf{0.34} & \textcolor{blue}{2.75} \\
\midrule
+ SparseVLM & 1424.32 & {42.11} & 72.74 & 24.97 & 27.60 & 73.62 & 19.98 & 48.93 & 40.90 \\
Throughput $\uparrow$ & 6.22 & 5.18 & 0.77 & 4.61 & 9.12 & 6.39 & 5.22 & 6.70 & 0.88 \\
Latency $\downarrow$ & 0.48 & 0.60 & 3.90 & 0.92 & 0.49 & 0.47 & 0.91 & 0.48 & 3.81 \\
\midrule
+ Focus & \underline{1835.20} & 42.56 & 75.76 & \textbf{41.70} & \textcolor{blue}{54.00} & \underline{75.51} & 59.20 & \textcolor{blue}{58.48} & \underline{42.20} \\
Throughput $\uparrow$ & 1.42 & 1.30 & 0.24 & 1.13 & 2.26 & 2.47 & 1.24 & 1.48 & 0.24 \\
Latency $\downarrow$ & 2.11 & 2.41 & 12.61 & 4.13 & 2.17 & 1.21 & 4.92 & 2.21 & 13.73 \\
\midrule
+ Daedal & \textcolor{blue}{1828.15} & 42.44 & \underline{77.27} & 40.22 & 49.40 & \textcolor{blue}{75.31} & \underline{59.34} & \underline{58.67} & \textcolor{blue}{42.10} \\
Throughput $\uparrow$ & \textbf{11.34} & \textcolor{blue}{5.73} & \underline{1.44} & \textcolor{blue}{5.22} & \textcolor{blue}{10.95} & \underline{8.71} & \textbf{8.40} & \textbf{9.49} & \underline{1.84} \\
Latency $\downarrow$ & \underline{0.26} & \underline{0.50} & \underline{2.24} & \textcolor{blue}{0.84} & \textcolor{blue}{0.45} & \textcolor{blue}{0.45} & \textcolor{blue}{0.77} & \underline{0.35} & \underline{1.83} \\
\midrule
+ Seer (ours) & \textbf{1852.05} & \textcolor{blue}{42.60} & \textbf{78.01} & \underline{41.12} & \underline{54.60} & \textbf{75.82} & \textbf{60.13} & \textbf{59.06} & \textbf{42.30} \\
Throughput $\uparrow$ & \underline{9.53} & \underline{5.99} & \textbf{1.58} & \textbf{7.85} & \underline{11.28} & \textbf{8.95} & \underline{8.30} & \textcolor{blue}{8.22} & \textbf{2.00} \\
Latency $\downarrow$ & \textbf{0.21} & \textcolor{blue}{0.52} & \textbf{1.90} & \textbf{0.59} & \underline{0.43} & \textbf{0.34} & \underline{0.73} & \textcolor{blue}{0.40} & \textbf{1.67} \\
\bottomrule
\end{tabular}
}
\end{table*}

\subsection{Datasets}
We list all evaluations on these multimodal (vision-language) understanding benchmarks below. 

\vspace{1em}
\noindent\textbf{Comprehensive reasoning}

\vspace{0.5em}
\noindent\textbf{MME.} The MME benchmark is designed to comprehensively assess multimodal large language models across a wide spectrum of perception and cognition tasks.

\vspace{0.5em}
\noindent\textbf{MMMU.} The MMMU benchmark is designed to access a model's proficiency in tasks requiring expert-level and domain-specific understanding and reasoning. It consists of over 10k questions derived from college-level examinations and tutorials across science, engineering and humanities. The questions challenge models on both advanced perception and deep knowledge. We evaluate all methods on the validation splits.

\vspace{0.5em}
\noindent\textbf{MMBench.} MMBench implements a multi-level framework to evaluate cross-specific task dimensions. The assessment is structured into different levels, enabling a comprehensive precise analysis of the strengths and weaknesses across skills. We evaluate methods' performance on dev splits.

\vspace{1em}
\noindent\textbf{Visual Question Answering}

\vspace{0.5em}
\noindent\textbf{ChartQA.} The ChartQA benchmark targets the complex challenge of question answering about charts, requiring models to integrate visual parsing with logical and mathematical reasoning.

\vspace{0.5em}
\noindent\textbf{SQA.} ScienceQA targets complex and multi-step reasoning within scientific domains. The benchmark features a dataset of questions across nature, language and social science. It is explicitly designed to evaluate multimodal understanding, sophisticated reasoning chains and interpretability. We evaluate methods on the image split to achieve visual ability.

\vspace{0.5em}
\noindent\textbf{GQA.} The GQA is composed of three parts: scene graphs, questions, and images. GQA offers a targeted evaluation methodology for visual question answering, grounded in detailed scene graph representations of images. The questions are generated to test a model's ability to reason about object attributes, the understanding of visual scenes and image spatial relationships. We evaluate all methods' performance on the GQA\_lite split.

\vspace{0.5em}
\noindent\textbf{DocVQA.} DocVQA focuses on visual question answering based on document images, pushing models to accurately parse dense text and structural elements within scanned documents.

\vspace{0.5em}
\noindent\textbf{InfoVQA.} InfoVQA evaluates a model's capability to comprehend infographics, demanding the joint interpretation of complex visual layouts, charts, and embedded textual information.

\vspace{0.5em}
\noindent\textbf{MathVista.} MathVista provides a fine-grained test bed for evaluating the mathematical reasoning of MLLMs in visual contexts, markedly widening the topical coverage and problem diversity. We conduct our evaluation on the test split.

\section{Extended Evaluation on Inference Efficiency}

\noindent\textbf{Analysis of Comprehensive Benchmarks.} As demonstrated in the comprehensive benchmark results across the LaViDa-LLaDA and MMaDA (Table~\ref{5_comparsion}), as well as the LaViDa-Dream (Table~\ref{6_comparsion}) architectures, our proposed Seer framework consistently achieves state-of-the-art efficiency. It delivers the highest throughput and lowest latency while rigorously preserving multimodal reasoning accuracy. While the baseline-adapted Daedal provides noticeable speedups, it often struggles to maintain optimal performance across all complex visual datasets. Most notably, Focus fails entirely to generalize to the multimodal diffusion paradigm; it actually exacerbates the computational bottleneck, resulting in strictly lower throughput and higher latency compared to the unaccelerated baseline in both Table 5 and Table 6, thereby offering no acceleration effect whatsoever. These findings underscore the superior adaptability and necessity of Seer's sparsity-aware truncation approach for practically optimizing DMLLMs.

\begin{figure*}[ht]
    \centering
    \includegraphics[width=\linewidth]{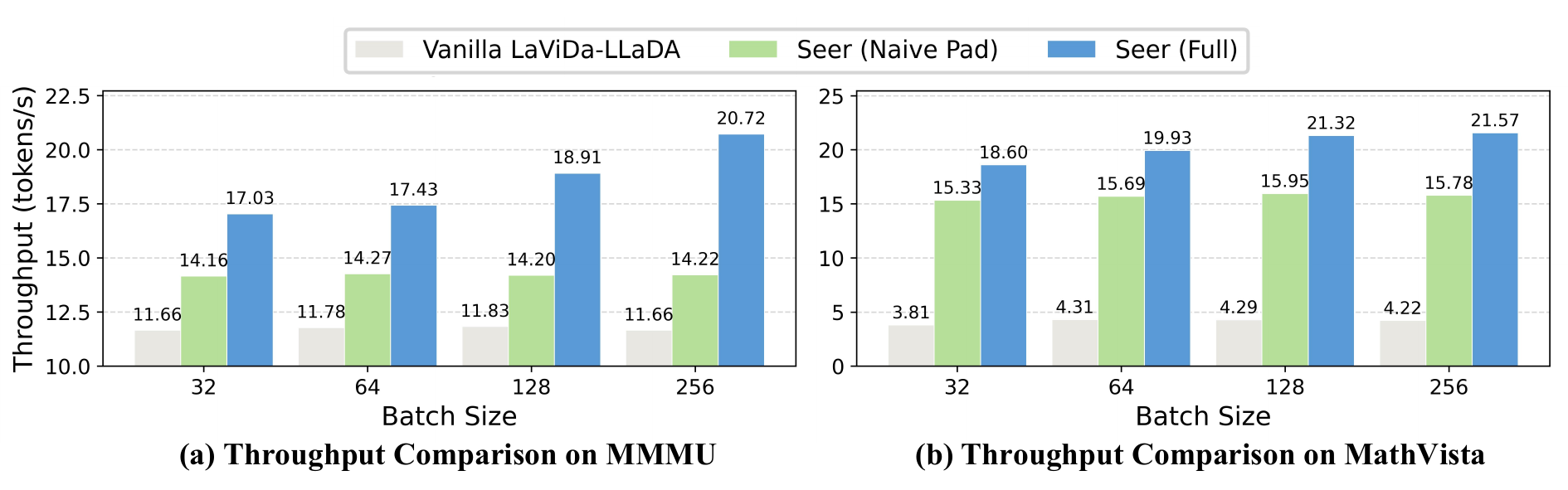} 
    \vspace{-8mm}
    \caption{\textbf{Throughput Scaling on LaViDa-LLaDA.} Impact of different execution strategies on inference throughput across varying batch sizes on the (a) MMMU and (b) MathVista datasets. Seer (Full) overcomes the padding waste bottleneck observed in Seer (Naive Pad), achieving significant and sustained throughput scaling at high concurrency.}
    \label{fig:throughput_llada}
\end{figure*}

\begin{figure*}[ht]
    \centering
    \includegraphics[width=\linewidth]{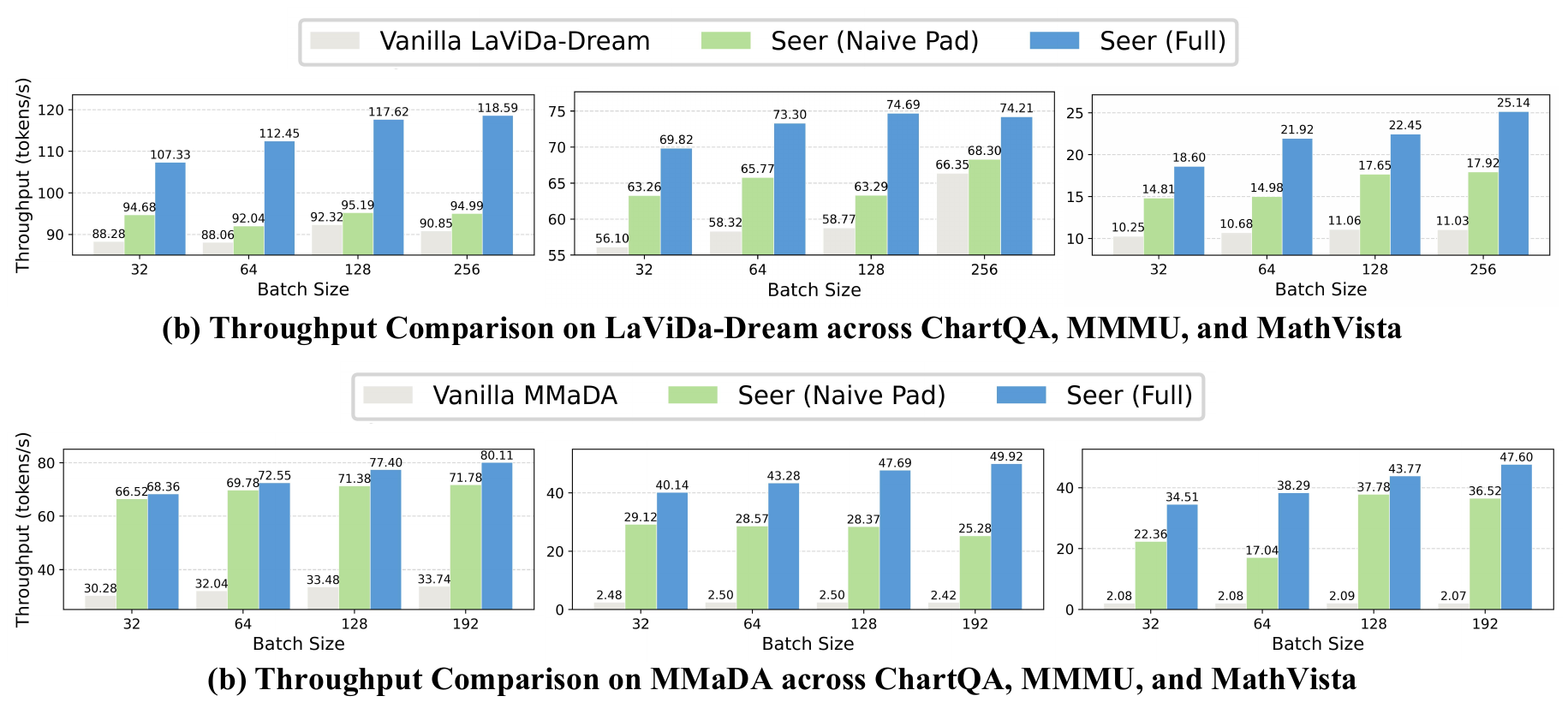} 
    \vspace{-8mm}
    \caption{\textbf{Throughput Scaling on LaViDa-Dream and MMaDA.} Impact of different execution strategies on inference throughput across varying batch sizes on the ChartQA, MMMU, and MathVista datasets for (a) LaViDa-Dream and (b) MMaDA. Consistent with LaViDa-LLaDA, Seer (Full) effectively translates theoretical reductions into robust end-to-end acceleration.}
    \label{fig:throughput_dream_mmada}
\end{figure*}


\textbf{Scaling Characteristics and System-Level Performance Gains.} 
We conducted comprehensive throughput comparison experiments across different concurrency scales on the LaViDa-LLaDA model (Figure~\ref{fig:throughput_llada}), as well as the LaViDa-Dream and MMaDA architectures (Figure~\ref{fig:throughput_dream_mmada}). Across all evaluated models and datasets, the results demonstrate that the complete Seer framework achieves significant and sustained acceleration as the batch size increases. For instance, Seer Full reaches peak throughputs of 21.57 tokens/s on LaViDa-LLaDA (MathVista, BS 256) and 118.59 tokens/s on LaViDa-Dream (ChartQA, BS 256). Notably, on the MMaDA architecture evaluated on MathVista, throughput increases from 2.07 tokens/s to 47.60 tokens/s at a batch size of 192, representing a 23$\times$ gain over the vanilla baseline. We also observe that when scaling to even larger batch sizes, the throughput saturates in compute-bound setups (e.g., LaViDa-LLaMA on MathVista); however, it continues to increase until out-of-memory (OOM) occurs in memory-bound setups (e.g., MMaDA at BS 256 in our setting). These experiments consistently reveal that while applying algorithmic truncation alone (Seer Naive Pad) yields marginal improvements, its throughput stagnates at larger batch sizes due to severe padding waste caused by intra-batch sequence length variance (i.e., the straggler effect). Conversely, Seer's proposed Hybrid Execution Routing effectively overcomes this system-level bottleneck by dynamically dispatching sequences into execution-friendly paths, successfully translating theoretical FLOP reductions into consistent end-to-end wall-clock acceleration across different DMLLM backbones.

\begin{table*}[!ht]
\centering
\caption{Compatibility of Seer with KV Cache. Benchmark results on LaViDa-LLaDA, LaViDa-Dream, and MMaDA architectures equipped with KV Cache. Benchmark abbreviations: MU (MMMU), MB (MMBench), IV (InfoVQA), CQ (ChartQA), SQ (SQA), DV (DocVQA), GQ (GQA), MV (MathVista). The better results between the baseline and Seer for each metric are highlighted in \textbf{bold}.}
\label{tab:kv_cache_compatibility}
\vspace{-3mm}
{
\begin{tabular}{l ccccccccc}
\toprule
Method & MME & MU & MB & IV & CQ & SQ & DV & GQ & MV \\
\midrule
\multicolumn{10}{c}{\textbf{LaViDa-LLaDA (w/ KV Cache)}} \\
\midrule
Baseline & 1694.92 & 43.78 & \textbf{73.48} & 36.58 & 57.40 & \textbf{71.24} & 58.67 & 56.10 & \textbf{44.90} \\
Throughput $\uparrow$ & 1.01 & 1.15 & 0.31 & 1.84 & 2.96 & 1.11 & 2.75 & 1.28 & 0.42 \\
Latency $\downarrow$ & 0.99 & 0.96 & 3.25 & 1.41 & 0.98 & 0.90 & 1.44 & 0.97 & 3.27 \\
\midrule
+ Seer (ours) & \textbf{1707.78} & \textbf{43.89} & \textbf{73.48} & \textbf{36.88} & \textbf{57.60} & 71.19 & \textbf{58.75} & \textbf{56.37} & 44.80 \\
Throughput $\uparrow$ & \textbf{1.43} & \textbf{1.63} & \textbf{1.55} & \textbf{3.66} & \textbf{4.14} & \textbf{1.73} & \textbf{5.22} & \textbf{1.89} & \textbf{1.89} \\
Latency $\downarrow$ & \textbf{0.70} & \textbf{0.65} & \textbf{0.64} & \textbf{0.71} & \textbf{0.70} & \textbf{0.58} & \textbf{0.76} & \textbf{0.67} & \textbf{0.64} \\

\midrule
\multicolumn{10}{c}{\textbf{LaViDa-Dream (w/ KV Cache)}} \\
\midrule
Baseline & 1856.96 & \textbf{42.00} & 75.76 & 41.17 & \textbf{53.40} & 75.31 & 56.32 & \textbf{57.80} & \textbf{41.80} \\
Throughput $\uparrow$ & 6.05 & 6.47 & 1.23 & 5.53 & 9.90 & 7.05 & 7.01 & 6.48 & 1.41 \\
Latency $\downarrow$ & 0.50 & 0.49 & 2.45 & 0.86 & 0.50 & 0.43 & 0.86 & 0.50 & 2.44 \\
\midrule
+ Seer (ours) & \textbf{1874.37} & 41.97 & \textbf{77.27} & \textbf{41.76} & 50.19 & \textbf{75.38} & \textbf{58.39} & 56.32 & 40.90 \\
Throughput $\uparrow$ & \textbf{23.62} & \textbf{13.85} & \textbf{3.35} & \textbf{28.13} & \textbf{24.69} & \textbf{14.92} & \textbf{19.19} & \textbf{14.19} & \textbf{4.45} \\
Latency $\downarrow$ & \textbf{0.13} & \textbf{0.23} & \textbf{0.90} & \textbf{0.17} & \textbf{0.20} & \textbf{0.20} & \textbf{0.32} & \textbf{0.23} & \textbf{0.77} \\

\midrule
\multicolumn{10}{c}{\textbf{MMaDA (w/ KV Cache)}} \\
\midrule
Baseline & 1306.95 & 31.00 & \textbf{37.69} & \textbf{12.35} & \textbf{9.56} & \textbf{53.89} & \textbf{8.85} & \textbf{47.38} & \textbf{51.70} \\
Throughput $\uparrow$ & 0.74 & 0.39 & 0.72 & 0.36 & 0.64 & 0.71 & 0.49 & 0.88 & 0.32 \\
Latency $\downarrow$ & 1.34 & 2.77 & 1.38 & 6.14 & 4.09 & 1.37 & 6.10 & 1.34 & 4.46 \\
\midrule
+ Seer (ours) & \textbf{1329.57} & \textbf{32.67} & 37.62 & 11.98 & 9.24 & 53.84 & 8.37 & 46.65 & \textbf{51.70} \\
Throughput $\uparrow$ & \textbf{5.75} & \textbf{5.09} & \textbf{4.80} & \textbf{8.83} & \textbf{10.08} & \textbf{4.43} & \textbf{9.65} & \textbf{6.56} & \textbf{6.05} \\
Latency $\downarrow$ & \textbf{0.17} & \textbf{0.21} & \textbf{0.21} & \textbf{0.23} & \textbf{0.25} & \textbf{0.22} & \textbf{0.27} & \textbf{0.18} & \textbf{0.21} \\
\bottomrule
\end{tabular}
}
\end{table*}

\section{Bounding-Box-Grounded Attention Analysis}
To provide quantitative support for the interpretation in Sec.~3.6, we analyze how early suffix truncation changes valid-prefix visual attention on a subset of 100 MME samples with explicit bounding-box annotations. These samples allow us to define question-relevant visual regions directly, enabling a region-aware comparison between the original model and Seer.

For each sample, let $V$ denote the set of visual tokens, and let $R \subseteq V$ denote the visual tokens whose image patches overlap with the ground-truth bounding box. Following common patch-region alignment practice, a visual token is assigned to $R$ if its patch overlaps with the bounding box by more than a threshold $\tau$ (we use $\tau=0.5$ in our implementation). The remaining visual tokens form the irrelevant region set $I = V \setminus R$.

Let $P$ denote the set of valid generated text tokens before the first \texttt{[EOS]} or \texttt{[EOT]}. For each valid prefix token $p \in P$, we extract its attention over visual tokens and renormalize the distribution within the visual modality:
\begin{equation}
\tilde{A}_{p \rightarrow v}
=
\frac{A_{p \rightarrow v}}
{\sum_{u \in V} A_{p \rightarrow u} + \epsilon},
\qquad v \in V,
\end{equation}
where $\epsilon$ is a small constant for numerical stability.

We then compute two complementary metrics. First, we use \emph{attention entropy} to measure how diffuse the valid-prefix visual attention is:
\begin{equation}
H_p
=
-\sum_{v \in V}
\tilde{A}_{p \rightarrow v}
\log(\tilde{A}_{p \rightarrow v} + \epsilon),
\end{equation}
and normalize it by $\log |V|$:
\begin{equation}
\hat{H}_p = \frac{H_p}{\log |V|}.
\end{equation}
Lower entropy indicates that the prefix attends to a more concentrated subset of visual tokens.

Second, we compute the \emph{irrelevant-region attention mass}, defined as the fraction of valid-prefix visual attention assigned to tokens outside the bounding box:
\begin{equation}
M^{\mathrm{irr}}_p
=
\sum_{v \in I}
\tilde{A}_{p \rightarrow v}.
\end{equation}
Lower values indicate weaker reliance on question-irrelevant visual context.

Unless otherwise specified, we report both metrics averaged over all valid prefix tokens, all attention heads, and the last four transformer layers at the final denoising step:
\begin{equation}
\hat{H}
=
\frac{1}{|P||\mathcal{L}||\mathcal{H}|}
\sum_{p \in P}
\sum_{l \in \mathcal{L}}
\sum_{h \in \mathcal{H}}
\hat{H}^{(l,h)}_p,
\end{equation}
\begin{equation}
M^{\mathrm{irr}}
=
\frac{1}{|P||\mathcal{L}||\mathcal{H}|}
\sum_{p \in P}
\sum_{l \in \mathcal{L}}
\sum_{h \in \mathcal{H}}
M^{\mathrm{irr},(l,h)}_p.
\end{equation}

\section{Compatibility with KV Cache}

\textbf{Orthogonality to Existing Caching Mechanisms.} To demonstrate that our Seer framework is orthogonal to existing attention-level optimizations, we conducted an extended evaluation integrating Seer with the standard KV Cache mechanism. We assessed the joint performance on the LaViDa-LLaDA, LaViDa-Dream, and MMaDA architectures across all nine reasoning benchmarks. As shown in Table~\ref{tab:kv_cache_compatibility}, even when the baseline models are inherently accelerated by KV caching, applying Seer still yields substantial and consistent improvements in both end-to-end throughput and latency. For instance, on the MMaDA model, Seer increases the actual throughput on the ChartQA dataset from 0.64 tokens/s to 10.08 tokens/s while reducing latency from 4.09s to 0.25s. Crucially, these significant speedups are achieved without compromising the multimodal reasoning scores across diverse tasks. This empirically confirms that Seer's dynamic sparsity-aware truncation provides independent, cumulative acceleration benefits that seamlessly complement and enhance foundational system-level optimizations like KV caching, proving its broad applicability as a plug-and-play solution.

\begin{table*}[!ht]
\centering

\begin{minipage}{0.48\textwidth}
\centering
\caption{Joint impact of Absolute Jump Threshold ($\tau_{jump}$) and Tolerance Ratio ($\gamma$) on generation accuracy and inference throughput (Evaluated on MMMU). The optimal Pareto frontier configuration is highlighted in bold.}
\label{tab:ablation_mmmu}

{\footnotesize
\setlength{\tabcolsep}{3.5pt}
\begin{tabular}{@{}ccccccc@{}}
\toprule
\textbf{Tol.} & \multicolumn{6}{c}{\textbf{Abs. Jump Threshold ($\tau_{jump}$)}} \\
\cmidrule(lr){2-7}
\textbf{($\gamma$)} & 0.01 & 0.02 & 0.03 & 0.04 & 0.05 & 0.06 \\
\midrule
0.2 & 43.9/0.82 & 44.0/0.78 & 44.0/0.75 & 44.0/0.71 & 44.0/0.67 & 44.0/0.64 \\
0.3 & 43.8/1.02 & 43.9/0.92 & 44.0/0.86 & 44.0/0.80 & 44.0/0.75 & 44.0/0.70 \\
0.4 & 42.1/1.35 & 43.6/1.18 & 43.9/1.02 & 44.0/0.92 & 44.0/0.85 & 44.0/0.78 \\
0.5 & 37.6/1.77 & 42.2/1.45 & 43.8/1.22 & 43.9/1.08 & 44.0/0.96 & 44.0/0.86 \\
0.6 & 28.5/2.21 & 38.6/1.72 & \textbf{43.9/1.39} & 43.8/1.21 & 43.9/1.05 & 44.0/0.92 \\
0.7 & 16.2/2.75 & 30.5/2.15 & 40.2/1.65 & 42.6/1.42 & 43.5/1.22 & 43.8/1.08 \\
0.8 & 8.5/3.12 & 21.4/2.65 & 33.5/2.05 & 39.2/1.72 & 42.1/1.45 & 43.5/1.25 \\
0.9 & 3.8/3.55 & 11.5/3.05 & 24.6/2.45 & 34.5/1.94 & 38.6/1.68 & 42.2/1.40 \\
1.0 & 1.5/3.87 & 6.2/3.45 & 15.5/2.85 & 28.5/2.35 & 34.2/1.95 & 39.5/1.62 \\
\bottomrule
\end{tabular}
}
\end{minipage}\hfill
\begin{minipage}{0.48\textwidth}
\centering
\caption{Joint impact of Absolute Jump Threshold ($\tau_{jump}$) and Tolerance Ratio ($\gamma$) on generation accuracy and inference throughput (Evaluated on MathVista). The optimal Pareto frontier configuration is highlighted in bold.}
\label{tab:ablation_mathvista}

{\footnotesize
\setlength{\tabcolsep}{3.5pt}
\begin{tabular}{@{}ccccccc@{}}
\toprule
\textbf{Tol.} & \multicolumn{6}{c}{\textbf{Abs. Jump Threshold ($\tau_{jump}$)}} \\
\cmidrule(lr){2-7}
\textbf{($\gamma$)} & 0.01 & 0.02 & 0.03 & 0.04 & 0.05 & 0.06 \\
\midrule
0.2 & 45.5/0.52 & 45.6/0.48 & 45.7/0.45 & 45.7/0.38 & 45.7/0.25 & 45.7/0.18 \\
0.3 & 44.8/0.85 & 45.5/0.72 & 45.6/0.65 & 45.7/0.52 & 45.7/0.40 & 45.7/0.30 \\
0.4 & 41.2/1.56 & 44.9/1.15 & 45.4/0.96 & 45.7/0.73 & 45.7/0.60 & 45.7/0.44 \\
0.5 & 35.6/2.20 & 42.6/1.65 & 45.3/1.25 & 45.6/1.08 & 45.7/0.82 & 45.7/0.62 \\
0.6 & 28.5/2.95 & 39.2/2.10 & \textbf{45.1/1.68} & 45.5/1.31 & 45.6/1.05 & 45.7/0.85 \\
0.7 & 15.2/3.66 & 32.5/2.76 & 41.5/2.11 & 43.2/1.75 & 45.1/1.40 & 45.5/1.12 \\
0.8 & 8.5/4.15 & 22.4/3.40 & 35.2/2.80 & 40.8/2.25 & 43.5/1.85 & 44.9/1.55 \\
0.9 & 4.1/4.50 & 12.5/4.02 & 26.5/3.35 & 36.4/2.70 & 40.2/2.30 & 43.2/1.97 \\
1.0 & 1.8/4.85 & 6.8/4.42 & 16.2/3.93 & 29.5/3.24 & 36.5/2.80 & 40.5/2.40 \\
\bottomrule
\end{tabular}
}
\end{minipage}

\end{table*}

\begin{figure*}[ht]
    \centering
    \includegraphics[width=0.8\linewidth]{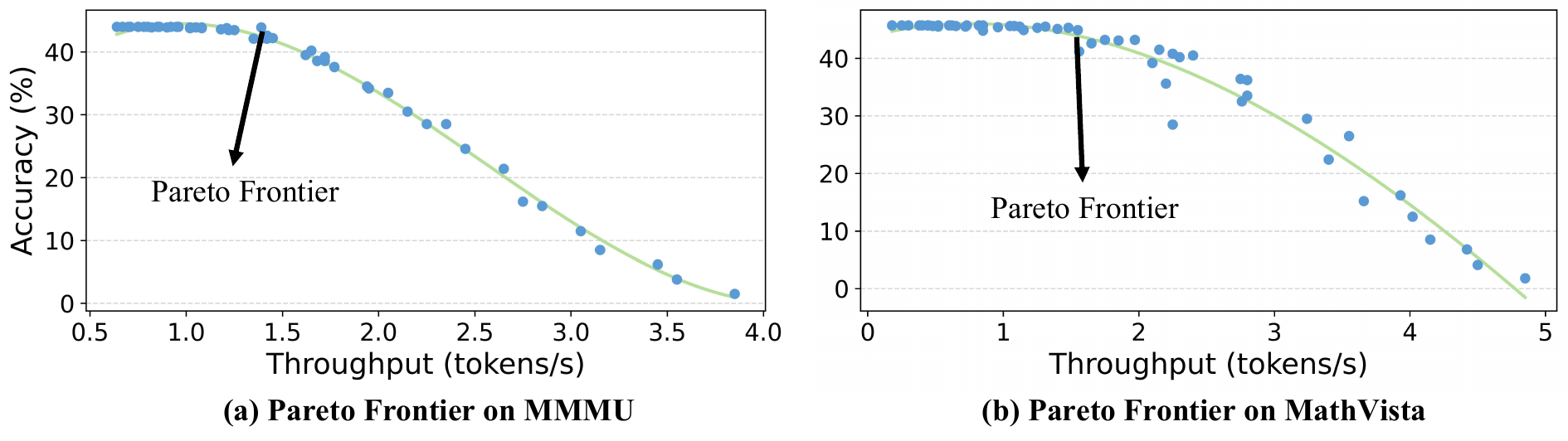} 
    \vspace{-4mm}
    \caption{\textbf{Pareto frontier of generation accuracy versus inference throughput.} (a) and (b) illustrate the trade-off on the MMMU and MathVista datasets, respectively. Each point represents a specific hyperparameter combination (Absolute Jump Threshold $\tau_{jump}$ and Tolerance Ratio $\gamma$) corresponding to Table \ref{tab:ablation_mmmu} and Table \ref{tab:ablation_mathvista}. The arrow highlights the optimal "knee" point on the Pareto frontier (our default configuration $\tau_{jump}=0.03, \gamma=0.6$), which achieves the best balance between preserving reasoning quality and maximizing inference acceleration.}
    \label{fig:appendix_scatter}
\end{figure*}

Table~\ref{tab:bbox_attention} reports the results on three representative DMLLMs. Across all models, Seer consistently reduces both attention entropy and irrelevant-region attention mass. This indicates that early suffix truncation makes valid-prefix visual attention less diffuse and reduces the fraction of attention assigned to regions outside the target box. These results provide quantitative evidence consistent with our interpretation that Seer suppresses suffix-mediated interference, although we do not view them as a causal proof of the mechanism.

\begin{table}[htbp]
\centering
\caption{Bounding-box-grounded attention analysis on 100 MME samples with explicit bounding boxes. Lower is better for both metrics.}
\label{tab:bbox_attention}
\normalsize
\setlength{\tabcolsep}{5pt}
\begin{tabular}{lcccc}
\toprule
\multirow{2}{*}{Model} 
& \multicolumn{2}{c}{\makecell{Normalized \\ Attention Entropy $\downarrow$}} 
& \multicolumn{2}{c}{\makecell{Irrelevant-Region \\ Attention Mass $\downarrow$}} \\
\cmidrule(lr){2-3} \cmidrule(lr){4-5}
& Baseline & Seer & Baseline & Seer \\
\midrule
LaViDa-LLaDA & 0.612 & 0.583 & 0.431 & 0.381 \\
LaViDa-Dream & 0.634 & 0.603 & 0.509 & 0.488 \\
MMaDA        & 0.671 & 0.568 & 0.462 & 0.404 \\
\bottomrule
\end{tabular}
\end{table}

As shown in Table~\ref{tab:bbox_attention}, Seer consistently reduces both normalized attention entropy and irrelevant-region attention mass across all three DMLLMs. This suggests that, after early suffix truncation, valid generated tokens attend to a less diffuse subset of visual tokens and allocate a smaller fraction of their visual attention to regions outside the target bounding box. Notably, the reduction is most pronounced on MMaDA, which also exhibits the highest baseline entropy, suggesting that models with more diffuse cross-modal attention may benefit more from suffix truncation. Overall, these findings provide region-grounded quantitative evidence that Seer reduces visually irrelevant interference in the retained text prefix, complementing the qualitative attention maps in Fig.~2.

\begin{figure*}[ht]
    \centering
    \includegraphics[width=\linewidth]{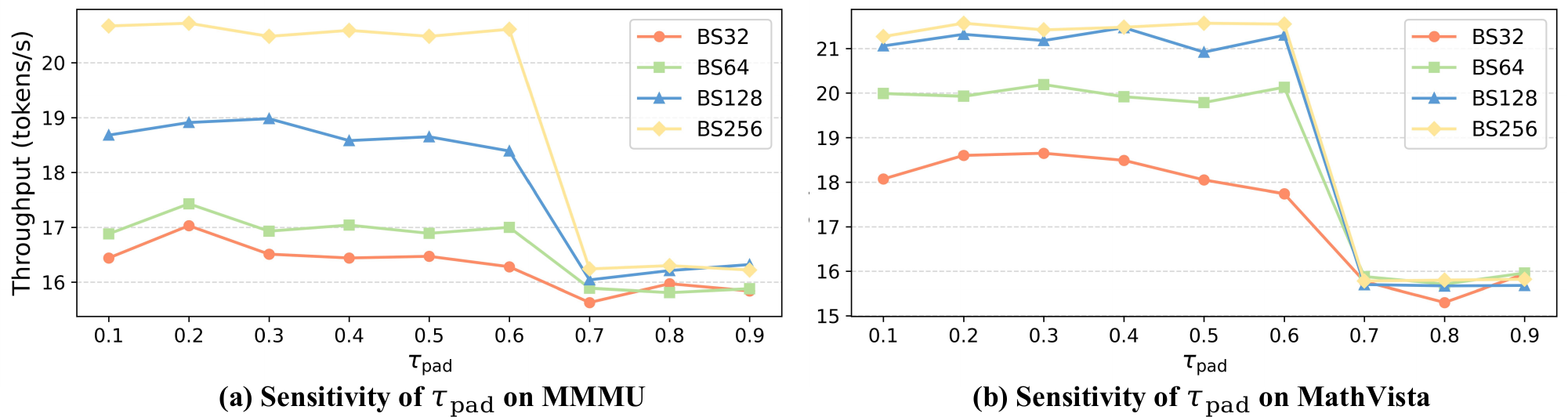} 
    \vspace{-5mm}
    \caption{\textbf{Sensitivity analysis of the padding waste threshold ($\tau_{pad}$).} Effect of varying $\tau_{pad}$ on inference throughput across different batch sizes on the (a) MMMU and (b) MathVista datasets. The system performance is highly robust and stable when $\tau_{pad} < 0.7$ (e.g., $0.1 \le \tau_{pad} \le 0.6$). However, reaching or exceeding this tipping point ($\tau_{pad} \ge 0.7$) causes a sharp throughput drop across all batch sizes, as excessively high thresholds force long-tailed sequences into static execution paths, reintroducing massive padding waste.}
    \label{fig:tpad_sensitivity_appendix}
\end{figure*}

\section{Additional Ablation Studies}

\textbf{Joint Impact of Hyperparameters on MMMU and MathVista.} To comprehensively evaluate the robustness of our dynamic truncation mechanism across diverse reasoning tasks, we extend the joint hyperparameter ablation study to the MMMU and MathVista datasets using the LaViDa-LLaDA architecture. As presented in Table~\ref{tab:ablation_mmmu} and Table~\ref{tab:ablation_mathvista}, the Absolute Jump Threshold ($\tau_{jump}$) and Tolerance Ratio ($\gamma$) jointly govern the trade-off between generation accuracy and inference throughput. We visualize this relationship in Figure~\ref{fig:appendix_scatter}, which clearly maps out a Pareto frontier for both benchmarks. Relaxing the boundary detection criteria (i.e., increasing $\gamma$ or decreasing $\tau_{jump}$) triggers more aggressive truncation, yielding exponential throughput gains but eventually leading to severe accuracy degradation due to the premature removal of valid semantic tokens. Conversely, overly strict criteria perfectly preserve accuracy but severely limit the acceleration benefits. Consistently, our default configuration ($\tau_{jump}=0.03, \gamma=0.6$) emerges as the optimal ``knee'' point on the Pareto frontiers across both datasets. At this specific configuration, Seer successfully strikes the best balance, achieving substantial inference speedups while strictly maintaining highly competitive reasoning quality (e.g., maintaining an accuracy of 43.9\% and 45.1\% on MMMU and MathVista, respectively), thereby demonstrating the task-agnostic reliability and generalizability of the proposed SNR-aware boundary detector.

\textbf{Sensitivity of Padding Waste Threshold on Diverse Tasks.} 
In addition to the semantic boundary detection parameters, the padding waste threshold ($\tau_{pad}$) acts as the critical gatekeeper in our Hybrid Execution Routing policy. To evaluate its generalizability, we conducted a sensitivity analysis on the MMMU and MathVista datasets, as illustrated in Figure~\ref{fig:tpad_sensitivity_appendix}. The results reveal a highly consistent and robust trend across both benchmarks: the end-to-end throughput remains stable and is largely insensitive to the exact value of $\tau_{pad}$ as long as it stays strictly below a critical tipping point ($\tau_{pad} < 0.7$). Within this safe zone (i.e., $\tau_{pad} \in [0.1, 0.6]$), the routing mechanism successfully filters out highly skewed batches (dispatching them to variable-length or eager paths) while reserving the highly optimized Static-Graph path strictly for uniform batches. However, once $\tau_{pad}$ reaches $0.7$ or above, a sharp and simultaneous drop in throughput is observed across all batch sizes. This occurs because an excessively lenient threshold forces long-tailed batches into static buckets, where a single straggler sequence coerces the entire batch to pad heavily, thereby nullifying the computational savings gained from Step-0 truncation. These findings confirm the indispensability of our padding-waste-aware routing design and demonstrate that setting a moderately conservative threshold (e.g., between 0.2 and 0.6) serves as a universally reliable default for diverse multimodal tasks.

\textbf{System Overhead Breakdown.} 
To further validate the efficiency of our system-level co-design, we profile the proportional time breakdown of the end-to-end inference pipeline across different system components on the LaViDa-LLaDA architecture. As detailed in Table~\ref{tab:time_breakdown} (with values averaged across four batch sizes), the actual model computation heavily dominates the inference process, accounting for 95.05\% to 99.52\% of the total time across the evaluated datasets. Crucially, the core mechanisms introduced by Seer---namely, the SNR-aware boundary detection and the hybrid execution routing/dispatch---introduce negligible latency overhead, collectively consuming less than 0.5\% of the total execution time. Furthermore, the overhead from sequence repacking (Packing/compaction) and the device-resident Triton fused kernel operations is strictly bounded (peaking at a combined 4.75\% on ChartQA) and even drops to 0\% on MathVista, where sequences are predominantly handled by static or eager paths without the need for varlen repacking. This highly lightweight profile conclusively demonstrates that Seer successfully bypasses CPU-GPU synchronization bottlenecks, ensuring that the theoretical FLOP reductions are seamlessly converted into robust wall-clock acceleration without incurring prohibitive system overhead.

\begin{table}[htbp]
\centering
\caption{Proportional breakdown of end-to-end inference time (\%) across different system components for the LaViDa-LLaDA architecture. The reported values are averaged across four batch sizes (32, 64, 128, and 256).}
\label{tab:time_breakdown}
\begin{tabular}{lccc}
\toprule
\textbf{Component} & \textbf{ChartQA} & \textbf{MMMU} & \textbf{MathVista} \\
\midrule
Model computation    & 95.05 & 97.425 & 99.525 \\
Boundary detection   & 0.10  & 0.225  & 0.275 \\
Routing / dispatch   & 0.10  & 0.175  & 0.200 \\
Packing / compaction & 1.70  & 0.725  & 0.000 \\
Triton fused kernel  & 3.05  & 1.450  & 0.000 \\
\bottomrule
\end{tabular}
\end{table}

\begin{table*}[!ht]
\centering
\caption{Sensitivity analysis of the look-ahead window size on the LaViDa-LLaDA architecture. The horizontal axis represents the varying local window sizes used in the SNR-aware boundary detector.}
\label{tab:window_size_sensitivity}
\begin{tabular}{ll ccccccc}
\toprule
\multirow{2}{*}{\textbf{Benchmark}} & \multirow{2}{*}{\textbf{Metric}} & \multicolumn{7}{c}{\textbf{Look-Ahead Window Size}} \\
\cmidrule(lr){3-9}
& & \textbf{2} & \textbf{3} & \textbf{4} & \textbf{5} & \textbf{6} & \textbf{8} & \textbf{10} \\
\midrule
\multirow{3}{*}{\textbf{ChartQA}}
& Score $\uparrow$ & 22.10 & 57.80 & 57.80 & 60.40 & 57.40 & 60.60 & 60.60 \\
& Throughput (tokens/s) $\uparrow$ & 5.74 & 3.18 & 3.11 & 2.20 & 1.71 & 1.47 & 1.47 \\
& Latency (s) $\downarrow$ & 0.4367 & 0.8816 & 0.9033 & 1.3219 & 1.7024 & 1.9831 & 1.9925 \\
\midrule
\multirow{3}{*}{\textbf{MMMU}}
& Score $\uparrow$ & 38.11 & 43.89 & 43.74 & 43.90 & 44.00 & 44.10 & 44.00 \\
& Throughput (tokens/s) $\uparrow$ & 1.91 & 1.39 & 0.99 & 0.83 & 0.62 & 0.59 & 0.60 \\
& Latency (s) $\downarrow$ & 0.5581 & 0.8100 & 1.1280 & 1.4479 & 1.8192 & 1.9047 & 1.8677 \\
\midrule
\multirow{3}{*}{\textbf{MathVista}}
& Score $\uparrow$ & 15.73 & 45.10 & 44.69 & 45.28 & 45.70 & 45.70 & 45.65 \\
& Throughput (tokens/s) $\uparrow$ & 3.70 & 1.68 & 0.65 & 0.21 & 0.16 & 0.16 & 0.16 \\
& Latency (s) $\downarrow$ & 0.2889 & 0.7500 & 2.1100 & 6.7300 & 8.8200 & 8.7200 & 8.6200 \\
\bottomrule
\end{tabular}
\end{table*}

\textbf{Sensitivity of Look-Ahead Window Size.} To evaluate the robustness of the local look-ahead mechanism in our SNR-aware boundary detector, we conducted a sensitivity analysis on the window size parameter using the LaViDa-LLaDA architecture across the ChartQA, MMMU, and MathVista benchmarks, as shown in Table~\ref{tab:window_size_sensitivity}. The window size determines the local context used to verify the MLP sparsity ``jump-to-plateau'' structural pattern. When the window is extremely small (e.g., size 2), the detection criterion becomes overly aggressive and brittle to local activation noise, leading to the premature truncation of semantically valid tokens; this causes a catastrophic degradation in reasoning accuracy, with scores plummeting to 22.10 on ChartQA and 15.73 on MathVista. Conversely, as the window size increases excessively (e.g., size 8 or 10), the boundary detector becomes overly conservative and struggles to trigger. While this safely preserves accuracy, the truncation mechanism is rendered largely ineffective, causing inference throughput to plummet and latency to climb back to unoptimized baseline levels (e.g., dropping to 0.16 tokens/s on MathVista). A moderate window size (e.g., size 3) emerges as the optimal configuration, striking a perfect balance that reliably captures the true semantic boundary without risking over-truncation, thereby yielding substantial end-to-end speedups while fully preserving generation quality.


\section{Extended Visualizations of Step-0 Layer-wise MLP Sparsity}
\label{sec:appendix_sparsity_visualizations}

To further substantiate the empirical observations in Section 3.2, we provide extended visualizations of Step-0 layer-wise MLP sparsity across diverse generation lengths, consistently demonstrating why early-layer sparsity serves as a robust semantic oracle. In the heatmaps below, the black bounding boxes highlight the early layers (L1--L6) where the ``Semantic Jump'' is most pronounced: semantically valid prefix tokens exhibit distinctively dense activations (lighter colors), whereas sparsity sharply spikes exactly at the predicted [EOT] token and subsequent [EOS] padding suffix (blue bounding boxes) to form a stable ``Padding Plateau.'' Furthermore, these visualizations justify our algorithmic design of strictly relying on early layers for boundary detection. Layer 0 exhibits uniform, uninformative high sparsity across all tokens, while deeper layers (L9--L31) suffer from extensive information mixing via bidirectional self-attention, which homogenizes the entire sequence into near-zero sparsity (pale yellow) and completely blurs the semantic boundary. Therefore, isolating the sparsity signal within the early layers is both empirically necessary and highly effective across varying sequence lengths.

\begin{figure*}[htbp]
    \centering
    \includegraphics[width=\linewidth]{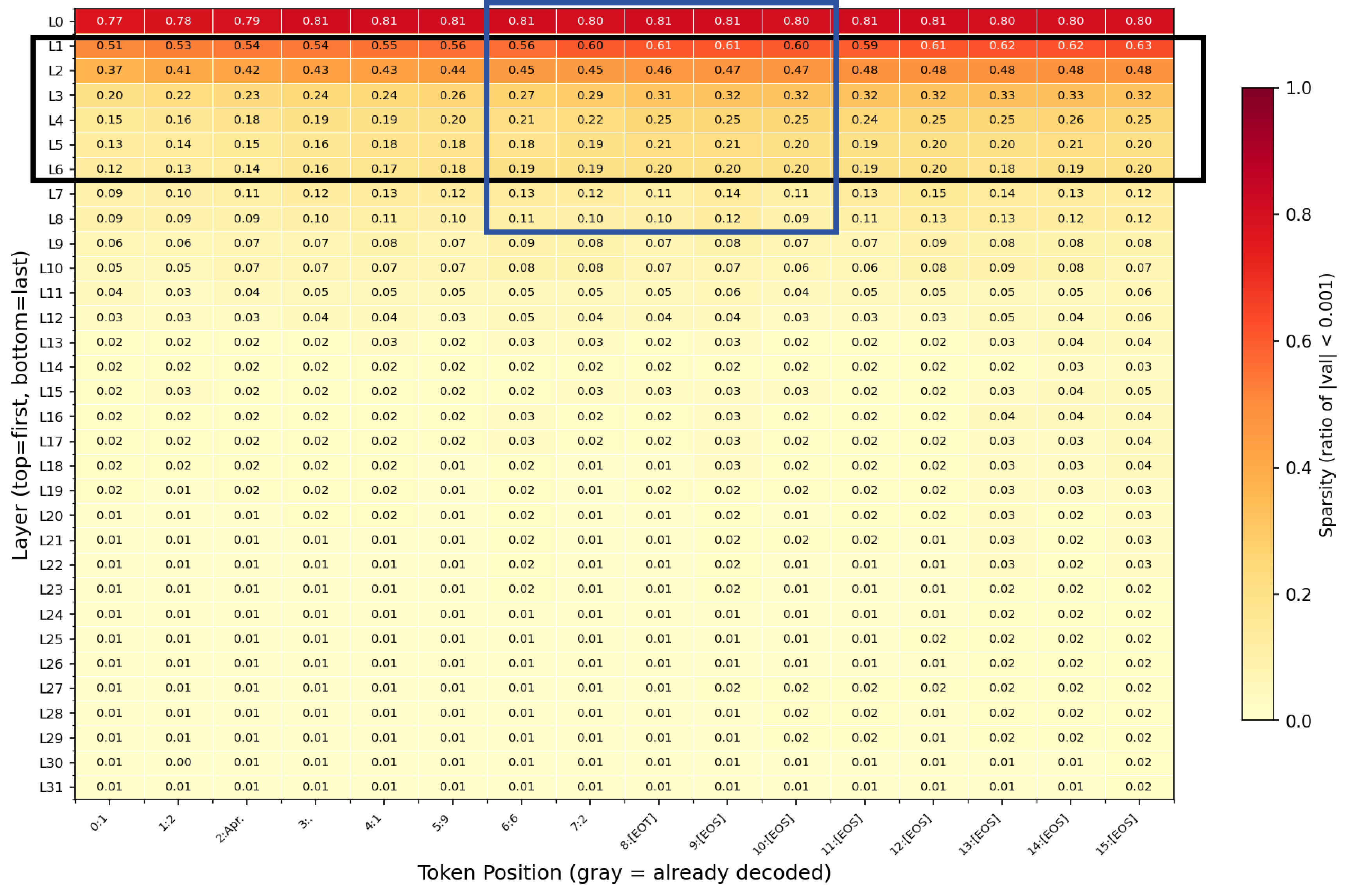}
    \vspace{-8mm}
    \caption{Step-0 MLP sparsity heatmap for an extremely short response (Generation Length = 16). The semantic boundary is distinctly captured by the sudden sparsity increase within the early layers (L1--L6, black box) at the [EOT] token.}
    \label{fig:heat_len16_pos2}
    \vspace{-8mm}
\end{figure*}

\begin{figure*}[htbp]
    \centering
    \includegraphics[width=\linewidth]{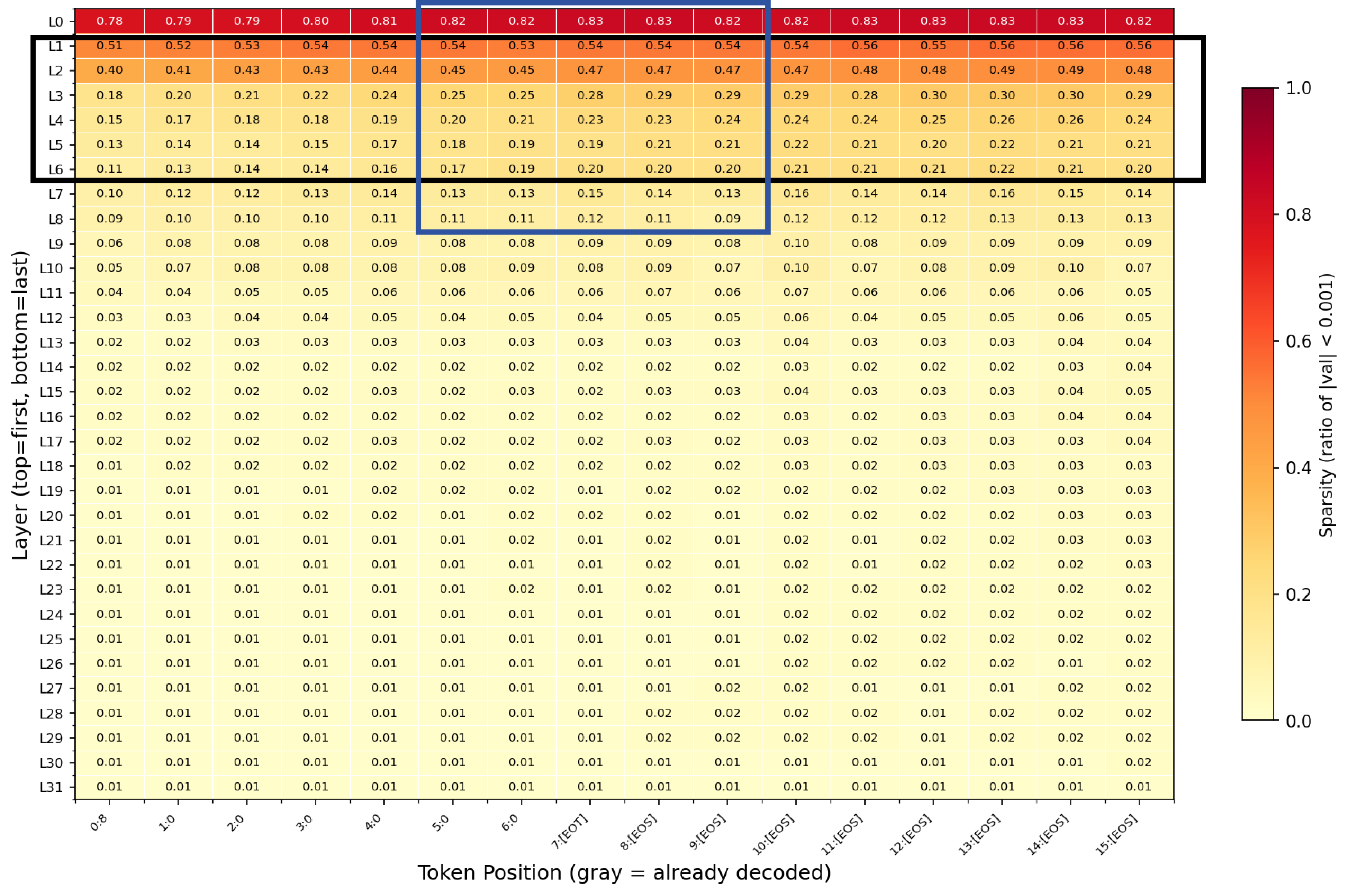}
    \vspace{-8mm}
    \caption{Step-0 MLP sparsity heatmap for a short response (Generation Length = 16). The blue box highlights the sharp ``Semantic Jump'' precisely at the transition to the [EOT] token.}
    \label{fig:heat_len16_pos3}
    \vspace{-8mm}
\end{figure*}

\begin{figure*}[htbp]
    \centering
    \includegraphics[width=\linewidth]{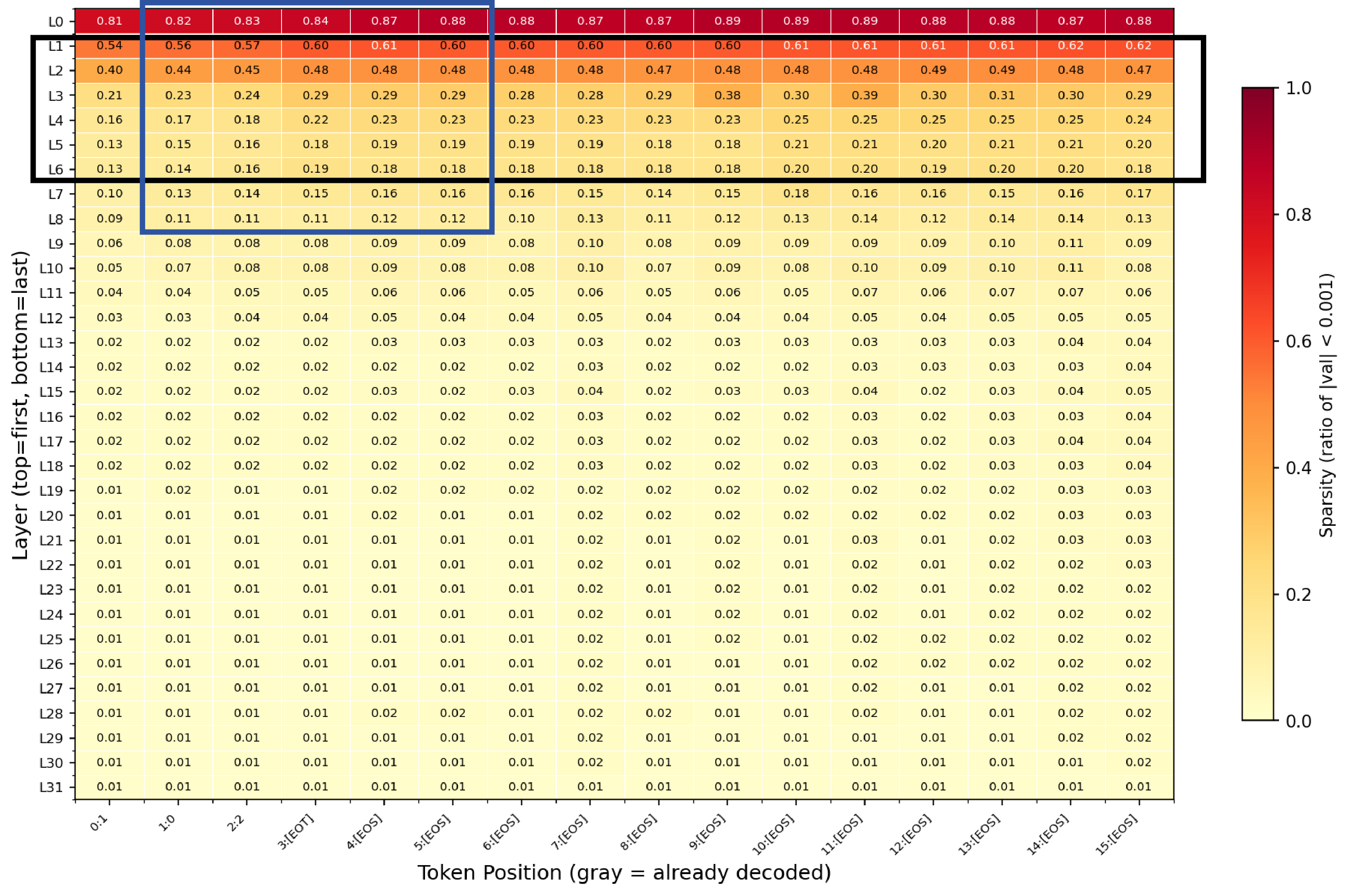}
    \vspace{-8mm}
    \caption{Step-0 MLP sparsity heatmap (Generation Length = 16). The early layers consistently maintain a stable ``Padding Plateau'' immediately following the valid semantic tokens.}
    \label{fig:heat_len16_pos7}
    \vspace{-8mm}
\end{figure*}

\begin{figure*}[htbp]
    \centering
    \includegraphics[width=\linewidth]{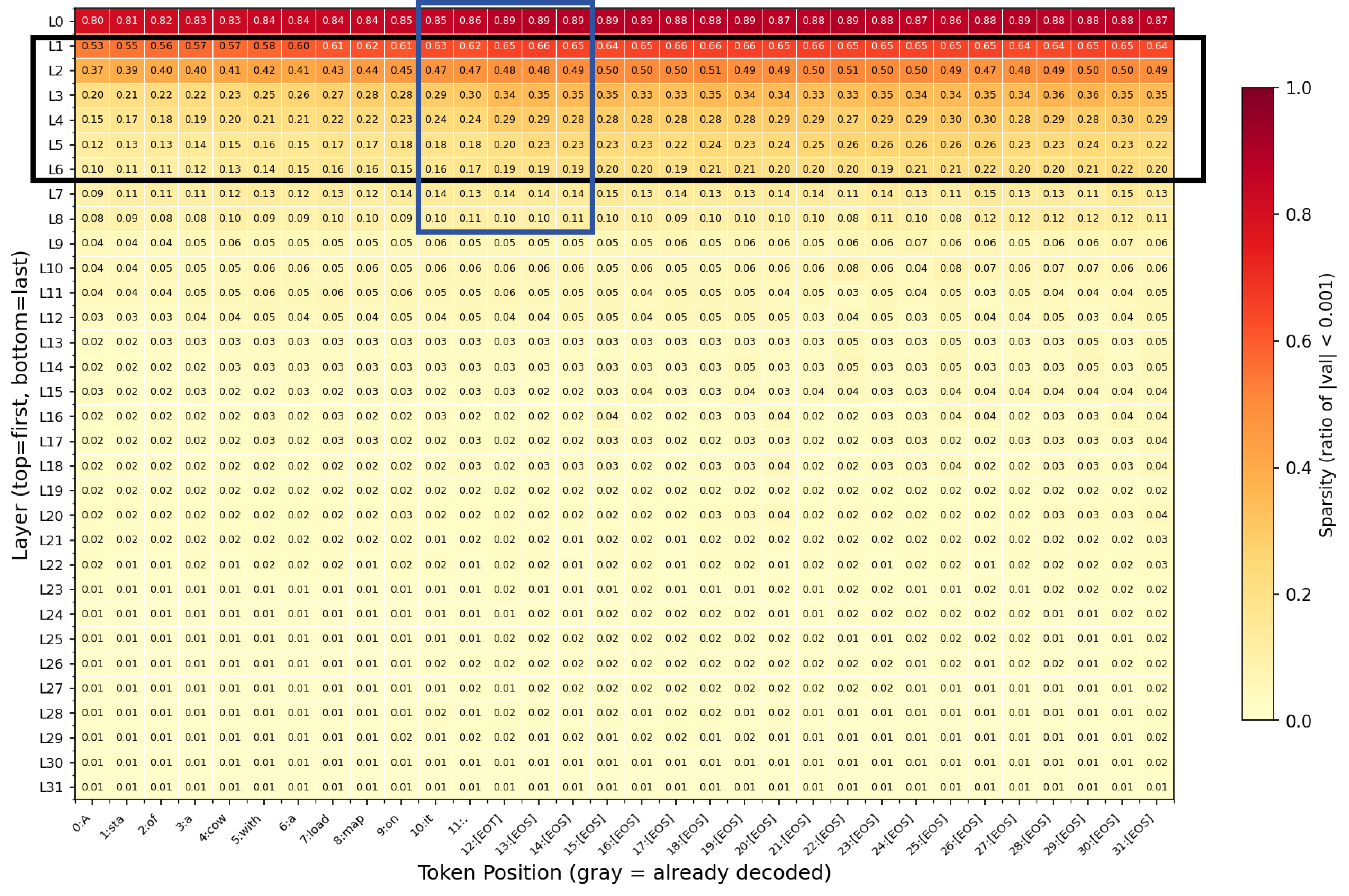}
    \vspace{-8mm}
    \caption{Step-0 MLP sparsity heatmap (Generation Length = 16) demonstrating the contrast between the informative early layers (black box) and the homogenized dense activations in deeper layers.}
    \label{fig:heat_len16_pos8}
    \vspace{-8mm}
\end{figure*}

\begin{figure*}[htbp]
    \centering
    \includegraphics[width=\linewidth]{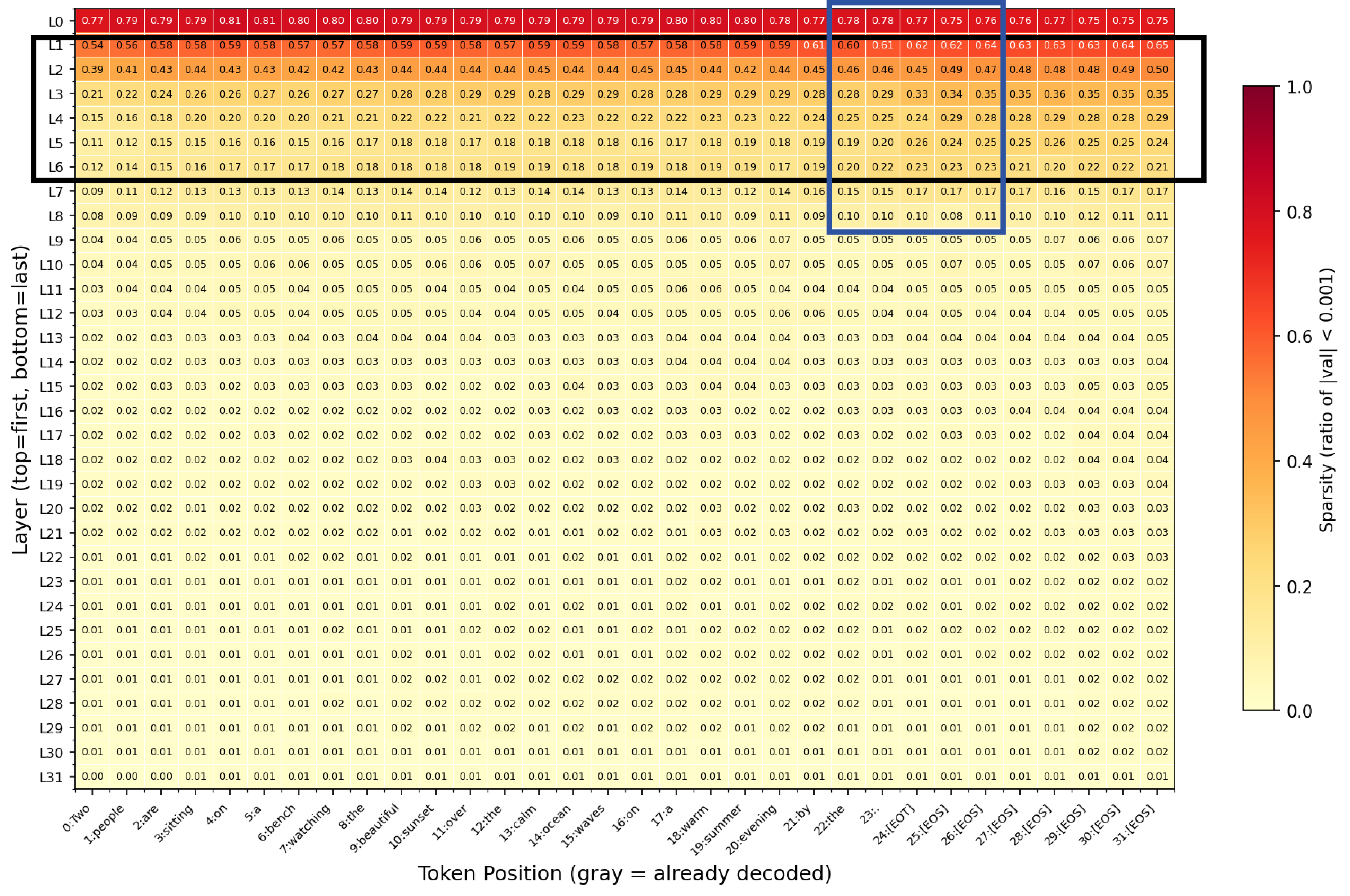}
    \vspace{-8mm}
    \caption{Step-0 MLP sparsity heatmap for a medium-length response (Generation Length = 32), showing a highly consistent boundary signal regardless of the prompt content.}
    \label{fig:heat_len32_pos10}
    \vspace{-8mm}
\end{figure*}

\begin{figure*}[htbp]
    \centering
    \includegraphics[width=\linewidth]{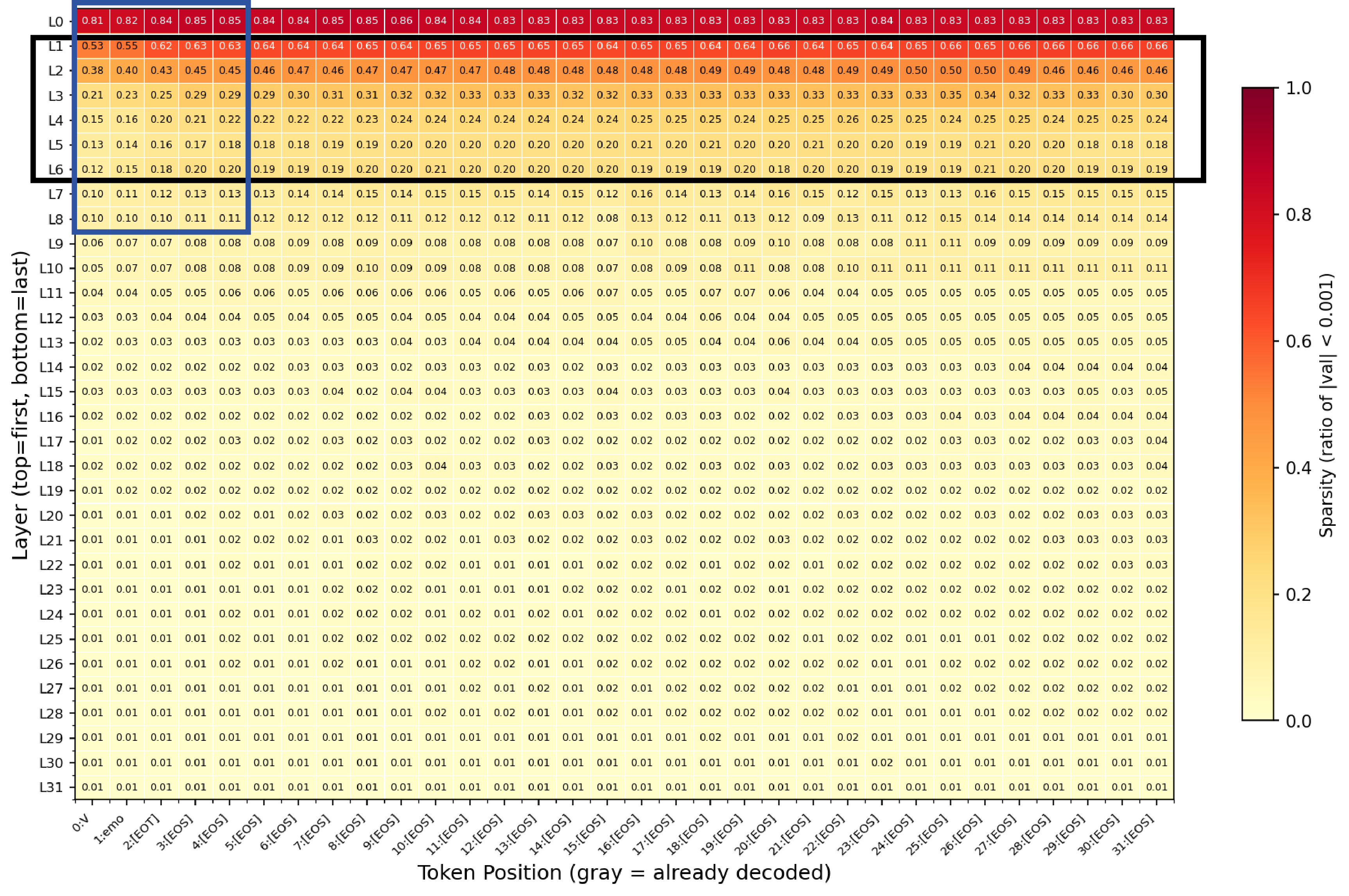}
    \vspace{-8mm}
    \caption{Step-0 MLP sparsity heatmap (Generation Length = 32). The color shift within the L1--L6 window accurately pinpoints the end of the semantic prefix.}
    \label{fig:heat_len32_pos12}
    \vspace{-8mm}
\end{figure*}

\begin{figure*}[htbp]
    \centering
    \includegraphics[width=\linewidth]{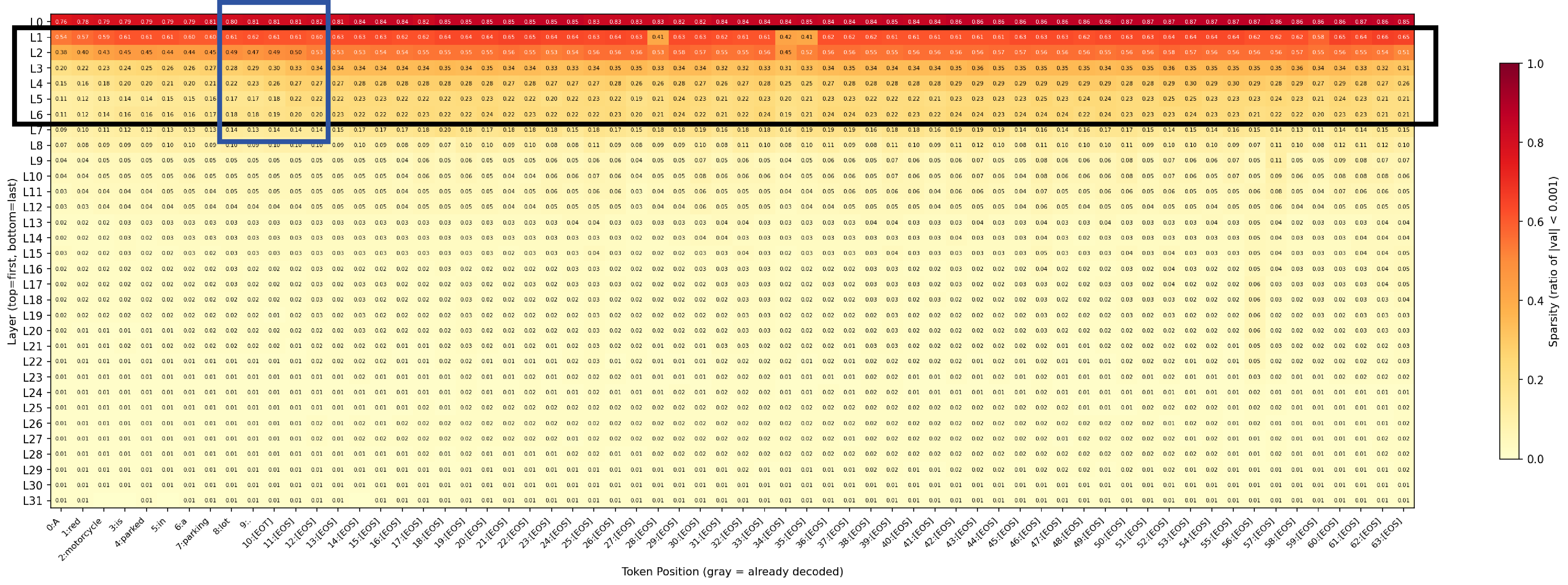}
    \vspace{-8mm}
    \caption{Step-0 MLP sparsity heatmap for a long response (Generation Length = 64). Even as the valid prefix extends, the early-layer sparsity jump remains remarkably robust.}
    \label{fig:heat_len64_pos24}
    \vspace{-8mm}
\end{figure*}

\begin{figure*}[htbp]
    \centering
    \includegraphics[width=\linewidth]{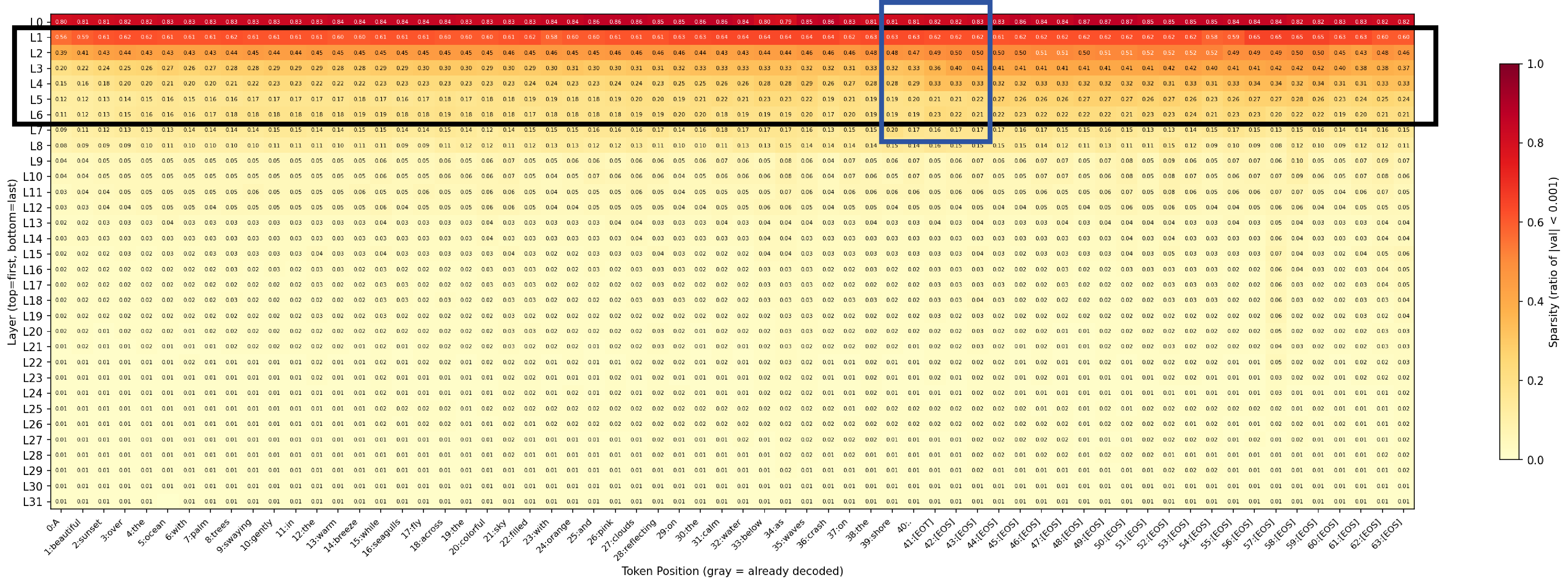}
    \vspace{-8mm}
    \caption{Step-0 MLP sparsity heatmap for an extremely long response (Generation Length = 64). The visualization confirms that the boundary detection phenomenon scales stably to extensive output sequences.}
    \label{fig:heat_len64_pos41}
    \vspace{-8mm}
\end{figure*}

\end{document}